\documentclass[nohyperref]{article}
\usepackage{hyperref}

\usepackage[accepted]{icml2022}
\usepackage{microtype}
\usepackage{graphicx}
\usepackage{subfigure}
\usepackage{booktabs} 
\usepackage{multicol,multirow}
\usepackage{bbding}
\usepackage{amsmath}
\usepackage{amssymb}
\usepackage{mathtools}
\usepackage{amsthm}

\newcommand{\blue}[1]{\textcolor{blue}{#1}}

\theoremstyle{plain}
\newtheorem{theorem}{Theorem}[section]
\newtheorem{proposition}[theorem]{Proposition}
\newtheorem{lemma}[theorem]{Lemma}

\theoremstyle{definition}

\newtheorem{assumption}[theorem]{Assumption}
\theoremstyle{remark}

\DeclareMathOperator*{\argmin}{arg\,min}
\DeclareMathOperator*{\argmax}{arg\,max}
\newcommand{\sumA}{\sum_{a \in \mathcal{A}}}
\newcommand{\sumE}{\sum_{e \in \mathcal{E}}}
\newcommand{\mE}{\mathbb{E}}
\newcommand{\mV}{\mathbb{V}}

\newcommand{\calD}{\mathcal{D}}
\newcommand{\calX}{\mathcal{X}}
\newcommand{\calA}{\mathcal{A}}
\newcommand{\calO}{\mathcal{O}}
\newcommand{\calU}{\mathcal{U}}
\newcommand{\calE}{\mathcal{E}}

\newcommand{\trueV}{V(\pi)}
\newcommand{\ips}{\hat{V}_{\mathrm{IPS}} (\pi; \calD)}
\newcommand{\mips}{\hat{V}_{\mathrm{MIPS}} (\pi; \calD)}
\newcommand{\ipsnoD}{\hat{V}_{\mathrm{IPS}} (\pi)}
\newcommand{\mipsnoD}{\hat{V}_{\mathrm{MIPS}} (\pi)}

\newcommand{\dr}{\hat{V}_{\mathrm{DR}} (\pi; \calD, \hat{q})}
\newcommand{\dm}{\hat{V}_{\mathrm{DM}} (\pi; \calD, \hat{q})}
\newcommand{\drs}{\hat{V}_{\mathrm{DRos}} (\pi; \calD, \hat{q}, \lambda)}
\newcommand{\mrdr}{\hat{V}_{\mathrm{MRDR}} (\pi; \calD, \hat{q}_{\mathrm{MRDR}})}
\newcommand{\switchdr}{\hat{V}_{\mathrm{SwitchDR}} (\pi; \calD, \hat{q}, \lambda)}
\newcommand{\drlam}{\hat{V}_{\mathrm{DR}-\lambda} (\pi; \calD, \hat{q}, \lambda)}

\newcommand{\mse}{\mathrm{MSE} ( \hat{V} (\pi ) )}
\newcommand{\bias}{\mathrm{Bias} ( \hat{V} (\pi) )}
\newcommand{\biasmips}{\mathrm{Bias} ( \hat{V}_{\mathrm{MIPS}} (\pi) )}
\newcommand{\var}{\mV_{\calD} \big[ \hat{V} (\pi; \calD ) \big]}
\newcommand{\mseratio}{\frac{\mathrm{MSE}(\hat{V}_{\mathrm{IPS}})}{\mathrm{MSE}(\hat{V}_{\mathrm{MIPS}})}}

\begin{document}

\twocolumn[
\icmltitle{Off-Policy Evaluation for Large Action Spaces via Embeddings}

\begin{icmlauthorlist}
\icmlauthor{Yuta Saito}{cornell}
\icmlauthor{Thorsten Joachims}{cornell}
\end{icmlauthorlist}
\icmlaffiliation{cornell}{Department of Computer Science, Cornell University, Ithaca, NY, USA}
\icmlcorrespondingauthor{Yuta Saito}{ys552@cornell.edu}
\icmlcorrespondingauthor{Thorsten Joachims}{tj@cs.cornell.edu}

\vskip 0.3in
]

\printAffiliationsAndNotice{} 

\begin{abstract}
Off-policy evaluation (OPE) in contextual bandits has seen rapid adoption in real-world systems, since it enables offline evaluation of new policies using only historic log data. Unfortunately, when the number of actions is large, existing OPE estimators -- most of which are based on inverse propensity score weighting -- degrade severely and can suffer from extreme bias and variance. This foils the use of OPE in many applications from recommender systems to language models. To overcome this issue, we propose a new OPE estimator that leverages \textit{marginalized} importance weights when \textit{action embeddings} provide structure in the action space. We characterize the bias, variance, and mean squared error of the proposed estimator and analyze the conditions under which the action embedding provides statistical benefits over conventional estimators. In addition to the theoretical analysis, we find that the empirical performance improvement can be substantial, enabling reliable OPE even when existing estimators collapse due to a large number of actions.
\end{abstract}

\section{Introduction} \label{sec:intro}
Many intelligent systems (e.g., recommender systems, voice assistants, search engines) interact with the environment through a \textit{contextual bandit} process where a policy observes a context, takes an action, and obtains a reward. Logs of these interactions provide valuable data for \textit{off-policy evaluation} (OPE), which aims to accurately evaluate the performance of new policies without ever deploying them in the field. OPE is of great practical relevance, as it helps avoid costly online A/B tests and can also act as subroutines for batch policy learning~\cite{dudik2014doubly,su2020doubly}. However, OPE is challenging, since the logs contain only partial-information feedback -- specifically the reward of the chosen action, but not the counterfactual rewards of all the other actions a different policy might choose.

When the action space is small, recent advances in the design of OPE estimators have led to a number of reliable methods with good theoretical guarantees~\cite{dudik2014doubly,swaminathan2015batch,wang2017optimal,farajtabar2018more,su2019cab,su2020doubly,metelli2021subgaussian}. Unfortunately, these estimators can degrade severely when the number of available actions is large.
Large action spaces are prevalent in many potential applications of OPE, such as recommender systems where policies have to handle thousands or millions of items (e.g., movies, songs, products). In such a situation, the existing estimators based on \textit{inverse propensity score} (IPS) weighting~\cite{horvitz1952generalization} can incur high bias and variance, and as a result, be impractical for OPE. First, a large action space makes it challenging for the logging policy to have common support with the target policies, and IPS is biased under support deficiency~\cite{sachdeva2020off}. Second, a large number of actions typically leads to high variance of IPS due to large importance weights. To illustrate, we find in our experiments that the variance and mean squared error of IPS inflate by a factor of over 300 when the number of actions increases from 10 to 5000 given a fixed sample size. While doubly robust (DR) estimators can somewhat reduce the variance by introducing a reward estimator as a control variate~\citep{dudik2014doubly}, they do not address the fundamental issues that come with large action spaces.

To overcome the limitations of the existing estimators when the action space is large, we leverage additional information about the actions in the form of \textit{action embeddings}. There are many cases where we have access to such prior information. For example, movies are characterized by auxiliary information such as genres (e.g., adventure, science fiction, documentary), director, or actors. We should then be able to utilize these supplemental data to infer the value of actions under-explored by the logging policy, potentially achieving much more accurate policy evaluation than the existing estimators. We first provide the conditions under which action embeddings provide another path for unbiased OPE, even with support deficient actions. We then propose the \textit{Marginalized IPS} (MIPS) estimator, which uses the \textit{marginal} distribution of action embeddings, rather than actual actions, to define a new type of importance weights. We show that MIPS is unbiased under an alternative condition, which states that the action embeddings should mediate every causal effect of the action on the reward. Moreover, we show that MIPS has a lower variance than IPS, especially when there is a large number of actions, and thus the vanilla importance weights have a high variance. We also characterize the gain in MSE provided by MIPS, which implies an interesting bias-variance trade-off with respect to the quality of the action embeddings. Including many embedding dimensions captures the causal effect better, leading to a smaller bias of MIPS. In contrast, using only a subset of the embedding dimensions reduces the variance more. We thus propose a strategy to intentionally violate the assumption about the action embeddings by discarding less relevant embedding dimensions for achieving a better MSE at the cost of introducing some bias. Comprehensive experiments on synthetic and real-world bandit data verify the theoretical findings, indicating that MIPS can provide an effective bias-variance trade-off in the presence of many actions.

\section{Off-Policy Evaluation} \label{sec:ope}
We follow the general contextual bandit setup, and an extensive discussion of related work is given in Appendix~\ref{app:related}. 
Let $x \in \calX \subseteq\mathbb{R}^{d_x}$ be a $d_x$-dimensional context vector drawn i.i.d. from an unknown distribution $p(x)$. Given context $x$, a possibly stochastic \textit{policy} $\pi(a|x)$ chooses action $a$ from a finite action space denoted as $\calA$. The reward $r \in [0, r_{\mathrm{max}}]$ is then sampled from an unknown conditional distribution $p(r|x,a)$. We measure the effectiveness of a policy $\pi$ through its \textit{value} 
\begin{align}
    \trueV := \mE_{p(x) \pi (a | x) p(r|x,a)} [r] = \mE_{p(x) \pi (a | x) } [q (x,a)],  
    \label{eq:value}
\end{align}
where $q(x,a) := \mE [r|x,a]$ denotes the expected reward given context $x$ and action $a$.

In OPE, we are given logged bandit data collected by a logging policy $\pi_0$. Specifically, let $\calD := \{(x_i,a_i,r_i)\}_{i=1}^n$ be a sample of logged bandit data containing $n$ independent observations drawn from the logging policy as $(x,a,r) \sim p(x)\pi_0(a|x)p(r|x,a)$. We aim to develop an estimator $\hat{V}$ for the value of a target policy $\pi$ (which is different from $\pi_0$) using only the logged data in $\calD$. The accuracy of $\hat{V}$ is quantified by its mean squared error (MSE)
\begin{align*}
    \mse 
    :&= \mE_{\calD} \Big[ \big(  \trueV - \hat{V} (\pi; \calD) \big)^2 \Big] \\
    & = \bias^2 + \var,
\end{align*}
where $\mE_{\calD}[\cdot]$ takes the expectation over the logged data and
\begin{align*}
    \bias &:= \mE_{\calD}[\hat{V} (\pi; \calD)] - \trueV ,  \\
    \var & := \mE_{\calD} \Big[ \big(  \hat{V} (\pi; \calD) - \mE_{\calD} [\hat{V} (\pi; \calD)] \big)^2 \Big].
\end{align*}

In the following theoretical analysis, we focus on the IPS estimator, since most advanced OPE estimators are based on IPS weighting~\cite{dudik2014doubly,wang2017optimal,su2019cab,su2020doubly,metelli2021subgaussian}. IPS estimates the value of $\pi$ by re-weighting the observed rewards as follows.
\begin{align*}
    \ips := \frac{1}{n} \sum_{i=1}^n \frac{\pi(a_i | x_i)}{\pi_0 (a_i | x_i)} r_i = \frac{1}{n} \sum_{i=1}^n w(x_i, a_i) r_i
\end{align*}
where $w(x,a) := \pi(a|x) / \pi_0(a|x)$ is called the \textit{(vanilla) importance weight}.

This estimator is unbiased (i.e., $ \mE_{\calD} [\ips] = \trueV $) under the following common support assumption. 
\begin{assumption} (Common Support) \label{assumption:common_support}
The logging policy $\pi_0$ is said to have common support for policy $\pi$ if $\pi (a | x) > 0 \rightarrow \pi_0(a | x) > 0$ for all $a \in \calA$ and $x \in \calX$.
\end{assumption}

The unbiasedness of IPS is desirable, making this simple re-weighting technique so popular. However, IPS can still be highly biased, particularly when the action space is large. \citet{sachdeva2020off} indicate that IPS has the following bias when Assumption~\ref{assumption:common_support} is not true.
\begin{align*}
    \big|\mathrm{Bias} (\ipsnoD) \big|
    = \mE_{p(x)} \left[ \sum_{a \in \calU_0(x,\pi_0)} \pi(a | x) q (x,a) \right],
\end{align*}
where $\calU_0(x,\pi_0) := \{a \in \calA \mid \pi_0 (a | x) = 0\} $ is the set of unsupported or deficient actions for context $x$ under $\pi_0$. Note that $\calU_0(x,\pi_0)$ can be large especially when $\calA$ is large. This bias is due to the fact that the logged dataset $\calD$ does not contain any information about the unsupported actions.

Another critical issue of IPS is that its variance can be large, which is given as follows~\cite{dudik2014doubly}.
\begin{align}
    n \mV_{\calD} \big[  \ips \big] 
    & =  \mE_{p(x)\pi_0(a|x)} [ w(x,a)^2 \sigma^2 (x,a) ] \notag \\
    & + \mV_{p(x)} \left[  \mE_{\pi_0(a|x)} [ w(x,a) q (x,a) ] \right] \notag \\
    & + \mE_{p(x)} \left[  \mV_{\pi_0(a|x)} [ w(x,a) q (x,a) ] \right], \label{eq:ips_variance}
\end{align}
where $ \sigma^2 (x,a) := \mV [r|x,a] $.
The variance consists of three terms. The first term reflects the randomness in the rewards. The second term represents the variance due to the randomness over the contexts. The final term is the penalty arising from the use of IPS weighting, and it is proportional to the weights and the true expected reward. The variance contributed by the first and third terms can be extremely large when the weights $w(x,a)$ have a wide range, which occurs when $\pi$ assigns large probabilities to actions that have low probability under $\pi_0$. The latter can be expected when the action space $\calA$ is large and the logging policy $\pi_0$ aims to have {\em universal support} (i.e., $\pi_0(a|x)>0$ for all $a$ and $x$). \citet{swaminathan2017off} also point out that the variance of IPS grows linearly with $w(x,a)$, which can be $\Omega(|\calA|)$.

This variance issue can be lessened by incorporating a reward estimator $\hat{q}(x,a) \approx q(x,a)$ as a control variate, resulting in the DR estimator~\cite{dudik2014doubly}. DR often improves the MSE of IPS due to its variance reduction property. However, DR still suffers when the number of actions is large, and it can experience substantial performance deterioration as we demonstrate in our experiments.

\section{The Marginalized IPS Estimator} \label{sec:method}
The following proposes a new estimator that circumvents the challenges of IPS for large action spaces. Our approach is to bring additional structure into the estimation problem, providing a path forward despite the minimax optimality of IPS and DR. In particular, IPS and DR achieve the minimax optimal MSE of at most $\calO (n^{-1} (\mE_{\pi_0}[w(x,a)^2 \sigma^2 (x,a) + w(x,a)^2 r_{\max}^2 ]))$, which means that they are impossible to improve upon in the worst case beyond constant factors~\cite{wang2017optimal,swaminathan2017off}, unless we bring in additional structure.

Our key idea for overcoming the limits of IPS and DR is to assume the existence of \textit{action embeddings} as prior information. The intuition is that this can help the estimator transfer information between similar actions. More formally, suppose we are given a $d_e$-dimensional \textit{action embedding} $e \in \calE \subseteq\mathbb{R}^{d_e}$ for each action $a$, where we merely assume that the embedding is drawn i.i.d.\ from some unknown distribution $p(e|x,a)$. The simplest example is to construct action embeddings using predefined category information (e.g., product category). Then, the embedding distribution is independent of the context and it is deterministic given the action. Our framework is also applicable to the most general case of continuous, stochastic, and context-dependent action embeddings. For example, product prices may be generated by a personalized pricing algorithm running behind the system. In this case, the embedding is continuous, depends on the user context, and can be stochastic if there is some randomness in the pricing algorithm. 

Using the action embeddings, we now refine the definition of the policy value as:
\begin{align*}
    \trueV &= \mE_{p(x) \pi (a | x) p(e|x,a) p(r|x,a,e)} [r].
\end{align*}
Note here that $q(x,a)=\mE_{p(e|x,a)}[q(x,a,e)]$ where $q(x,a,e):= \mE [r|x,a,e]$, so the above refinement does not contradict the original definition given in Eq.~\eqref{eq:value}.

A logged bandit dataset now contains action embeddings for each observation in $\calD= \{(x_i, a_i, e_i, r_i)\}_{i=1}^n$, where each tuple is generated by the logging policy as $(x,a,e,r) \sim p(x)\pi_0(a|x)p(e|x,a)p(r|x,a,e)$. Our strategy is to leverage this additional information for achieving a more accurate OPE for large action spaces. 

\begin{table*}[h]
\begin{minipage}{\textwidth}
    \caption{A toy example illustrating the benefits of marginal importance weights} \label{tab:example}
    \vspace{0.1in}
  \begin{minipage}[t]{.33\textwidth}
    \centering
    \scalebox{0.9}{
    \begin{tabular}{c|ccc}
        \toprule
         & $\pi_0(a|x_1)$ & $\pi(a|x_1)$ & $w(x_1,a)$ \\ \midrule
        $a_1$ & 0.0 & 0.2 & N/A  \\
        $a_2$ & 0.2 & 0.8 & 4.0  \\
        $a_3$ & 0.8 & 0.0 & 0.0  \\
        \bottomrule
    \end{tabular}}
  \end{minipage}
  \hfill
  \begin{minipage}[t]{.33\textwidth}
    \centering
    \scalebox{0.9}{
      \begin{tabular}{c|ccc}
        \toprule
         & $p(e_1|a)$ & $p(e_2|a)$ & $p(e_3|a)$ \\ \midrule
        $a_1$ & 0.25 & 0.25 & 0.5  \\
        $a_2$ & 0.5 & 0.25 & 0.25  \\
        $a_3$ & 0.25 & 0.5 & 0.25  \\
        \bottomrule
    \end{tabular}}
  \end{minipage}
  \hfill
  \begin{minipage}[t]{.33\textwidth}
    \centering
    \scalebox{0.9}{
      \begin{tabular}{c|ccc}
        \toprule
         & $p(e|x_1,\pi_0)$ & $p(e|x_1,\pi)$ & $w(x_1,e)$ \\ \midrule
        $e_1$ & 0.3 & 0.45 & 1.5  \\
        $e_2$ & 0.45 & 0.25 & 0.55  \\
        $e_3$ & 0.25 & 0.3 & 1.2  \\
        \bottomrule
      \end{tabular}}
  \end{minipage}
\end{minipage}
\end{table*}

To motivate our approach, we introduce two properties characterizing an action embedding.

\begin{assumption} (Common Embedding Support) \label{assumption:common_embed_support}
The logging policy $\pi_0$ is said to have common embedding support for policy $\pi$ if $p (e|x,\pi) > 0 \rightarrow p(e|x,\pi_0) > 0$ for all $e \in \calE$ and $x \in \calX$, where $p(e | x,\pi) := \sumA p(e|x,a) \pi(a | x) $ is the \textit{marginal} distribution over the action embedding space given context $x$ and policy $\pi$.
\end{assumption}

Assumption~\ref{assumption:common_embed_support} is analogous to Assumption~\ref{assumption:common_support}, but requires only the common support with respect to the action embedding space, which can be substantially more compact than the action space itself. Indeed, Assumption~\ref{assumption:common_embed_support} is weaker than common support of IPS (Assumption~\ref{assumption:common_support}).\footnote{First, if Assumption~\ref{assumption:common_support} is true, Assumption~\ref{assumption:common_embed_support} is also true because $p(e|x, a)$ remains the same for the target and logging policies. Table~\ref{tab:example} will provide a counterexample for the opposite statement (i.e., Assumption~\ref{assumption:common_embed_support} does not imply Assumption~\ref{assumption:common_support}).}
Next, we characterize the expressiveness of the embedding in the ideal case, but we will relax this assumption later.

\begin{assumption} (No Direct Effect) \label{assumption:no_direct_effect}
Action $a$ has no direct effect on the reward $r$, i.e., $a \perp r \mid x,e$. 
\end{assumption}

\begin{figure}[t]
\centering
\includegraphics[clip, width=7.0cm]{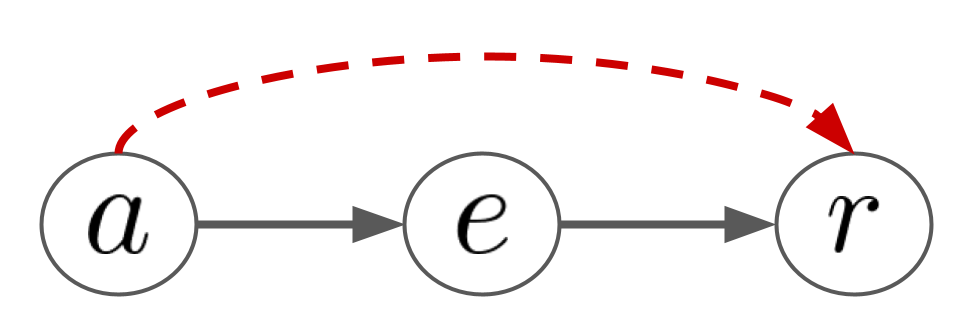}
\caption{Causal Graph Consistent with Assumption~\ref{assumption:no_direct_effect}} \label{fig:no_direct_effect}
\vspace{0.1in}
\raggedright
\fontsize{8.5pt}{8.5pt} \selectfont \textit{Note}:
Grey arrows indicate the existence of causal effect of the tail variable on the head variable. 
The dashed red arrow is a direct causal effect that is ruled out by Assumption~\ref{assumption:no_direct_effect}.
\end{figure}

As illustrated in Figure~\ref{fig:no_direct_effect}, Assumption~\ref{assumption:no_direct_effect} requires that every possible effect of $a$ on $r$ be fully mediated by the observed embedding $e$. For now, we rely on the validity of Assumption~\ref{assumption:no_direct_effect}, as it is convenient for introducing the proposed estimator. However, we later show that it is often beneficial to strategically discard some embedding dimensions and violate the assumption to achieve a better MSE.

We start the derivation of our new estimator with the observation that Assumption~\ref{assumption:no_direct_effect} gives us another path to unbiased estimation of the policy value without Assumption~\ref{assumption:common_support}.

\begin{proposition} \label{prop:identification}
Under Assumption~\ref{assumption:no_direct_effect}, we have
\begin{align*}
    \trueV & = \mE_{p(x) p(e | x,\pi) p(r|x,e) } [r] 
\end{align*}
See Appendix~\ref{app:identification} for the proof.
\end{proposition}

Proposition~\ref{prop:identification} provides another expression of the policy value \textit{without} explicitly relying on the action variable $a$. This new expression naturally leads to the following \textit{marginalized inverse propensity score} (MIPS) estimator, which is our main proposal.
\begin{align*}
    \mips 
    & := \frac{1}{n} \sum_{i=1}^n \!\frac{p(e_i | x_i,\pi)}{p(e_i | x_i,\pi_0)} r_i  = \frac{1}{n}\! \sum_{i=1}^n w(x_i, e_i) r_i,
\end{align*}
where $w(x,e) := p(e|x,\pi)/p(e|x,\pi_0)$ is the \textit{marginal importance weight} defined with respect to the marginal distribution over the action embedding space.

To obtain an intuition for the benefits of MIPS, we provide a toy example in Table~\ref{tab:example} with $\calX = \{x_1\}$, $\calA = \{a_1,a_2,a_3\}$, and $\calE = \{e_1,e_2,e_3\}$ (a special case of our formulation with a discrete embedding space). The left table describes the logging and target policies with respect to $\calA$ and implies that Assumption~\ref{assumption:common_support} is violated ($\pi_0(a_1|x_1)=0.0$). The middle table describes the conditional distribution of the action embedding $e$ given action $a$ (e.g., probability of a movie $a$ belonging to a genre $e$). The right table describes the marginal distributions over $\calE$, which are calculable from the other two tables. By considering the marginal distribution, Assumption~\ref{assumption:common_embed_support} is ensured in the right table, even if Assumption~\ref{assumption:common_support} is not true in the left table. Moreover, the maximum importance weight is smaller for the right table ($\max_{e \in \calE} w (x_1,e) < \max_{a \in \calA} w (x_1,a)$), which may contribute to a variance reduction of the resulting estimator.

Below, we formally analyze the key statistical properties of MIPS and compare them with those of IPS, including the realistic case where Assumption~\ref{assumption:no_direct_effect} is violated.

\subsection{Theoretical Analysis}
First, the following proposition shows that MIPS is unbiased under assumptions different from those of IPS.

\begin{proposition} \label{prop:unbiased}
Under Assumptions~\ref{assumption:common_embed_support} and~\ref{assumption:no_direct_effect}, MIPS is unbiased, i.e., $\mE_{\calD}[\mips] = \trueV$ for any $\pi$. See Appendix~\ref{app:unbiased} for the proof.
\end{proposition}

Proposition~\ref{prop:unbiased} states that, even when $\pi_0$ fails to provide common support over $\calA$ such that IPS is biased, MIPS can still be unbiased if $\pi_0$ provides common support over $\calE$ (Assumption~\ref{assumption:common_embed_support}) and $e$ fully captures the causal effect of $a$ on $r$ (Assumption~\ref{assumption:no_direct_effect}).

Having multiple estimators that enable unbiased OPE under different assumptions is in itself desirable, as we can choose the appropriate estimator depending on the data generating process. However, it is also helpful to understand \textit{how violations of the assumptions influence the bias of the resulting estimator}. In particular, for MIPS, it is difficult to verify whether Assumption~\ref{assumption:no_direct_effect} is true in practice. The following theorem characterizes the bias of MIPS.

\begin{theorem} (Bias of MIPS) \label{thm:bias}
If Assumption~\ref{assumption:common_embed_support} is true, but Assumption~\ref{assumption:no_direct_effect} is violated, MIPS has the following bias.
\begin{align*}
    &\biasmips  &\\
    &= \mE_{p(x)p(e|x,\pi_0)} 
    \bigg[ 
        \sum_{a < b} \;  \pi_0(a|x,e) \pi_0(b|x,e) \\
        & \qquad \qquad \qquad \qquad \qquad \times (q(x,a,e) - q(x,b,e)) \\ 
        & \qquad \qquad \qquad \qquad \qquad \times (w(x,b) - w(x,a))
    \bigg],
\end{align*}
where $a,b \in \calA$. See Appendix~\ref{app:bias} for the proof.
\end{theorem}

Theorem~\ref{thm:bias} suggests that three factors contribute to the bias of MIPS when Assumption~\ref{assumption:no_direct_effect} is violated. The first factor is the \textit{predictivity of the action embeddings with respect to the actual actions}.
When action $a$ is predictable given context $x$ and embedding $e$, $\pi_0(a|x,e)$ is close to zero or one (deterministic), meaning that $\pi_0(a|x,e) \pi_0(b|x,e)$ is close to zero. This suggests that even if Assumption~\ref{assumption:no_direct_effect} is violated, action embeddings that identify the actions well still enable a nearly unbiased estimation of MIPS. 
The second factor is the \textit{amount of direct effect of the action on the reward}, which is quantified by $q(x,a,e) - q(x,b,e)$. When the direct effect of $a$ on $r$ is small, $q(x,a,e) - q(x,b,e)$ also becomes small and so is the bias of MIPS. In an ideal situation where Assumption~\ref{assumption:no_direct_effect} is satisfied, we have $q(x,a,e) = q(x,b,e) = q(x,e)$, thus MIPS is unbiased, which is consistent with Proposition~\ref{prop:unbiased}. Note that the first two factors suggest that, to reduce the bias, the action embeddings should be \textit{informative} so that they are either predictive of the actions or mediate a large amount of the causal effect.
The final factor is the \textit{similarity between logging and target policies} quantified by $w(x,a) - w(x,b)$.
When Assumption~\ref{assumption:no_direct_effect} is satisfied, MIPS is unbiased for any target policy, however, Theorem~\ref{thm:bias} suggests that if the assumption is not true, MIPS produces a larger bias for target policies dissimilar from the logging policy.\footnote{When $\pi = \pi_0$, the bias is zero regardless of the other factors as $ w(x,a) = w(x,b)=1$, meaning that \textit{on-policy estimation} is always unbiased, which is quite intuitive.}

Next, we analyze the variance of MIPS, which we show is never worse than that of IPS and can be substantially lower.
\begin{theorem} (Variance Reduction of MIPS) \label{thm:variance_reduction}
Under Assumptions~\ref{assumption:common_support},~\ref{assumption:common_embed_support}, and~\ref{assumption:no_direct_effect}, we have
\begin{align*}
    & n \left( \mV_{\calD} [\ips ] - \mV_{\calD} [\mips ] \right)\\
    & = \mE_{p(x)p(e|x,\pi_0)} \left[ \mE_{p(r|x,e)} \left[ r^2 \right] \mV_{\pi_0(a|x,e)} \left[ w(x,a) \right]  \right],
\end{align*}
which is non-negative. Note that the variance reduction is also lower bounded by zero even when Assumption~\ref{assumption:no_direct_effect} is not true. See Appendix~\ref{app:variance_reduction} for the proof.
\end{theorem}

There are two factors that affect the amount of variance reduction. The first factor is the second moment of the reward with respect to $p(r|x,e)$. This term becomes large when, for example, the reward is noisy even after conditioning on the action embedding $e$. The second factor is the variance of $w(x,a)$ with respect to the conditional distribution $\pi_0(a|x,e)$, which becomes large when \textbf{(i)} $w(x,a)$ has a wide range or \textbf{(ii)} there remain large variations in $a$ even after conditioning on action embedding $e$ so that $\pi_0(a|x,e)$ remains stochastic. Therefore, MIPS becomes increasingly favorable compared to IPS for larger action spaces where the variance of $w(x,a)$ becomes larger. Moreover, to obtain a large variance reduction, the action embedding should ideally not be unnecessarily predictive of the actions.

Finally, the next theorem describes the gain in MSE we can obtain from MIPS when Assumption~\ref{assumption:no_direct_effect} is violated.

\begin{theorem} (MSE Gain of MIPS) \label{thm:mse_gain}
Under Assumptions~\ref{assumption:common_support} and~\ref{assumption:common_embed_support}, we have
\begin{align*}
    & n \left(\mathrm{MSE} ( \ipsnoD ) - \mathrm{MSE} ( \mipsnoD ) \right) \\
    & = \mE_{x,a,e \sim \pi_0} \left[ \left(w(x,a)^2 - w(x,e)^2\right) \cdot \mE_{p(r|x,a,e)} [r^2] \right]  \\
    & \quad + 2V(\pi) \biasmips + (1-n) \biasmips^2 .
\end{align*}
See Appendix~\ref{app:mse_gain} for the proof.
\end{theorem}

Note that IPS can have some bias when Assumption~\ref{assumption:common_support} is not true, possibly producing a greater MSE gain for MIPS.

\subsection{Data-Driven Embedding Selection} \label{sec:slope}
The analysis in the previous section implies a clear bias-variance trade-off with respect to the \textit{quality of the action embeddings}. Specifically, Theorem~\ref{thm:bias} suggests that the action embeddings should be as \textit{informative} as possible to reduce the bias when Assumption~\ref{assumption:no_direct_effect} is violated. On the other hand, Theorem~\ref{thm:variance_reduction} suggests that the action embeddings should be as \textit{coarse} as possible to gain a greater variance reduction. Theorem~\ref{thm:mse_gain} summarizes the bias-variance trade-off in terms of MSE. 

A possible criticism to MIPS is Assumption~\ref{assumption:no_direct_effect}, as it is hard to verify whether this assumption is satisfied using only the observed logged data. However, the above discussion about the bias-variance trade-off implies that it might be effective to strategically violate Assumption~\ref{assumption:no_direct_effect} by discarding some embedding dimensions. This \textit{action embedding selection} can lead to a large variance reduction at the cost of introducing some bias, possibly improving the MSE of MIPS. To implement the action embedding selection, we can adapt the estimator selection method called SLOPE proposed in \citet{su2020adaptive} and \citet{tucker2021improved}. SLOPE is based on Lepski’s principle for bandwidth selection in nonparametric statistics~\citep{lepski1997optimal} and is used to tune the hyperparameters of OPE estimators. A benefit of SLOPE is that it avoids estimating the bias of the estimator, which is as difficult as OPE. Appendix~\ref{app:slope} describes how to apply SLOPE to the action embedding selection in our setup, and Section~\ref{sec:empirical} evaluates its benefit empirically.

\subsection{Estimating the Marginal Importance Weights} \label{sec:estimate_iw}
When using MIPS, we might have to estimate $w(x,e)$ depending on how the embeddings are given.
A simple approach to this is to utilize the following transformation.
\begin{align}
    w(x,e) = \mE_{\pi_0(a|x,e)} \left[ w(x,a) \right]. \label{eq:w_x_c}
\end{align}
Eq.~\eqref{eq:w_x_c} implies that we need an estimate of $\pi_0(a|x,e)$, which we compute by regressing $a$ on $(x,e)$.
We can then estimate $w(x,e) $ as $\hat{w}(x,e) = \mE_{\hat{\pi}_0(a|x,e)} \left[ w(x,a) \right]$.\footnote{Appendix~\ref{app:estimated_iw} describes the bias and variance of MIPS with estimated marginal importance weights $\hat{w}(x,e)$.} This procedure is easy to implement and tractable, even when the embedding space is high-dimensional and continuous. Note that, even if there are some deficient actions, we can directly estimate $w(x,e)$ by solving density ratio estimation as binary classification as done in~\citet{sondhi2020balanced}.

\section{Empirical Evaluation} \label{sec:empirical}
We first evaluate MIPS on synthetic data to identify the situations where it enables a more accurate OPE. Second, we validate real-world applicability on data from an online fashion store. Our experiments are conducted using the \textit{OpenBanditPipeline} (OBP)\footnote{\href{https://github.com/st-tech/zr-obp}{\blue{https://github.com/st-tech/zr-obp}}}, an open-source software for OPE provided by~\citet{saito2020open}. Our experiment implementation is available at \href{https://github.com/usaito/icml2022-mips}{\blue{https://github.com/usaito/icml2022-mips}}.

\subsection{Synthetic Data} \label{sec:synthetic}
For the first set of experiments, we create synthetic data to be able to compare the estimates to the ground-truth value of the target policies. To create the data, we sample 10-dimensional context vectors $x$ from the standard normal distribution.
We also sample $d_e$-dimensional categorical action embedding $e \in \calE$ from the following conditional distribution given action $a$.
\begin{align}
    p(e \mid a) = \prod_{k=1}^{d_e} \frac{\exp( \alpha_{a,e_k} )}{ \sum_{e' \in \calE_k} \exp( \alpha_{a,e'}) } , \label{eq:embed_dist}
\end{align} 
which is independent of the context $x$ in the synthetic experiment. $\{\alpha_{a,e_k}\}$ is a set of parameters sampled independently from the standard normal distribution.
Each dimension of $\calE$ has a cardinality of $10$, i.e., $\calE_k = \{1,2,\dots,10\}$.
We then synthesize the expected reward as
\begin{align}
    q(x,e) = \sum_{k = 1}^{d_e} \eta_k \cdot \left( x^{\top} M x_{e_k} + \theta_x^{\top} x +  \theta_e^{\top} x_{e_k}  \right), \label{eq:synthetic_reward}
\end{align}
where $M$, $\theta_x$, and $\theta_e$ are parameter matrices or vectors to define the expected reward. 
These parameters are sampled from a uniform distribution with range $[-1,1]$. 
$x_{e_k}$ is a context vector corresponding to the $k$-th dimension of the action embedding, which is unobserved to the estimators.
$\eta_k$ specifies the importance of the $k$-th dimension of the action embedding, which is sampled from Dirichlet distribution so that $\sum_{k = 1}^{d_e} \eta_k=1$.
Note that if we observe all dimensions of $\calE$, then $q(x,e) = q(x,a,e)$. On the other hand, $q(x,e) \neq q(x,a,e)$, if there are some missing dimensions, which means that Assumption~\ref{assumption:no_direct_effect} is violated.

We synthesize the logging policy $\pi_0$ by applying the softmax function to $q(x,a) = \mE_{p(e|a)}[q(x,e)]$ as
\begin{align}
    \pi_0(a \mid x) =  \frac{\exp( \beta \cdot q(x,a))}{ \sum_{a' \in \calA} \exp( \beta \cdot q(x,a')) } , \label{eq:synthetic_logging}
\end{align}
where $\beta$ is a parameter that controls the optimality and entropy of the logging policy.
A large positive value of $\beta$ leads to a near-deterministic and well-performing logging policy, while lower values make the logging policy increasingly worse. In the main text, we use $\beta=-1$, and additional results for other values of $\beta$ can be found in Appendix~\ref{app:additional_results}.

In contrast, the target policy $\pi$ is defined as 
\begin{align*}
    \pi(a \mid x) = (1 - \epsilon) \cdot \mathbb{I} \big\{a = \argmax_{a' \in \calA} q(x,a') \big\} + \epsilon / |\calA|,
\end{align*}
where the noise $\epsilon \in [0,1]$ controls the quality of $\pi$. In the main text, we set $\epsilon=0.05$, which produces a near-optimal and near-deterministic target policy.  We share additional results for other values of $\epsilon$ in Appendix~\ref{app:additional_results}.

To summarize, we first sample context $x$ and define the expected reward $q(x,e)$ as in Eq.~\eqref{eq:synthetic_reward}. We then sample discrete action $a$ from $\pi_0$ based on Eq.~\eqref{eq:synthetic_logging}. Given action $a$, we sample categorical action embedding $e$ based on Eq.~\eqref{eq:embed_dist}. Finally, we sample the reward from a normal distribution with mean $q(x,e)$ and standard deviation $\sigma=2.5$. Iterating this procedure $n$ times generates logged data $\calD$ with $n$ independent copies of $(x,a,e,r)$.

\begin{figure*}[th]
\centering
\begin{minipage}{\hsize}
    \begin{center}
        \includegraphics[clip, width=15cm]{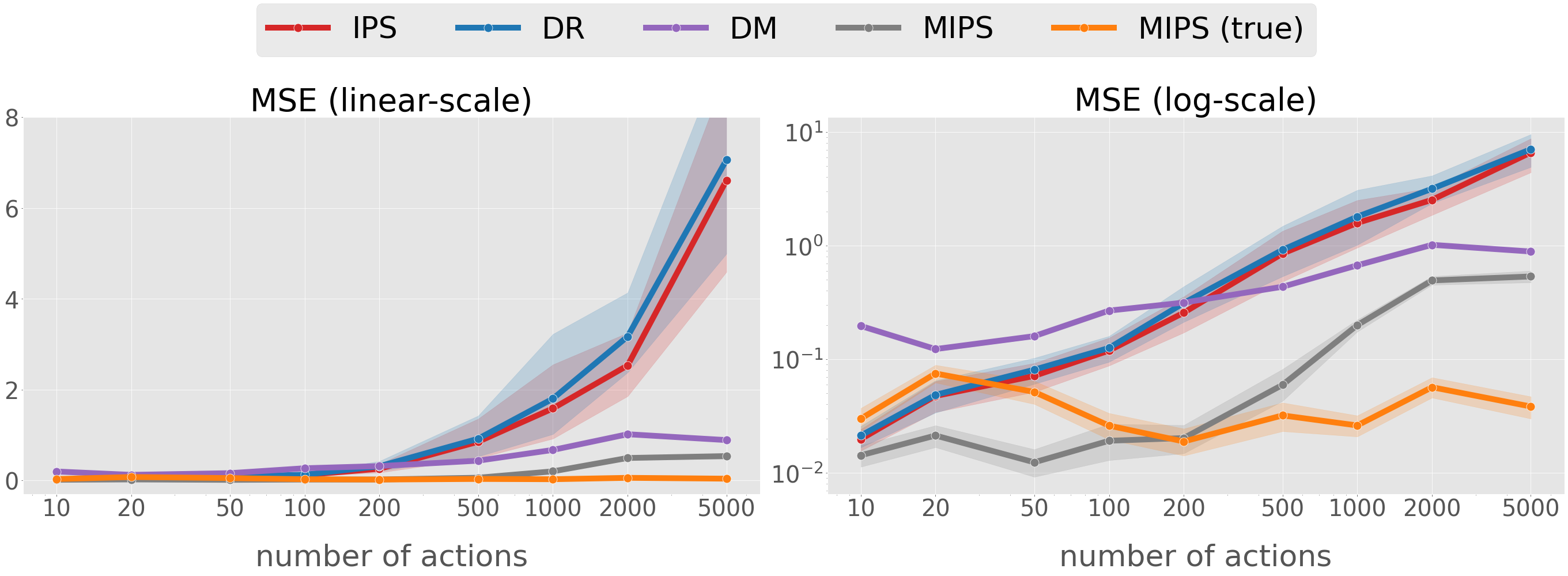}
    \end{center}
    \vspace{-2mm}
    \caption{MSE (both on linear- and log-scale) with \textbf{varying number of actions}.}
    \label{fig:varying_num_actions}
\end{minipage}
\end{figure*}

\begin{figure*}[th]
\centering
\begin{minipage}{\hsize}
    \begin{center}
        \includegraphics[clip, width=15cm]{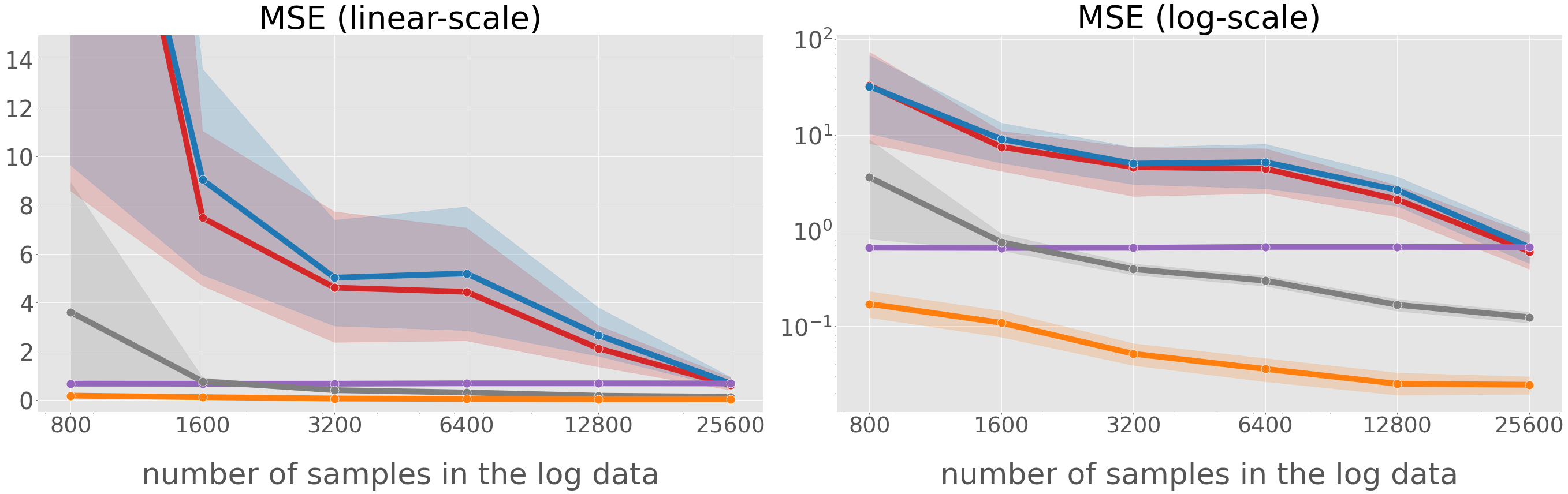}
    \end{center}
    \vspace{-2mm}
    \caption{MSE (both on linear- and log-scale) with \textbf{varying number of samples}.}
    \label{fig:varying_sample_size}
\end{minipage}
\end{figure*}
\subsubsection{Baselines}
We compare our estimator with Direct Method (DM), IPS, and DR.\footnote{Appendix~\ref{app:additional_results} provides more comprehensive experiment results including Switch-DR~\citep{wang2017optimal}, DR with Optimistic Shrinkage (DRos)~\citep{su2020doubly}, and DR-$\lambda$~\citep{metelli2021subgaussian} as additional baseline estimators. The additional experimental results suggest that all of these existing
estimators based on IPS weighting experience significant accuracy deterioration with large action spaces due to either large bias or variance. Moreover, we observe that MIPS is more robust and outperforms all these baselines in a range of settings.} We use the Random Forest~\citep{breiman2001random} implemented in \textit{scikit-learn}~\citep{pedregosa2011scikit} along with 2-fold cross-fitting~\cite{newey2018crossfitting} to obtain $\hat{q}(x,e)$ for DR and DM. We use the Logistic Regression of \textit{scikit-learn} to estimate $\hat{\pi}_{0}(a|x,e)$ for MIPS. We also report the results of MIPS with the true importance weights as ``MIPS (true)". MIPS (true) provides the best performance we could achieve by improving the procedure for estimating the importance weights of MIPS.

\begin{figure*}[th]
\centering
\includegraphics[clip, width=14.5cm]{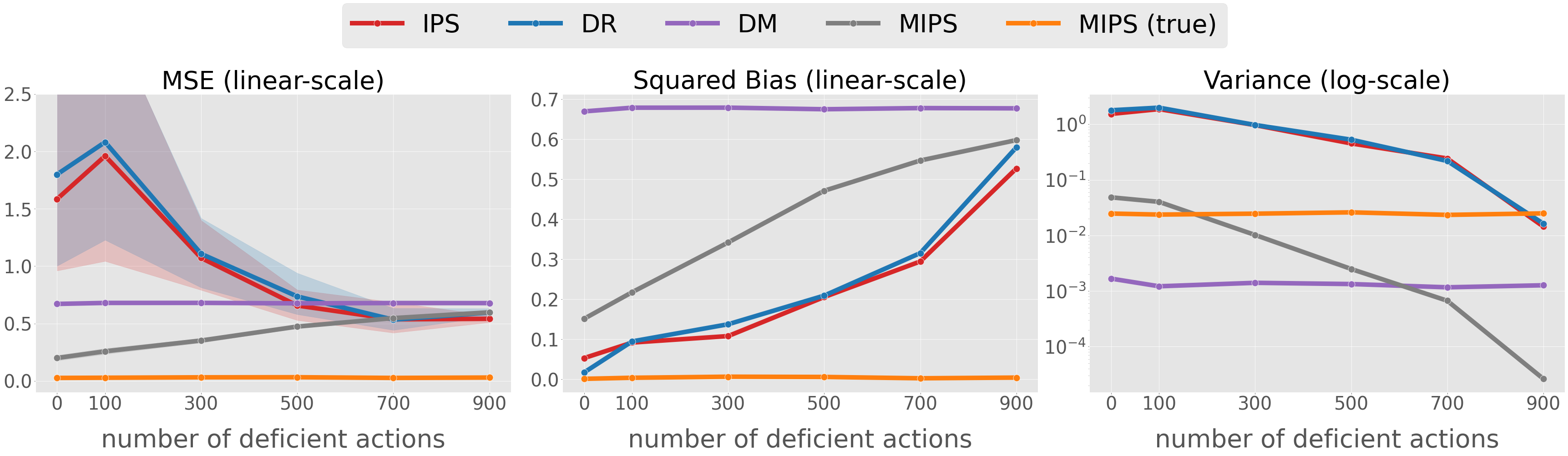}
\vspace{-2mm}
\caption{MSE, Squared Bias, and Variance with \textbf{varying number of deficient actions}.}
\label{fig:varying_deficient_actions}
\end{figure*}

\begin{figure*}[th]
\centering
\begin{minipage}{\hsize}
    \begin{center}
        \includegraphics[clip, width=14.5cm]{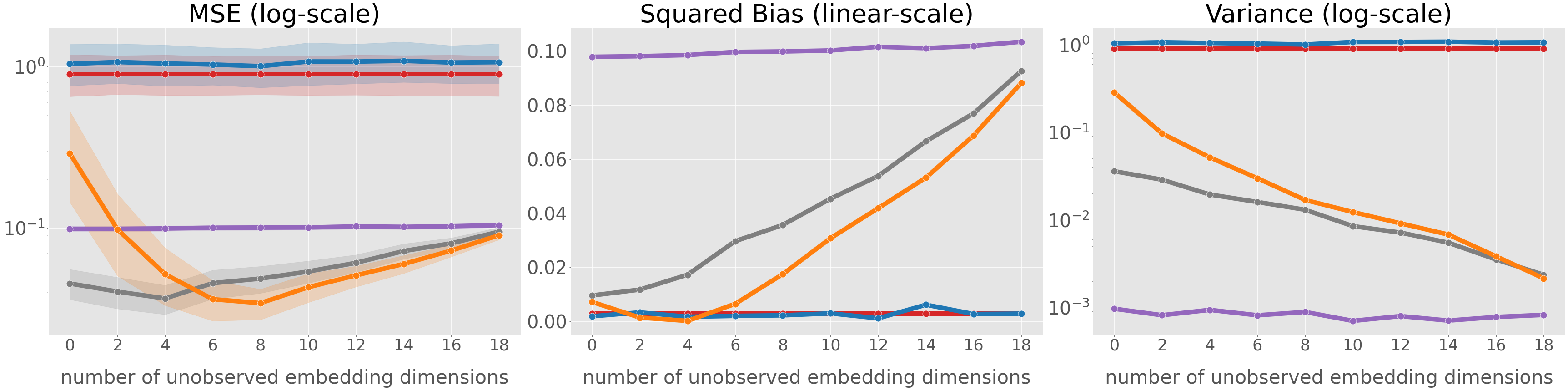}
    \end{center}
    \vspace{-2mm}
    \caption{MSE, Squared Bias, and Variance with \textbf{varying number of unobserved dimensions in action embeddings}.}
    \label{fig:violating_assumptions}
\end{minipage}
\\ \vspace{0.2in}
\begin{minipage}{\hsize}
    \begin{center}
        \includegraphics[clip, width=14.5cm]{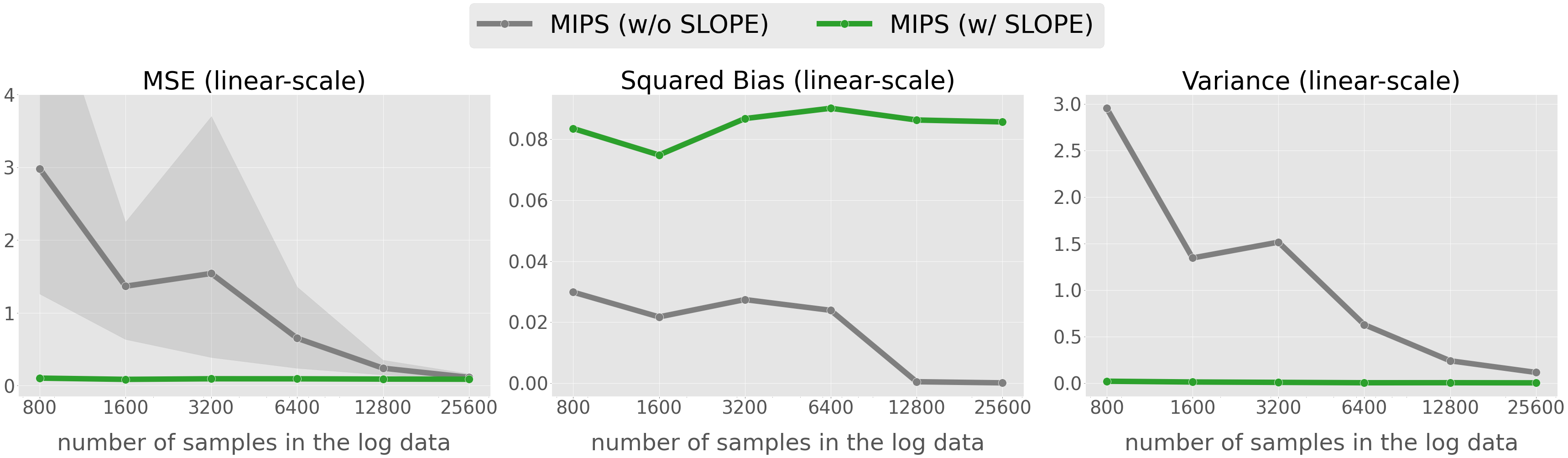}
    \end{center}
    \vspace{-2mm}
    \caption{MSE, Squared Bias, and Variance of MIPS \textbf{w/ or w/o action embedding selection (SLOPE)}.}
    \label{fig:slope}
\end{minipage}
\end{figure*}
\subsubsection{Results}
The following reports and discusses the MSE, squared bias, and variance of the estimators computed over 100 different sets of logged data replicated with different seeds.

\paragraph{How does MIPS perform with varying numbers of actions?}
First, we evaluate the estimators' performance when we vary the number of actions from 10 to 5000. The sample size is fixed at $n=10000$.
Figure~\ref{fig:varying_num_actions} shows how the number of actions affects the estimators' MSE (both on linear- and log-scale). We observe that MIPS provides substantial improvements over IPS and DR particularly for larger action sets. More specifically, when $|\calA|=10$, $\mseratio = 1.38$, while $\mseratio=12.38$ for $|\calA|=5000$, indicating a significant performance improvement of MIPS for larger action spaces as suggested in Theorem~\ref{thm:variance_reduction}. MIPS is also consistently better than DM, which suffers from high bias. The figure also shows that MIPS (true) is even better than MIPS in large action sets, mostly due to the reduced bias when using the true marginal importance weights. This observation implies that there is room for further improvement in how to estimate the marginal importance weights.

\paragraph{How does MIPS perform with varying sample sizes?}
Next, we compare the estimators under varying numbers of samples ($n \in \{800,1600,3200,6400,12800,25600\} $). The number of actions is fixed at $|\calA|=1000$. Figure~\ref{fig:varying_sample_size} reports how the estimators' MSE changes with the size of logged bandit data. We can see that MIPS is appealing in particular for small sample sizes where it outperforms IPS and DR by a larger margin than in large sample regimes ($\mseratio = 9.10$ when $n=800$, while $\mseratio = 4.87$ when $n=25600$). With the growing sample size, MIPS, IPS, and DR improve their MSE as their variance decreases. In contrast, the accuracy of DM does not change across different sample sizes, but it performs better than IPS and DR because they converge very slowly in the presence of many actions. In contrast, MIPS is better than DM except for $n=800$, as the bias of MIPS is much smaller than that of DM. Moreover, MIPS becomes increasingly better than DM with the growing sample size, as the variance of MIPS decreases while DM remains highly biased.

\paragraph{How does MIPS perform with varying numbers of deficient actions?}
We also compare the estimators under varying numbers of deficient actions ($|\calU_0| \in \{0,100,300,500,700,900\}$) with a fixed action set ($|\calA|=1000$). Figure~\ref{fig:varying_deficient_actions} shows how the number of deficient actions affects the estimators' MSE, squared bias, and variance. The results suggest that MIPS (true) is robust and not affected by the existence of deficient actions. In addition, MIPS is mostly better than DM, IPS, and DR even when there are many deficient actions. However, we also observe that the gap between MIPS and MIPS (true) increases for large numbers of deficient actions due to the bias in estimating the marginal importance weights. Note that the MSE of IPS and DR decreases with increasing number of deficient actions, because their variance becomes smaller with a smaller number of supported actions, even though their bias increases as suggested by~\citet{sachdeva2020off}.

\paragraph{How does MIPS perform when Assumption~\ref{assumption:no_direct_effect} is violated?}
Here, we evaluate the accuracy of MIPS when Assumption~\ref{assumption:no_direct_effect} is violated. To adjust the amount of violation, we modify the action embedding space and reduce the cardinality of each dimension of $\calE$ to 2 (i.e., $\calE_k=\{0,1\}$), while we increase the number of dimensions to 20 ($d_e=20$). This leads to $|\calE|=2^{20}=1,048,576$, and we can now drop some dimensions to increase violation.
In particular, when we observe all dimensions of $\calE$, Assumption~\ref{assumption:no_direct_effect} is perfectly satisfied. However, when we withhold $\{0,2,4,\ldots,18\}$ embedding dimensions, the assumption becomes increasingly invalid. When many dimensions are missing, the bias of MIPS is expected to increase as suggested in Theorem~\ref{thm:bias}, potentially leading to a worse MSE.

Figure~\ref{fig:violating_assumptions} shows how the MSE, squared bias, and variance of the estimators change with varying numbers of unobserved embedding dimensions. Somewhat surprisingly, we observe that MIPS and MIPS (true) perform better when there are some missing dimensions, even if it leads to the violated assumption. Specifically, the MSE of MIPS and MIPS (true) is minimized when there are 4 and 8 missing dimensions (out of 20), respectively. This phenomenon is due to the reduced variance. The third column of Figure~\ref{fig:violating_assumptions} implies that the variance of MIPS and MIPS (true) decreases substantially with an increasing number of unobserved dimensions, while the bias increases with the violated assumption as expected. These observations suggest that MIPS can be highly effective despite the violated assumption.

\paragraph{How does data-driven embedding selection perform combined with MIPS?} 
The previous section showed that there is a potential to improve the accuracy of MIPS by selecting a subset of dimensions for estimating the marginal importance weights. We now evaluate whether we can effectively address this embedding selection problem.

Figure~\ref{fig:slope} compares the MSE, squared bias, and variance of MIPS and MIPS with SLOPE (MIPS w/ SLOPE) using the same embedding space as in the previous section. Note that we vary the sample size $n$ and fix $|\calA|=1000$. The results suggest that the data-driven embedding selection provides a substantial improvement in MSE for small sample sizes. As shown in the second and third columns in Figure~\ref{fig:slope}, the embedding selection significantly reduces the variance at the cost of introducing some bias by strategically violating the assumption, which results in a better MSE.

\begin{figure}[t]
\begin{center}
\includegraphics[clip,width=8cm]{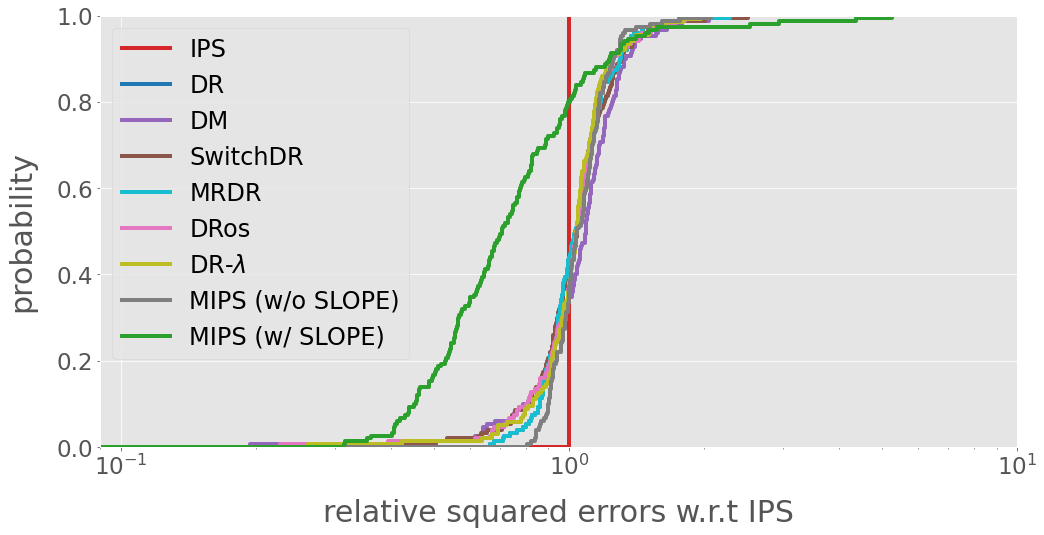}
\end{center}
\vspace{-3mm}
\caption{CDF of relative squared error w.r.t IPS.}
\label{fig:results_real}
\end{figure}

\paragraph{Other benefits of MIPS.}
MIPS has additional benefits over the conventional estimators. In fact, in addition to the case with many actions, IPS is also vulnerable when logging and target policies differ substantially and the reward is noisy (see Eq.~\eqref{eq:ips_variance}). 
Appendix~\ref{app:additional_results} empirically investigates the additional benefits of MIPS with varying logging/target policies and varying noise levels with a fixed action set. We observe that MIPS is substantially more robust to the changes in policies and added noise than IPS or DR, which provides further arguments for the applicability of MIPS.

\subsection{Real-World Data} \label{sec:real_world}
To assess the real-world applicability of MIPS, we now evaluate MIPS on real-world bandit data. In particular, we use the Open Bandit Dataset (OBD)\footnote{https://research.zozo.com/data.html}~\citep{saito2020open}, a publicly available logged bandit dataset collected on a large-scale fashion e-commerce platform. We use 100,000 observations that are randomly sub-sampled from the ``ALL" campaign of OBD. The dataset contains user contexts $x$, fashion items to recommend as action $a \in \calA$ where $|\calA|=240$, and resulting clicks as reward $r \in \{0,1\}$. OBD also includes 4-dimensional action embedding vectors such as hierarchical category information about the fashion items.

The dataset consists of two sets of logged bandit data collected by two different policies (uniform random and Thompson sampling) during an A/B test of these policies. We regard uniform random and Thompson sampling as logging and target policies, respectively, to perform an evaluation of OPE estimators. Appendix~\ref{app:procedure_with_obd} describes the detailed experimental procedure to evaluate the accuracy of the estimators on real-world bandit data.

\paragraph{Results.}
We evaluate MIPS (w/o SLOPE) and MIPS (w/ SLOPE) in comparison to DM, IPS, DR, Switch-DR, More Robust DR~\citep{farajtabar2018more}, DRos, and DR-$\lambda$. We apply SLOPE to tune the built-in hyperparameters of Switch-DR, DRos, and DR-$\lambda$.
Figure~\ref{fig:results_real} compares the estimators by drawing the cumulative distribution function (CDF) of their squared errors estimated with 150 different bootstrapped samples of the logged data. Note that the squared errors are normalized by that of IPS. We find that MIPS (w/ SLOPE) outperforms IPS in about 80\% of the simulation runs, while other estimators, including MIPS (w/o SLOPE), work similarly to IPS. This result demonstrates the real-world applicability of our estimator as well as the importance of implementing action embedding selection in practice. We report qualitatively similar results for other sample sizes (from 10,000 to 500,000) in Appendix~\ref{app:procedure_with_obd}.

\section{Conclusion and Future Work} \label{sec:conclusion}
We explored the problem of OPE for large action spaces. In this setting, existing estimators based on IPS suffer from impractical variance, which limits their applicability. This problem is highly relevant for practical applications, as many real decision making problems such as recommender systems have to deal with a large number of discrete actions. To achieve an accurate OPE for large action spaces, we propose the MIPS estimator, which builds on the \textit{marginal} importance weights computed with \textit{action embeddings}. We characterize the important statistical properties of the proposed estimator and discuss when it is superior to the conventional ones. Extensive experiments demonstrate that MIPS provides a significant gain in MSE when the vanilla importance weights become large due to large action spaces, substantially outperforming IPS and related estimators. 

Our work raises several interesting research questions. For example, this work assumes the existence of some predefined action embeddings and analyzes the resulting statistical properties of MIPS. Even though we discussed how to choose which embedding dimensions to use for OPE (Section~\ref{sec:slope}), it would be intriguing to develop a more principled method to optimize or learn (possibly continuous) action embeddings from the logged data for further improving MIPS. Developing a method for accurately estimating the marginal importance weight would also be crucial to fill the gap between MIPS and MIPS (true) observed in our experiments. It would also be interesting to explore off-policy learning using action embeddings and possible applications of marginal importance weighting to other estimators that depend on the vanilla importance weight such as DR.

\section*{Acknowledgements}
This research was supported in part by NSF Awards IIS-1901168 and IIS-2008139. Yuta Saito was supported by the Funai Overseas Scholarship. All content represents the opinion of the authors, which is not necessarily shared or endorsed by their respective employers and/or sponsors. 

\bibliography{main.bbl}
\bibliographystyle{icml2022}

\newpage
\appendix
\onecolumn

\section{Related Work} \label{app:related}

\paragraph{Off-Policy Evaluation:}
Off-policy evaluation of counterfactual policies has extensively been studied in both contextual bandits~\citep{dudik2014doubly,wang2017optimal,liu2018breaking,farajtabar2018more,su2019cab,su2020doubly,kallus2020optimal,metelli2021subgaussian} and reinforcement learning (RL)~\citep{jiang2016doubly,thomas2016data,xie2019towards,kallus2020double,liu2020understanding}.
There are three main approaches in the literature. The first approach is DM, which estimates the policy value based on the estimated reward $\hat{q}$. DM has a lower variance than IPS, and is also proposed as an approach to deal with support deficient data~\citep{sachdeva2020off} where IPS is biased. A drawback is that it is susceptible to misspecification of the reward function. This misspecification issue is problematic, as the extent of misspecification cannot be easily evaluated for real-world data~\citep{farajtabar2018more,voloshin2019empirical}. The second approach is IPS, which estimates the value of a policy by applying importance weighting to the observed reward. With some assumptions for identification such as common support, IPS is unbiased and consistent. However, IPS can suffer from high bias and variance when the action space is large. It can have a high bias when the logging policy fails to satisfy the common support condition, which is likely to occur for large action spaces~\citep{sachdeva2020off}. Variance is also a critical issue especially when the action space is large, as the importance weights are likely to take larger values. The weight clipping~\cite{swaminathan2015counterfactual,su2019cab,su2020doubly} and normalization~\citep{swaminathan2015self} are often used to address the variance issue, but they produce additional bias. Thus, DR has gained particular attention as the third approach. This estimator is a hybrid of the previous two approaches, and can achieve a lower bias than DM, and a lower variance than IPS~\citep{dudik2014doubly,farajtabar2018more}. It can also achieve the lowest possible asymptotic variance, a property known as \textit{efficiency}~\citep{narita2019efficient}. Several recent works have extended DR to improve its performance with small samples~\citep{wang2017optimal,su2020doubly} or under model misspecification~\citep{farajtabar2018more}. Though there are a number of extensions of DR both in bandits (as described above) and RL~\citep{jiang2016doubly,thomas2016data,kallus2020double}, none of them tackle the large discrete action space. \citet{demirer2019semi} describe an estimator for finitely many possible actions as a special case of their main proposal, which is for continuous action spaces. However, this method is based on a linearity assumption of the reward function, which rarely holds in practice. Moreover, the bias arises from violating the assumption and the variance reduction due to the additional assumption are not analyzed. \citet{kallus2018policy} formulate the problem of OPE for continuous action spaces and propose some estimators building on the kernel smoothing in nonparametric statistics. Specifically, kernel functions are used to infer the rewards among similar continuous actions where the bias-variance trade-off is controlled by a bandwidth hyperparameter. If every dimension of the action embedding space $\calE$ is continuous, the continuous-action estimators of \citet{kallus2018policy} might be applied to our setup under smoothness assumption. However, this naive application can suffer from the curse of dimensionality where the kernel smoothing performs dramatically worse as the number of embedding dimensions increases. In contrast, MIPS avoids the curse of dimensionality by estimating the marginal importance weights via supervised classification as in Section~\ref{sec:estimate_iw}.

Note that there is an estimator called marginalized importance sampling in OPE of RL~\citep{liu2018breaking,xie2019towards,liu2020off}. This method estimates the state marginal distribution and applies importance weighting with respect to this marginal distribution rather than the trajectory distribution. Although marginalization is a key trick of this estimator, it is aimed at resolving the curse of horizon, a problem specific to RL. In contrast, our approach utilizes the marginal distribution over action embeddings to deal with large action spaces. Applications of our estimator are not limited to RL.   

\paragraph{Off-Policy Evaluation for Slate and Ranking Policies:}
Another line of work that shares the similar motivation to ours is OPE of slate or ranking policies~\citep{swaminathan2017off,li2018offline,mcinerney2020counterfactual,saito2020doubly,su2020doubly,vlassis2021control,lopez2021learning,kiyohara2022doubly}. In this setting, the estimators have to handle the combinatorial action space, which could be very large even if the number of unique actions is not. Therefore, some additional assumptions are imposed to make the combinatorial action space tractable. 
A primary problem setting in this direction is OPE for slate bandit policies, where it is assumed that only a single, slate-level reward is observed for each data. \citet{swaminathan2017off} tackle this setting by positing a linearity assumption on the reward function.
The proposed pseudoinverse (PI) estimator was shown to provide an exponential gain in the sample complexity over IPS.
Following this seminal work, \citet{su2020doubly} extend their Doubly Robust with Optimistic Shrinkage, originally proposed for the general OPE problem, to the slate action case. \citet{vlassis2021control} improve the PI estimator by optimizing a set of control variates.
Although PI is compelling, applications of this class of estimators are limited to the specific problem of slate bandits. On the other hand, our framework is more general and applicable not only to slate bandits, but also to other problem instances including OPE for ranking policies with observable slot-level rewards (described below) or general contextual bandits with large action spaces. In addition, all estimators for slate bandits rely on the linearity assumption, while our MIPS builds on a different assumption about the quality of the action embedding.

Another similar setting is OPE for ranking policies where it is assumed that the rewards for every slot in a ranking (slot-level rewards) are observable, a setting also known as semi-bandit feedback. PI and its variants discussed above are applicable to this setting, but \citet{mcinerney2020counterfactual} empirically verify that the PI estimators do not work well, as they do not utilize additional information about the slot-level rewards.
To leverage slot-level rewards to further improve OPE, assumptions are made to capture different types of user behaviors to control the bias-variance trade-off in OPE. For example, \citet{li2018offline} assume that users interact with items presented in different positions of a ranking totally independently. In contrast, \citet{mcinerney2020counterfactual} and \citet{kiyohara2022doubly} assume that users go down a ranking from top to bottom. These assumptions correspond to click models such as cascade model in information retrieval~\citep{guo2009efficient,chuklin2015click} and are useful in reducing the variance. However, whether these assumptions are reasonable depends highly on a ranking interface and real user behavior. If the assumption fails to capture real user behavior, this approach can produce unexpected bias. For example, the cascade model is only applicable when a ranking interface is vertical, however, real-world ranking interfaces are often more complex~\cite{guo2020debiasing}. Moreover, real-world user behaviors are often too diverse to model with a single, universal assumption~\citep{borisov2016neural}. In contrast, our approach is applicable to any ranking interfaces, once they are represented as action embeddings, without assuming any particular user behavior. Moreover, ours is more general in that its application is not limited to information retrieval and recommender systems, but includes robotics, education, conversational agents, or personalized medicine where click models are not applicable.

\paragraph{Reinforcement Learning for Large Action Spaces:}
Although we focus on OPE, there have been several attempts to enable high-performance policy learning for large action spaces.
A typical approach is to factorize the action space into binary sub-spaces~\citep{pazis2011generalized,dulac2012fast}.
For example, \citet{pazis2011generalized} represent each action with a binary format and train a value function for each bit.
On the other hand, \citet{van2009using} and \citet{dulac2015deep} assume the existence of continuous representations of discrete actions as prior knowledge. They perform policy gradients with the continuous actions and search the nearest discrete action. Similar to these works, we assume the existence of some predefined action embeddings and propose to use that prior information to enable an accurate OPE for large action spaces. We also analyze the bias-variance trade-off of the resulting estimator and relate it to the quality of the action embeddings. Some recent works also tackle how to learn useful action representations from only available data. \citet{tennenholtz2019natural} achieve this by leveraging expert demonstrations, while \citet{chandak2019learning} perform supervised learning to predict the state transitions and obtain action representations with no prior knowledge. Following these works, it may be valuable to develop an algorithm to optimize or learn (possibly continuous) action embeddings from the data to further improve OPE for large action spaces.

\paragraph{Multi-Armed Bandits with Side Information:}
There are two prominent approaches to deal with large or infinite action spaces in the \textit{online} bandit literature~\citep{krishnamurthy2019contextual,slivkins2019introduction}.
The first one is the parametric approach such as linear or combinatorial bandits, which assumes that the expected reward can be represented as a parametric function of the action such as a linear function~\citep{chu2011contextual,agrawal2013thompson}.
There is also a nonparametric approach, which typically makes much weaker assumptions about the rewards, e.g., Lipschitz assumptions.
Lipschitz bandits have been studied to address large, structured action spaces such as the $[0, 1]$ interval, where the applications range from dynamic pricing to ad auction. A basic idea in this literature is that similar arms should have similar quality, as per Lipschitz-continuity or some corresponding assumptions on the structure of the action space. The Lipschitz assumption was introduced by \citet{agrawal1995continuum} to the bandit setting. 
\citet{kleinberg2004nearly} optimally solve this problem in the worst case. \citet{kleinberg2019bandits} and \citet{bubeck2011x} rely on the zooming algorithms, which gradually zoom in to the more promising regions of the action space to achieve data-dependent regret bounds. Further works extend this direction by relaxing the assumptions with various local definitions, as well as incorporating contexts into account, as surveyed in Section 4 of \citet{slivkins2019introduction}.

\paragraph{Causal Inference with Surrogates:}
From a statistical standpoint, causal inference with surrogates is also related~\citep{athey2019surrogate,athey2020combining,kallus2020role,chen2021semiparametric}. Its aim is to identify and estimate the causal effect of some treatments (e.g., job training) on a \textit{primary outcome}, which is unobservable without waiting for decades (e.g., lifetime earnings)~\citep{athey2019surrogate}. Instead of waiting for a long period to collect the data, these works assume the availability of \textit{surrogate outcomes} such as test scores and college attendance rates, which could be observed in a much shorter period. In particular, \citet{athey2019surrogate} build on what is called the surrogacy condition to identify the average treatment effect of treatments on the primary outcome. The surrogacy condition is analogous to Assumption~\ref{assumption:no_direct_effect} and states that there should not be any direct effect of treatments on the primary outcome. Although our formulation and assumptions share a similar structure, we would argue that our motivation is to enable an accurate OPE of decision making policies for large action spaces, which is quite different from identifying the average causal effect of binary treatments on a long-term outcome.

\section{Proofs, Derivations, and Additional Analysis} \label{app:proof}

\subsection{Proof of Proposition~\ref{prop:identification}} \label{app:identification}

\begin{proof}
\begin{align}
    \trueV 
    & = \mE_{p(x) \pi(a|x) p(e|x,a) } [q (x, a, e)] \notag \\ 
    & = \mE_{p(x) \pi(a|x) p(e|x,a) } [q (x,e)] \label{eq:a_1_1} \\ 
    & = \mE_{p(x)} \left[\sumA \pi(a|x) \sumE p(e|x,a) \cdot q (x,e) \right] \notag \\ 
    & = \mE_{p(x)} \left[\sumE  q (x,e) \cdot \left( \sumA \pi(a|x) \cdot p(e|x,a) \right) \right] \label{eq:a_1_2} \\ 
    & = \mE_{p(x)} \left[\sumE  p(e|x,\pi) \cdot q (x,e)  \right] \notag \\ 
    & = \mE_{p(x)p(e|x,\pi ) } [ q (x,e) ] \notag \\ 
    & = \mE_{p(x)p(e|x,\pi )p(r|x,e) } [ r ] \notag
\end{align}
where we use Assumption~\ref{assumption:no_direct_effect} in Eq.~\eqref{eq:a_1_1} and $ p(e|x,\pi) = \sumA \pi(a|x) p(e|x,a) $ in Eq.~\eqref{eq:a_1_2}.
\end{proof}

\subsection{Proof of Proposition~\ref{prop:unbiased}} \label{app:unbiased}
\begin{proof}
From the linearity of expectation, we have $\mE_{\calD} [ \mips ] = \mE_{p(x) \pi_0 (a|x) p(e|x,a) p(r|x,a,e)} [w(x,e)r]$. Thus, we calculate only the expectation of $ w(x,e) r $ (RHS of the equation) below.
\begin{align}
     & \mE_{p(x) \pi_0 (a|x) p(e|x,a) p(r|x,a,e)} [w(x,e)r] \notag \\
     & = \mE_{p(x) \pi_0 (a|x) p(e|x,a) } [ w(x,e) \cdot q (x,a,e) ] \notag \\
     & = \mE_{p(x) \pi_0 (a|x) p(e|x,a) } [ w(x,e) \cdot q (x,e) ] \label{eq:a_2_1} \\
     & = \mE_{p(x)} \left[\sumA \pi_0 (a|x) \sumE p(e|x,a)  \frac{p(e|x,\pi)}{p(e|x,\pi_0)} q (x,e) \right] \notag \\
     & = \mE_{p(x)} \left[\sumE \frac{p(e|x,\pi)}{p(e|x,\pi_0)} \cdot q (x,e) \cdot \left( \sumA  p(e|x,a) \cdot \pi_0 (a|x) \right) \right] \notag \\
     & = \mE_{p(x)} \left[\sumE \frac{p(e|x,\pi)}{p(e|x,\pi_0)}  \cdot p(e|x,\pi_0) \cdot q (x,e) \right] \notag \\
     & = \mE_{p(x)p(e|x,\pi)} [ q (x,e)  ]  \label{eq:a_2_2} \\
     & = \mE_{p(x)p(e|x,\pi)p(r|x,e)} [ r  ] \notag \\
     & = \trueV \notag
\end{align}
where we use Assumption~\ref{assumption:no_direct_effect} in Eq.~\eqref{eq:a_2_1} and $ p(e|x,\pi_0) = \sumA \pi_0 (a|x) p(e|x,a) $ in Eq.~\eqref{eq:a_2_2}.
\end{proof}

\subsection{Proof of Theorem~\ref{thm:bias}} \label{app:bias}
To prove Theorem~\ref{thm:bias}, we first state a lemma.
 
\begin{lemma} For real-valued, bounded functions $f: \mathbb{N} \rightarrow \mathbb{R}, g: \mathbb{N} \rightarrow \mathbb{R}, h: \mathbb{N} \rightarrow \mathbb{R}$ where $\sum_{a \in [m]} g(a) = 1 $, we have
\begin{align}
    \sum_{a \in [m]} f(a)g(a) \Big(h(a) - \sum_{b \in [m]} g(b)h(b)\Big)
    = \sum_{a<b\le m} g(a)g(b) (h(a) - h(b)) (f(a) - f(b)) \label{eq:a_5_1}
\end{align}
\begin{proof} We prove this lemma via induction. First, we show the $m=2$ case below.
\begin{align*}
    &f(1)g(1)\left( h(1) - (g(1)h(1) + g(2)h(2) ) \right) + f(2)g(2)\left( h(2) - (g(1)h(1) + g(2)h(2) ) \right)     \\
    &= f(1)g(1)h(1) - f(1)g(1)(g(1)h(1) + g(2)h(2) ) + f(2)g(2) h(2) - f(2)g(2)(g(1)h(1) + g(2)h(2) ) \\
    &= f(1)g(1)h(1) - f(1)g(1)((1 - g(2))h(1) + g(2)h(2) ) + f(2)g(2) h(2) - f(2)g(2)(g(1)h(1) + (1-g(1))h(2) )  \\
    &= - f(1)g(1)( - g(2)h(1) + g(2)h(2) ) - f(2)g(2)(g(1)h(1) -g(1)h(2) )  \\
    &= - f(1)g(1)g(2)(h(2) - h(1)) + f(2)g(1)g(2)(h(2) - h(1))  \\
    &= g(1)g(2) (h(2) - h(1)) (f(2) - f(1) )  \\
\end{align*}
Note that $g(1)+g(2)=1$ from the statement.

Next, we assume Eq.~\eqref{eq:a_5_1} is true for the $m=k-1$ case and show that it is also true for the $m=k$ case.
First, note that
\begin{align*}
    &\sum_{a<b\le k} g(a)g(b) (h(a) - h(b)) (f(a) - f(b)) \\
    &= \sum_{a<b\le k-1} g(a)g(b) (h(a) - h(b))(f(a) - f(b)) + \sum_{a \in [k-1]} g(a)g(k) (h(a)-h(k)) (f(a)-f(k)) \\
\end{align*}

Then, we have
\begin{align}
    & \sum_{a \in [k]} f(a)g(a) \Big(h(a) - \sum_{b \in [k]} g(b)h(b)\Big) \notag \\
    & = \sum_{a\in [k-1]} f(a)g(a) \Big(h(a) - \sum_{b \in [k]} g(b)h(b)\Big) 
    + f(k)g(k) \Big( h(k) - \sum_{b \in [k]} g(b)h(b)\Big) \notag \\
    & = \sum_{a\in [k-1]} f(a)g(a) \left( \Big(h(a) - \sum_{b\in [k-1]} g(b)h(b)\Big) - g(k)h(k) \right)
    + f(k)g(k)  h(k) - f(k)g(k) \sum_{a\in [k]} g(a)h(a) \notag \\
    & = \sum_{a\in [k-1]} f(a)g(a) \Big(h(a) - \sum_{b\in [k-1]} g(b)h(b)\Big) - g(k)h(k) \sum_{a\in [k-1]} f(a)g(a)
    + f(k)g(k)h(k) - f(k)g(k) \sum_{a\in [k]} g(a)h(a) \notag \\
    & = \sum_{a\in [k-1]} f(a)g(a) \Big(h(a) - \sum_{b\in [k-1]} g(b)h(b)\Big) \notag \\
    & \quad  - g(k)h(k) \sum_{a\in [k-1]} f(a)g(a)  + f(k)h(k)g(k) - f(k)g(k) \sum_{a\in [k-1]} g(a)h(a) - f(k)g(k)g(k)h(k) \notag \\
    & = \big(1-g(k) \big) \sum_{a\in [k-1]} f(a) \tilde{g}(a) \left(\big(1-g(k) \big) \Big(h(a) - \sum_{b\in [k-1]} \tilde{g}(b)h(b) \Big) + g(k)h(a) \right) \notag \\
    & \quad  - g(k)h(k) \sum_{a\in [k-1]} f(a)g(a)  + f(k)h(k)g(k) - f(k)g(k) \sum_{a\in [k-1]} g(a)h(a) - f(k)g(k)h(k) \left(1 - \sum_{a\in [k-1]} g(a) \right)  \notag \\
    & = \big(1-g(k) \big)^2 \sum_{a\in [k-1]} f(a) \tilde{g}(a) \Big(h(a) - \sum_{b\in [k-1]} \tilde{g}(b)h(b) \Big) \notag \\
    & \quad + g(k) \sum_{a\in [k-1]} f(a)g(a)h(a) - g(k)h(k) \sum_{a\in [k-1]} f(a)g(a) - f(k)g(k) \sum_{a\in [k-1]} g(a)h(a) + f(k)g(k)h(k) \sum_{a\in [k-1]} g(a) \label{eq:a_5_2}
\end{align}
where we use $ g(k) = 1 - \sum_{a\in [k-1]} g(a)$ and define $\tilde{g}(a) := g(a) / (\sum_{a\in [k-1]} g(a)) = g(a) / (1- g(k)) $.

The first term of Eq.~\eqref{eq:a_5_2} is the $m=k-1$ case, so we have the following from the assumption of induction.
\begin{align*}
    \big(1-g(k) \big)^2 \sum_{a\in [k-1]} f(a) \tilde{g}(a) \Big(h(a) - \sum_{b\in [k-1]} \tilde{g}(b)h(b) \Big) 
    &= \big(1-g(k) \big)^2 \sum_{a<b\le k-1} \tilde{g}(a)\tilde{g}(b) (h(a) - h(b)) (f(a) - f(b)) \\
    &=\sum_{a<b\le k-1} g(a)g(b) (h(a) - h(b)) (f(a) - f(b))
\end{align*}
Note that $\sum_{a\in [k-1]} \tilde{g}(a) = 1$. Rearranging the remaining terms of Eq.~\eqref{eq:a_5_2} yields
\begin{align*}
    & \sum_{a \in [k]} f(a)g(a) \Big(h(a) - \sum_{b \in [k]} g(b)h(b)\Big) \\
    & = \sum_{a<b\le k-1} g(a)g(b) (h(a) - h(b)) (f(a) - f(b)) + \sum_{a \in [k-1]} g(a)g(k)(h(a)-h(k)) (f(a)-f(k))
\end{align*}
Implying that the $m=k$ case is true if the $m=k-1$ case is true.
\end{proof}
\end{lemma}

We then use the above Lemma to prove Theorem~\ref{thm:bias}.
\begin{proof}
\begin{align}
    \biasmips 
    & = \mE_{p(x) \pi_0 (a|x) p(e|x,a) p(r|x,a,e)} [w(x,e)r] - V(\pi) \notag \\
    & = \mE_{p(x) \pi_0 (a|x) p(e|x,a)} [ w(x,e) \cdot q(x,a,e) ] - \mE_{p(x) \pi(a|x) p(e|x,a) } [ q(x,a,e)] \notag \\
    & = \mE_{p(x) \pi_0 (a|x)} \left[\sumE  p(e|x,a) \cdot w(x,e) \cdot q(x,a,e) \right]  
    - \mE_{p(x) \pi(a|x) } \left[\sumE p(e|x,a) \cdot q(x,a,e) \right] \notag \\
    & = \mE_{p(x)} \left[\sumA \pi_0 (a|x) \sumE  \frac{p(e|x,\pi_0) \cdot \pi_0(a|x,e)}{\pi_0(a|x)} \cdot w(x,e) \cdot q(x,a,e) \right] \notag \\
    & \quad- \mE_{p(x)} \left[\sumA \pi(a|x) \sumE  \frac{p(e|x,\pi_0) \cdot \pi_0(a|x,e)}{\pi_0(a|x)} \cdot q(x,a,e) \right] \label{eq:a_3_1} \\
    & = \mE_{p(x)} \left[\sumE p(e|x,\pi_0) \cdot w(x,e) \sumA \pi_0(a|x,e) \cdot q(x,a,e) \right] \notag \\
    & \quad- \mE_{p(x)} \left[\sumE  p(e|x,\pi_0)  \sumA w(x,a) \cdot \pi_0(a|x,e) \cdot q(x,a,e) \right]  \notag \\
    & = \mE_{p(x)p(e|x,\pi_0)} \left[ w(x,e) \sumA \pi_0(a|x,e) \cdot q(x,a,e) \right] \notag \\
    & \quad - \mE_{p(x)p(e|x,\pi_0)} \left[  \sumA w(x,a) \cdot \pi_0(a|x,e) \cdot q(x,a,e) \right] \notag \\
    & = \mE_{p(x)p(e|x,\pi_0)} \left[\sumA w(x,a) \cdot \pi_0(a|x,e)  \sum_{b \in \calA} \pi_0(b|x,e) \cdot q(x,b,c) \right] \notag \\
    & \quad - \mE_{p(x)p(e|x,\pi_0)} \left[  \sumA w(x,a) \cdot \pi_0(a|x,e) \cdot q(x,a,e) \right] \label{eq:a_3_2} \\
    & = \mE_{p(x)p(e|x,\pi_0)} \left[  \sumA w(x,a) \cdot \pi_0(a|x,e) \cdot \left( \Big(  \sum_{b \in \calA} \pi_0(b|x,e) \cdot q(x,b,c) \Big) - q(x,a,e) \right) \right]  \notag
\end{align}
where we use $p(e|x,a) = \frac{p(e|x,\pi_0) \pi_0(a|x,e)}{\pi_0(a|x)} $ in Eq.~\eqref{eq:a_3_1} and $w(x,e) = \mE_{\pi_0(a|x,e) } [w(x,a)] $ in Eq.~\eqref{eq:a_3_2}.
 
By applying Lemma A.1 to the last line (setting $f(a) = w(\cdot,a), g(a) = \pi_0(a | \cdot,\cdot), h(a) = q(\cdot,a,\cdot)$), we get the final expression of the bias. 
\end{proof}

\subsection{Proof of Theorem~\ref{thm:variance_reduction}} \label{app:variance_reduction}
\begin{proof}
Under Assumptions~\ref{assumption:common_support},~\ref{assumption:common_embed_support}, and~\ref{assumption:no_direct_effect}, IPS and MIPS are both unbiased. Thus, the difference in their variance is attributed to the difference in their second moment, which is calculated below.
\begin{align}
    & \mV_{p(x) \pi_0 (a|x) p(e|x,a) p(r|x,a,e)} [w(x,a)r] 
    - \mV_{p(x) \pi_0 (a|x) p(e|x,a) p(r|x,a,e)} [w(x,e)r] \notag \\
    & = \mE_{p(x) \pi_0 (a|x) p(e|x,a) p(r|x,a,e)} [ w(x,a)^2 \cdot r^2 ] 
    - \mE_{p(x) \pi_0 (a|x) p(e|x,a) p(r|x,a,e)} [ w(x,e)^2 \cdot r^2 ] \notag \\
    & = \mE_{p(x) \pi_0 (a|x) p(e|x,a) } \left[ w(x,a)^2 \cdot \mE_{p(r|x,a,e)} [r^2]   \right]  - \mE_{p(x) \pi_0(a|x) p(e|x,a)} \left[ w(x,e)^2 \cdot \mE_{p(r|x,a,e)} [r^2]   \right] \notag \\
    & = \mE_{p(x) \pi_0 (a|x) p(e|x,a) } \left[ \left(w(x,a)^2  - w(x,e)^2 \right) \cdot \mE_{p(r|x,e)} [r^2]  \right] \label{eq:a_4_1} \\  
    & = \mE_{p(x)} \left[\sumA \pi_0 (a| x) \sumE p(e|x,a) \cdot \left(w(x,a)^2  - w(x,e)^2 \right) \cdot \mE_{p(r|x,e)} [r^2]  \right] \notag \\  
    & = \mE_{p(x)} \left[\sumA \pi_0 (a| x) \sumE \frac{p(e|x,\pi_0) \cdot \pi_0(a|x,e)}{\pi_0(a|x)} \cdot \left(w(x,a)^2  - w(x,e)^2 \right) \cdot \mE_{p(r|x,e)} [r^2]  \right] \label{eq:a_4_2} \\ 
    & = \mE_{p(x)} \left[  \sumE p(e|x,\pi_0) \cdot \mE_{p(r|x,e)} [r^2] \sumA \pi_0(a|x,e) \cdot \left(w(x,a)^2  - w(x,e)^2 \right)  \right] \notag \\ 
    & = \mE_{p(x)p(e|x,\pi_0) } \left[  \mE_{p(r|x,e)} [r^2]  \cdot \left( \Big( \sumA \pi_0(a|x,e) \cdot w(x,a)^2 \Big) - w(x,e)^2 \right)  \right] \notag 
\end{align}
where we use Assumption~\ref{assumption:no_direct_effect} in Eq.~\eqref{eq:a_4_1}, $p(e|x,a) = \frac{p(e|x,\pi_0) \pi_0(a|x,e)}{\pi_0(a|x)} $ in Eq.~\eqref{eq:a_4_2}. Here, we have
\begin{align*}
    \left( \sumA \pi_0(a|x,e) \cdot w(x,a)^2 \right) - w(x,e)^2  
    & = \left( \sumA \pi_0(a|x,e) \cdot w(x,a)^2 \right) - \left( \sumA \pi_0(a|x,e) \cdot w(x,a)  \right)^2 \\
    & =  \mE_{\pi_0(a|x,e)} \left[ w(x,a)^2 \right]  - \left( \mE_{\pi_0(a|x,e)} \left[ w(x,a) \right]  \right)^2 \\
    & = \mV_{\pi_0(a|x,e)} \left[ w(x,a) \right]
\end{align*}
where $w(x,e) = \mE_{\pi_0(a|x,e)} [ w(x,a) ] $.

Therefore,
\begin{align*}
   & \mE_{p(x) \pi_0 (a|x) p(e|x,a) p(r|x,a,e)} [ w(x,a)^2 \cdot r^2 ] 
    - \mE_{p(x) \pi_0 (a|x) p(e|x,a) p(r|x,a,e)} [ w(x,e)^2 \cdot r^2 ] \\
   & = \mE_{p(x)p(e|x,\pi_0) } \left[  \mE_{p(r|x,e)} [r^2]  \cdot \left( \Big( \sumA \pi_0(a|x,e) \cdot w(x,a)^2 \Big) - w(x,e)^2 \right)  \right]  \\
   & = \mE_{p(x)p(e|x,\pi_0) } \left[  \mE_{p(r|x,e)} [r^2]  \cdot \mV_{\pi_0(a|x,e)} \left[ w(x,a) \right] \right]  
\end{align*}
Finally, as samples are independent, $n\mV_{\calD} [\ips] = \mV_{p(x) \pi_0 (a|x) p(e|x,a) p(r|x,a,e)} [ w(x,a) r ] $ and $n\mV_{\calD} [\mips] = \mV_{p(x) \pi_0 (a|x) p(e|x,a) p(r|x,a,e)} [w(x,e)r] $ .
\end{proof}

\subsection{Proof of Theorem~\ref{thm:mse_gain}} \label{app:mse_gain}
\begin{proof}
First, we express the MSE gain of MIPS over the vanilla IPS with their bias and variance as follows.
\begin{align*}
    \mathrm{MSE} \big( \ipsnoD \big) - \mathrm{MSE} \big( \mipsnoD \big) 
    = \mV_{\calD} [ \hat{V}_{\mathrm{IPS}} (\pi; \calD) ]  - \mV_{\calD} [ \hat{V}_{\mathrm{MIPS}} (\pi; \calD) ] - \biasmips^2
\end{align*}
Since the samples are assumed to be independent, we can simply rescale the MSE gain as follows.
\begin{align*}
    n\left(\mathrm{MSE} \big( \ipsnoD \big) - \mathrm{MSE} \big( \mipsnoD \big)  \right)
    = \mV_{x,a,r} [ w(x,a) r ]  - \mV_{x,e,r} [w(x,e)r] - n\biasmips^2
\end{align*}

Below, we calculate the difference in variance.
\begin{align*}
    & \mV_{p(x) \pi_0 (a|x) p(e|x,a) p(r|x,a,e)} [ w(x,a) r ]  - \mV_{p(x) \pi_0 (a|x) p(e|x,a) p(r|x,a,e)} [w(x,e)r] \\
    & = \mE_{p(x) \pi_0 (a|x) p(e|x,a) p(r|x,a,e)} [ w(x,a)^2 \cdot r^2 ] - V(\pi)^2 \\
    & \quad -  \left( \mE_{p(x) \pi_0 (a|x) p(e|x,a) p(r|x,a,e)} [ w(x,e)^2 \cdot r^2 ] - \left(V(\pi) + \biasmips \right)^2 \right) \\
    & = \mE_{p(x) \pi_0 (a|x) p(e|x,a)} \left[ \left(w(x,a)^2 - w(x,e)^2\right) \cdot \mE_{p(r|x,a,e)} [r^2] \right] - V(\pi)^2  + \left(V(\pi)^2 + 2 V (\pi) \biasmips + \biasmips^2 \right) \\
    & = \mE_{p(x) \pi_0 (a|x) p(e|x,a)} \left[ \left(w(x,a)^2 - w(x,e)^2\right) \cdot \mE_{p(r|x,a,e)} [r^2] \right]  +  2 V (\pi) \biasmips +  \biasmips^2 
\end{align*}

Thus, we have
\begin{align*}
    &n \left(\mV_{\calD} [ \hat{V}_{\mathrm{IPS}} (\pi; \calD) ]  - \mV_{\calD} [ \hat{V}_{\mathrm{MIPS}} (\pi; \calD) ] - \biasmips^2 \right) \\
    & = \mE_{p(x) \pi_0 (a|x) p(e|x,a)} \left[ \left(w(x,a)^2 - w(x,e)^2\right) \cdot \mE_{p(r|x,a,e)} [r^2] \right] +  2 V (\pi) \biasmips + (1 - n) \biasmips^2 
\end{align*}
\end{proof}
The first term becomes large when the scale of the marginal importance weights is smaller than that of the vanilla importance weights. The second term becomes large when the value of $\pi$ is large and MIPS overestimates it by a large margin. The third term can take a large negative value when the sample size is large and the bias of MIPS is large. This summarizes the bias-variance trade-off between the vanilla IPS and MIPS. When the sample size is small, the first and second terms in the MSE gain are dominant, and MIPS is more appealing due to its variance reduction property. However, as the sample size gets larger, the bias becomes dominant, and IPS is expected to overtake MIPS at some point. We would argue that, when the action space is large, the variance reduction of MIPS often provides the gain in MSE, as the variance components are more dominant, which is supported by our experiment.

\subsection{Derivation of Eq.~\eqref{eq:w_x_c} in Section~\ref{sec:estimate_iw}} \label{app:w_x_c}
\begin{align}
    w(x,e) 
    & = \frac{p(e|x,\pi)}{p(e|x,\pi_0)} \notag \\
    & = \frac{\sumA p(e|x, a)  \cdot \pi(a|x) }{p(e|x,\pi_0)} \notag \\
    & = \frac{p(e|x,\pi_0) \sumA (\pi_0(a|x,e) / \pi_0(a|x)) \cdot \pi(a|x) }{ p(e|x,\pi_0)} \label{eq:a_6_1} \\
    & = \sumA \pi_0(a|x,e)  \frac{\pi(a|x) }{\pi_0(a|x)} \notag \\
    & = \mE_{\pi_0(a|x,e)} \left[ w(x,a) \right] \notag
\end{align}
where we use $ p(e|x, a) = \frac{p(e|x,\pi_0) \pi_0(a|x,e)}{\pi_0(a|x)} $ in Eq.~\eqref{eq:a_6_1}.

\subsection{Bias and Variance of MIPS with Estimated Marginal Importance Weights} \label{app:estimated_iw}

\begin{theorem} (Bias of MIPS with Estimated Marginal Importance Weights)
If Assumption~\ref{assumption:common_embed_support} is true, but Assumption~\ref{assumption:no_direct_effect} is violated, MIPS with the estimated marginal importance weight $\hat{w}(x,e)$ has the following bias.
\begin{align*}
    \mathrm{Bias} ( \hat{V}_{\mathrm{MIPS}} (\pi; \hat{w}) )  
    = \biasmips - \mE_{p(x)p(e|x,\pi)} \left[  \delta(x,e) q (x,\pi_0,e) \right] ,
\end{align*}
where $\hat{V}_{\mathrm{MIPS}} (\pi; \hat{w}) := n^{-1} \sum_{i=1}^n \hat{w}(x_i,e_i) r_i$, $\delta(x,e)  := 1 - (\hat{w}(x,e) / w(x,e)) $, and $q(x,\pi_0,e)  :=  \sumA \pi_0(a|x,e) \cdot q(x,a,e) $.
\begin{proof}
\begin{align}
    \mathrm{Bias} ( \hat{V}_{\mathrm{MIPS}} (\pi; \hat{w}) ) 
    & = \mE_{p(x) \pi_0 (a|x) p(e|x,a) p(r|x,a,e)} [ \hat{w}(x,e) r ] -  V(\pi)     \label{eq:a_7_1} \\
    & =   \mE_{p(x) \pi_0 (a|x) p(e|x,a) p(r|x,a,e)} \big[ (\hat{w}(x,e) - w(x,e)) \cdot r \big] + \biasmips  \label{eq:a_7_2} 
\end{align}
where we use $ \mE_{\calD} [ \hat{V}_{\mathrm{MIPS}} (\pi; \calD, \hat{w}) ] =  \mE_{p(x) \pi_0 (a|x) p(e|x,a) p(r|x,a,e)} [ \hat{w}(x,e) r ] $ (as samples are assumed to be independent) in Eq.~\eqref{eq:a_7_1} and decompose the bias into the bias of MIPS with the true $w(x,e)$ and bias due to the estimation error of $\hat{w}(x,e)$ in Eq.~\eqref{eq:a_7_2}. We know the bias of MIPS with the true weight from Theorem~\ref{thm:bias}, so we calculate only the bias due to estimating the weight.
\begin{align}
    &\mE_{p(x) \pi_0 (a|x) p(e|x,a) p(r|x,a,e)} [ \left(\hat{w}(x,e) - w(x,e) \right) \cdot r ] \notag \\
    & = \mE_{p(x) \pi_0 (a|x) p(e|x,a)} [ \left(\hat{w}(x,e) - w(x,e) \right) \cdot q(x,a,e) ] \notag \\
    & = \mE_{p(x)} \left[\sumA \pi_0 (a|x) \sumE p(e|x,a) \cdot \left(\hat{w}(x,e) - w(x,e) \right) \cdot q(x,a,e)  \right] \notag \\
    & = \mE_{p(x)} \left[\sumA \pi_0 (a|x) \sumE \frac{p(e|x,\pi_0) \cdot \pi_0(a|x,e)}{\pi_0(a|x)} \cdot \left(\hat{w}(x,e) - w(x,e) \right) \cdot q(x,a,e)  \right] \label{eq:a_7_3} \\
    & = \mE_{p(x)} \left[\sumE p(e|x,\pi_0) \cdot \left(\hat{w}(x,e) - w(x,e) \right) \sumA \pi_0(a|x,e) \cdot q(x,a,e)  \right] \notag \\
    & = - \mE_{p(x)} \left[\sumE p(e|x,\pi) \cdot \delta(x,e) \cdot q(x,\pi_0,e)  \right] \label{eq:a_7_4} \\
    & = - \mE_{p(x)p(e|x,\pi)} \left[ \delta(x,e) \cdot q(x,\pi_0,e)  \right] \notag
\end{align}
where we use $ p(e|x, a) = \frac{p(e|x,\pi_0) \pi_0(a|x,e)}{\pi_0(a|x)} $ in Eq.~\eqref{eq:a_7_3} and  $q(x,\pi_0,e) = \sumA \pi_0(a|x,e) q(x,a,e) $ in Eq.~\eqref{eq:a_7_4}.
\end{proof}
\end{theorem}

\begin{theorem} (Variance of MIPS with Estimated Marginal Importance Weights)
Under Assumptions~\ref{assumption:common_embed_support} and~\ref{assumption:no_direct_effect}, we have
\begin{align*}
    n \mV_{\calD} ( \hat{V}_{\mathrm{MIPS}} (\pi; \calD, \hat{w})  ) 
    & = \mE_{p(x) p(e|x,\pi)} \left[ (1-\delta(x,e))^2 w(x,e) \sigma^2 (x,\pi_0,e) \right] \\
    & \quad + \mE_{p(x)} \left[ \mV_{\pi_0 (a|x) p(e|x,a)} \left[ \hat{w}(x,e) q (x,a,e) \right] \right] \\
    & \quad + \mV_{p(x)} \left[ \mE_{p(e|x,\pi)} \left[ (1-\delta(x,e)) q (x,\pi_0,e) \right] \right] 
\end{align*}
where $\delta(x,e)  := 1 - (\hat{w}(x,e) / w(x,e)) $, $q (x,\pi_0,e) := \sumA \pi_0(a|x,e) \cdot q(x,a,e) $, and $\sigma^2 (x,\pi_0,e) := \sumA \pi_0(a|x,e) \cdot \sigma^2(x,a,e) $.
\begin{proof}
Since the samples are assumed to be independent, we have $$ n \mV_{\calD} ( \hat{V}_{\mathrm{MIPS}} (\pi; \calD, \hat{w})  )  =  \mV_{p(x) \pi_0 (a|x) p(e|x,a) p(r|x,a,e)} \left[ \hat{w}(x,e) r  \right].$$
Below we apply the law of total variance twice to the RHS of the above equation.
\begin{align}
    \mV_{p(x) \pi_0 (a|x) p(e|x,a) p(r|x,a,e)} \left[ \hat{w}(x,e) r  \right]
    & = \mE_{p(x) \pi_0 (a|x) p(e|x,a)} \left[ \hat{w}(x,e)^2 \cdot \mV_{p(r|x,a,e)} [r] \right] \notag \\
    & \quad + \mV_{p(x) \pi_0 (a|x) p(e|x,a)} \left[ \hat{w}(x,e) \cdot \mE_{p(r|x,a,e)} [r] \right] \notag \\
    & = \mE_{p(x) \pi_0 (a|x) p(e|x,a)} \left[ \hat{w}(x,e)^2 \cdot \sigma^2 (x,a,e) \right] \notag \\
    & \quad + \mV_{p(x) \pi_0 (a|x) p(e|x,a)} \left[ \hat{w}(x,e) \cdot q (x,a,e) \right] \notag \\
    & = \mE_{p(x) \pi_0 (a|x) p(e|x,a)} \left[ \hat{w}(x,e)^2 \cdot \sigma^2 (x,a,e) \right] \notag \\
    & \quad + \mE_{p(x)} \left[ \mV_{\pi_0 (a|x) p(e|x,a)} \left[ \hat{w}(x,e) \cdot q (x,a,e) \right] \right] \notag \\
    & \quad + \mV_{p(x)} \left[ \mE_{\pi_0 (a|x) p(e|x,a)} \left[ \hat{w}(x,e) \cdot q (x,a,e) \right] \right] \notag \\
    & = \mE_{p(x) p(e|x,\pi)} \left[ (1-\delta(x,e))^2 \cdot w(x,e) \cdot \sigma^2 (x,\pi_0,e) \right] \notag \\
    & \quad + \mE_{p(x)} \left[ \mV_{\pi_0 (a|x) p(e|x,a)} \left[ \hat{w}(x,e) \cdot q (x,a,e) \right] \right] \notag \\
    & \quad + \mV_{p(x)} \left[ \mE_{p(e|x,\pi)} \left[ (1-\delta(x,e)) \cdot q (x,\pi_0,e) \right] \right] \label{eq:a_7_5}
\end{align}
where we use $ \mE_{\pi_0 (a|x) p(e|x,a)} [ \hat{w}(x,e)^2  \sigma^2 (x,a,e)] = \mE_{p(e|x,\pi)} [ (1-\delta(x,e))^2 w(x,e) \sigma^2 (x,\pi_0,e)] $ and $ \mE_{\pi_0 (a|x) p(e|x,a)} [ \hat{w}(x,e) q (x,a,e) ] = \mE_{p(e|x,\pi)} [ (1-\delta(x,e)) q (x,\pi_0,e) ] $ in Eq.~\eqref{eq:a_7_5}
\end{proof}
\end{theorem}

\subsection{Bias of MIPS with Deficient Embedding Support} \label{app:deficient_embed_support}

\begin{theorem} (Bias of MIPS with Deficient Embedding Support)
If Assumption~\ref{assumption:no_direct_effect} is true, but Assumption~\ref{assumption:common_embed_support} is violated, MIPS has the following bias.
\begin{align*}
    \big|\mathrm{Bias} (\mipsnoD) \big|
    = \mE_{p(x)} \left[ \sum_{e \in \calU_0^e(x,\pi_0)} p(e | x,\pi) q (x,e) \right],
\end{align*}
where $\calU_0^e(x,\pi_0) := \{e \in \calE \mid p (e | x,\pi_0) = 0\} $ is the space of unsupported embeddings for context $x$ under $\pi_0$.
\begin{proof}
We follow Proposition 1 of~\citet{sachdeva2020off} to derive the bias under deficient embedding support.
\begin{align}
    \mathrm{Bias} (\mipsnoD) 
    &= \mE_{p(x)\pi_0(a|x)p(e|x,a)p(r|x,a,e)} [w(x,e) r ] - V(\pi) \notag \\
    & = \mE_{p(x)} \left[ \sum_{e\in (\calU_0^e(x,\pi_0))^c} w(x,e) q(x,e) \sumA \pi_0(a|x)p(e|x,a) \right] - \mE_{p(x)p(e|x,\pi)} [q(x,e)] \label{eq:a_8_1} \\
    & =  \mE_{p(x)} \left[ \sum_{e\in (\calU_0^e(x,\pi_0))^c} p(e|x,\pi) q(x,e) - \sumE p(e|x,\pi) q(x,e) \right] \label{eq:a_8_2} \\
    & = - \mE_{p(x)} \left[ \sum_{e\in \calU_0^e(x,\pi_0)} p(e|x,\pi) q(x,e) \right] \notag
\end{align}
where Eq.~\eqref{eq:a_8_1} is due to Assumption~\ref{assumption:no_direct_effect} and Eq.~\eqref{eq:a_8_2} is from $ p(e|x, a) = \frac{p(e|x,\pi_0) \pi_0(a|x,e)}{\pi_0(a|x)} $.
\end{proof}
\end{theorem}

\section{Data-Driven Action Feature Selection Based on \citet{tucker2021improved} and \citet{su2020adaptive}} \label{app:slope}
\citet{wang2017optimal} and \citet{su2020doubly} describe a procedure for data-driven estimator selection, which is used to tune the built-in hyperparameters of their own estimators. However, their methods need to estimate the bias (or its loose upper bound as a proxy) of the estimator as a subroutine, which is as difficult as OPE itself.
\citet{su2020adaptive} develop a generic data-driven method for estimator selection for OPE called SLOPE, which is based on Lepski’s principle~\citep{lepski1997optimal} and does not need a bias estimator.
\citet{tucker2021improved} improve the theoretical analysis of \citet{su2020adaptive}, resulting in a refined procedure called SLOPE++.

Given a finite set of estimators $\{\hat{V}_m\}_{m=1}^M$, which is often constructed by varying the value of hyperparameters, the estimator selection problem aims at identifying the estimator that minimizes some notion of estimation error such as the following absolute error with respect to a given target policy $\pi$.
\begin{align*}
    m^* := \argmin_{m \in [M]} \, \left| \trueV - \hat{V}_m (\pi; \calD) \right|,
\end{align*}
where $\calD$ is a given logged bandit dataset.

For solving this selection problem, SLOPE++ requires the following monotonicity assumption (SLOPE requires a slightly stronger assumption).
\begin{assumption} (Monotonicity) \label{assumption:monotonicity}
\begin{enumerate}
    \item $\mathrm{Bias} (\hat{V}_m) \le \mathrm{Bias} (\hat{V}_{m+1}), \; \forall m \in [M] $
    \item $\mathrm{CNF} (\hat{V}_{m+1}) \le \mathrm{CNF} (\hat{V}_{m}), \; \forall m \in [M] $
\end{enumerate}
where $\mathrm{CNF} (\hat{V}) $ is a high probability bound on the deviation of $\hat{V}$, which requires that the following holds with a probability at least $1 - \delta$.
\begin{align*}
    \left| \mE_{\calD} \Big[ \hat{V} (\pi; \calD) \Big] - \hat{V} (\pi; \calD) \right|   \le \mathrm{CNF} (\hat{V}),
\end{align*}
which we can generally bound with high confidence using techniques such as concentration inequalities.
\end{assumption}

Based on this assumption, \citet{tucker2021improved} derive the following universal bound.

\begin{theorem} (Theorem 1 of \citet{tucker2021improved})
Given $\delta>0$, high confidence bound $\mathrm{CNF} (\hat{V}_{m})$ on the deviations, and that we have ordered the candidate estimators such that $\mathrm{CNF} (\hat{V}_{m+1}) \le \mathrm{CNF} (\hat{V}_{m})$. Selecting the estimator as
\begin{align}
    \hat{m}:=\max \left\{m:\big|\hat{V}_{m}-\hat{V}_{j}\big| \leq \mathrm{CNF}(m)+(\sqrt{6}-1) \mathrm{CNF}(j), j<m\right\}
    \label{eq:slope}
\end{align}
ensures that with probability at least $1 - \delta$, 
\begin{align*}
    \left| \hat{V}_{\hat{m}}  - \hat{V}_{m^*}  \right| \le (\sqrt{6}+3) \min_{m} \left(\max _{j \leq m} \mathrm{Bias}(j)+\mathrm{CNF}(m)\right).
\end{align*}
Under Assumption~\ref{assumption:monotonicity}, the bound simplifies to 
\begin{align*}
    \left| \hat{V}_{\hat{m}}  - \hat{V}_{m^*}  \right| \le (\sqrt{6}+3) \min_{m} \left(\mathrm{Bias}(m)+\mathrm{CNF}(m)\right).
\end{align*}
In contrast, when the set of estimators is not ordered with respect to $\mathrm{CNF}(\cdot)$, we have a looser bound as below.
\begin{align*}
    \left| \hat{V}_{\hat{m}}  - \hat{V}_{m^*}  \right| \le (\sqrt{6}+3) \min_{m} \left(\max _{j \leq m} \mathrm{Bias}(j)+\max _{k \leq m}\mathrm{CNF}(k)\right).
\end{align*}
Note that \citet{tucker2021improved} also provide the corresponding universal upper bound with respect to MSE in their Corollary 1.1.
\end{theorem}

We build on the selection procedure given in Eq.~\eqref{eq:slope} to implement data-driven action feature selection.
Specifically, in our case, the task is to identify which dimensions of the action embedding $e$ we should use to minimize the MSE of the resulting MIPS as follows.
\begin{align*}
    \min_{\calE \subseteq \mathcal{V}} \; \mathrm{Bias} \big(\hat{V}_{\mathrm{MIPS}}\left(\pi; \calE\right)\big)^{2}+\mathbb{V}_{\calD} \big[\hat{V}_{\mathrm{MIPS}} \big(\pi ; \calD, \calE \big) \big]
\end{align*}
where $\mathcal{V}:=\left\{\calE_{1}, \calE_{2}, \ldots, \calE_{k}\right\}$ is a set of available action features. Note that we make the dependence of MIPS on the action embedding space $\calE$ explicit in the above formulation.

As described in Theorems~\ref{thm:bias},~\ref{thm:variance_reduction}, and~\ref{thm:mse_gain}, we should use as many dimensions as possible to reduce the bias, while we should use as coarse information as possible to gain a large variance reduction.
For identifying useful features to compute the marginal importance weights, we construct a set of estimators $\{\hat{V}_{\mathrm{MIPS}}\left(\pi ; \calD, \calE\right)\}_{\calE \subseteq \mathcal{V}}$ and simply apply Eq.~\eqref{eq:slope}. Note that when the number of embedding dimensions is not small, the brute-force search over all possible combinations of the embedding dimensions is not tractable. Thus, we sometime define the action embedding search space $\mathcal{V}$ via a greedy procedure to make the embedding selection tractable. In our experiments, we perform action embedding selection based on the greedy version of SLOPE++, and we estimate a high probability bound on the deviation ($\mathrm{CNF}(\hat{V})$) based on the Student’s $t$ distribution as done in~\citet{thomas2015confidence}. The MIPS estimator along with the exact and greedy versions of embedding dimension selection is now implemented in the OBP package.\footnote{\href{https://github.com/st-tech/zr-obp}{\blue{https://github.com/st-tech/zr-obp}}}.

\section{Experiment Details and Additional Results} \label{app:experiment_detail}

\subsection{Baseline Estimators} \label{app:baselines}
Below, we define and describe the baseline estimators in detail.

\paragraph{Direct Method (DM)}

DM is defined as follows.
\begin{align*}
    \dm := \frac{1}{n} \sum_{i=1}^n \mE_{\pi(a|x_i)} [\hat{q} (x_i, a)] = \frac{1}{n} \sum_{i=1}^n \sumA \pi(a|x_i) \hat{q} (x_i, a), 
\end{align*}
where $\hat{q}(x,a)$ estimates $q(x,a)$ based on logged bandit data.
The accuracy of DM depends on the quality of $\hat{q}(x,a)$.
If $\hat{q}(x,a)$ is accurate, so is DM. 
However, if $\hat{q}(x,a)$ fails to estimate the expected reward accurately, the final estimator is no longer consistent.
As discussed in Appendix~\ref{app:related}, the misspecification issue is challenging, as it cannot be easily detected from available data~\citep{farajtabar2018more,voloshin2019empirical}.
This is why DM is often described as a high bias estimator.

\paragraph{Doubly Robust (DR)~\citep{dudik2014doubly}}

DR is defined as follows.
\begin{align*}
    \dr := \frac{1}{n} \sum_{i=1}^n \left\{ \mE_{\pi(a|x_i)} [\hat{q} (x_i, a)] + w(x_i,a_i)  (r_i-\hat{q}(x_i, a_i) ) \right\},
\end{align*}
which combines DM and IPS in a way to reduce the variance.
More specifically, DR utilizes $\hat{q}$ as a control variate.
If the expected reward is correctly specified, DR is \textit{semiparametric efficient} meaning that it achieves the minimum possible asymptotic variance among regular estimators~\citep{narita2019efficient}.
A problem is that, if the expected reward is misspecified, this estimator can have a larger asymptotic MSE compared to IPS.

\paragraph{Switch Doubly Robust (Switch-DR)~\citep{wang2017optimal}}
Although DR generally reduces the variance of IPS and is also minimax optimal~\citep{wang2017optimal}, it can still suffer from the variance issue in practice, particularly when the importance weights are large due to a weak overlap between target and logging policies.
Switch-DR is introduced to further deal with the variance issue and is defined as follows.
\begin{align*}
    \switchdr := \frac{1}{n} \sum_{i=1}^n \left\{\mE_{\pi(a|x_i)} [\hat{q} (x_i, a)] + w(x_i,a_i) \mathbb{I}\{ w(x_i,a_i) \le \lambda \} (r_i-\hat{q}(x_i, a_i) ) \right\},
\end{align*}
where $\mathbb{I} \{\cdot\}$ is the indicator function and $\lambda \ge 0$ is a hyperparameter.
When $\lambda=0$, Switch-DR becomes DM, while $ \lambda \to \infty $ leads to DR.
Switch-DR is also minimax optimal when $\lambda$ is appropriately set~\citep{wang2017optimal}.

\paragraph{More Robust Doubly Robust~\citep{farajtabar2018more}}

MRDR uses an expected reward estimator ($\hat{q}_{\mathrm{MRDR}}$) derived by minimizing the variance of the resulting DR estimator.
This estimator is defined as $\mrdr := \hat{V}_{\mathrm{DR}} (\pi; \calD, \hat{q}_{\mathrm{MRDR}}),$ where $\hat{q}_{\mathrm{MRDR}}$ is derived by minimizing the (empirical) variance objective: $\hat{q}_{\mathrm{MRDR}} \in \argmin_{\hat{q} \in \mathcal{Q}} \mV_{n} (\dr),$
where $\mathcal{Q}$ is a function class for $\hat{q}$. When $\mathcal{Q}$ is well-specified, then $\hat{q}_{\mathrm{MRDR}} = q$. 
The main point is that, even if $\mathcal{Q}$ is misspecified, MRDR is still expected to perform reasonably well, as the target function is the resulting variance. To implement MRDR, we follow \citet{farajtabar2018more} and \citet{su2020doubly}, and derive $\hat{q}_{\mathrm{MRDR}}$ by minimizing the weighted squared loss with respect to the reward prediction on the logged data.

\paragraph{Doubly Robust with Optimistic Shrinkage~\citep{su2020doubly}}
DRos is defined via minimizing an upper bound of the MSE and is defined as follows.
\begin{align*}
    \drs := \frac{1}{n} \sum_{i=1}^n \left\{\mE_{\pi(a|x_i)} [\hat{q} (x_i, a)] + \frac{\lambda w(x_i, a_i)}{w(x_i, a_i)^2+\lambda}  (r_i-\hat{q}(x_i, a_i) )\right\},
\end{align*}
where $\lambda \ge 0$ is a hyperparameter. When $\lambda = 0$, DRos is equal to DM, while $\lambda \rightarrow \infty$ makes DRos identical to DR.
DRos is aimed at improving the small sample performance of DR, but is indeed biased due to the weight shrinkage.

\paragraph{DR-$\lambda$~\citep{metelli2021subgaussian}}
DR-$\lambda$ is a recent estimator building on a ``smooth shrinkage" of the importance weights to mitigate the heavy-tailed behavior of the previous estimators. This estimator is defined as follows.
\begin{align*}
    \drlam & := \frac{1}{n} \sum_{i=1}^n \left\{\mE_{\pi(a|x_i)} [\hat{q} (x_i, a)] + \frac{w(x_i, a_i)}{1 - \lambda + \lambda w(x_i, a_i)}  (r_i-\hat{q}(x_i, a_i) )\right\},
\end{align*}

where $\lambda \in [0,1]$ is a hyperparameter. Note that \citet{metelli2021subgaussian} define a more general weight, $ ((1-\lambda) w(x,a)^s + \lambda ) ^{\frac{1}{s}}$, with an additional hyperparameter $s$. The above instance is a special case with $s=1$, which is the main proposal of \citet{metelli2021subgaussian}.

\begin{figure*}[h]
\centering
\begin{minipage}{\hsize}
    \begin{center}
        \includegraphics[clip, width=16.5cm]{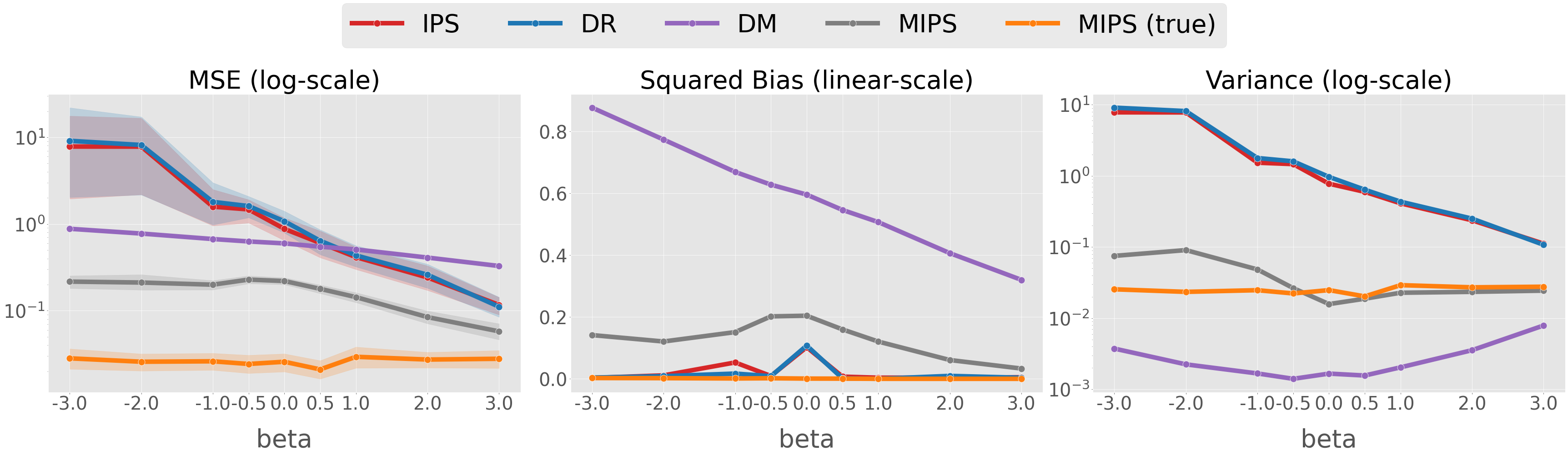}
    \end{center}
    \vspace{-2mm}
    \caption{MSE, Squared Bias, and Variance with \textbf{varying logging policies} ($\beta$)}
    \label{fig:varying_beta}
\end{minipage}
\\ \vspace{0.4in}
\begin{minipage}{\hsize}
    \begin{center}
        \includegraphics[clip, width=16.5cm]{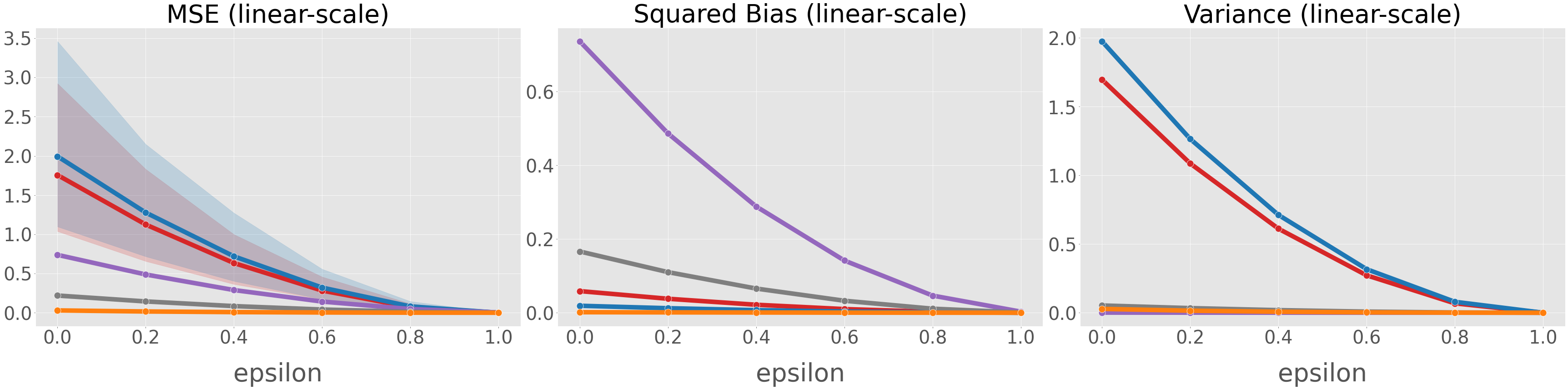}
    \end{center}
    \vspace{-2mm}
    \caption{MSE, Squared Bias, and Variance with \textbf{varying target policies} ($\epsilon$)}
    \label{fig:varying_eps}
\end{minipage}
\\ \vspace{0.4in}
\begin{minipage}{\hsize}
    \begin{center}
        \includegraphics[clip, width=16.5cm]{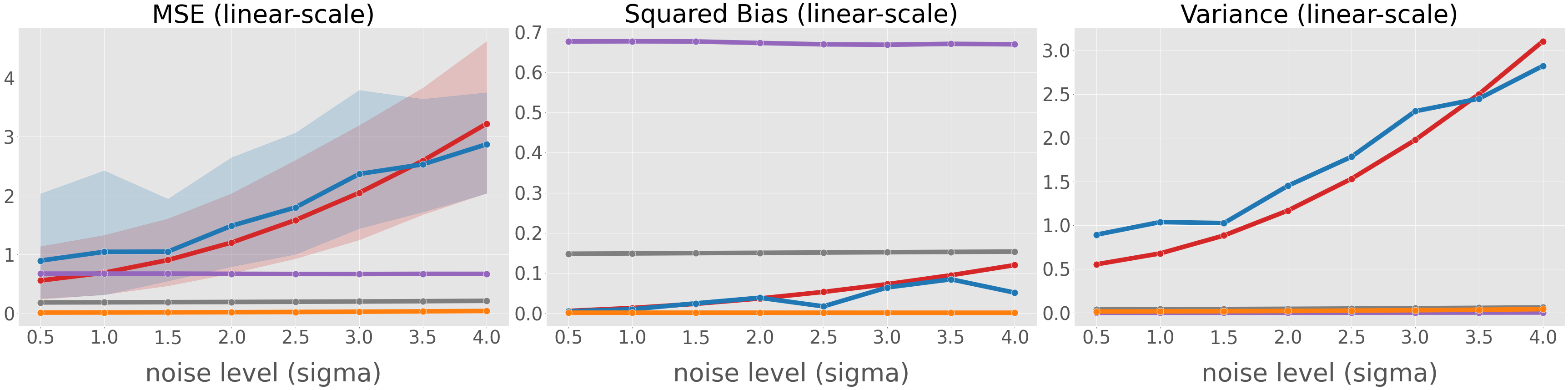}
    \end{center}
    \vspace{-2mm}
    \caption{MSE, Squared Bias, and Variance with \textbf{varying noise levels} ($\sigma$)}
    \label{fig:varying_noise}
\end{minipage}
\vskip 0.1in
\raggedright
\fontsize{9pt}{9pt}\selectfont \textit{Note}:
We set $n=10,000$ and $|\calA|=1,000$. 
For Figure~\ref{fig:varying_beta}, we fix $\epsilon=0.05, \sigma=2.5$, for Figure~\ref{fig:varying_eps}, we fix $\beta=-1, \sigma=2.5$, and for Figure~\ref{fig:varying_noise}, we fix $\epsilon=0.05, \beta=-1$.
The results are averaged over 100 different sets of synthetic logged data replicated with different random seeds.
The shaded regions in the MSE plots represent the 95\% confidence intervals estimated with bootstrap.
\end{figure*}
\subsection{Additional Results on Synthetic Bandit Data} \label{app:additional_results}
In this section, we explore two additional research questions regarding the estimators' performance for different logging/target policies and different levels of noise on the rewards. We demonstrate that MIPS works particularly better than other baselines when the target and logging policies differ greatly and the reward is noisy. After discussing the two research questions, we report detailed experimental results regarding the research questions addressed in the main text with additional baselines.

\paragraph{How does MIPS perform with varying logging and target policies?}
We compare the MSE, squared bias, and variance of the estimators (DM, IPS, DR, MIPS, and MIPS with the true weights) with varying logging and target policies. We can do this by varying the values of $\beta$ and $\epsilon$ as described in Section~\ref{sec:empirical}. Note that we set $\beta=-1$ and $\epsilon=0.05$ for all synthetic results in the main text.

First, Figure~\ref{fig:varying_beta} reports the results with varying logging policies ($\beta \in \{-3,-2,-1,0,1,2,3\}$) and with a near-optimal/near-deterministic target policy defined by $\epsilon=0.05$ (fixed). A large negative value of $\beta$ leads to a worse logging policy, meaning that it creates a large discrepancy between logging and target policies in this setup. The left column of Figure~\ref{fig:varying_beta} demonstrates that the MSEs of the estimators generally become larger for larger negative values of $\beta$ as expected. Most notably, the MSEs of IPS and DR blow up for $\beta=-3,-2$ due to their inflated variance as suggested in the right column of the same figure. On the other hand, MIPS and MIPS (true) work robustly for a range of logging policies, suggesting the strong variance reduction for the case with a large discrepancy between policies. DM also suffers from a larger discrepancy between logging and target policies due to its increased bias caused by the extrapolation error issue.

Next, Figure~\ref{fig:varying_eps} shows the results with varying target policies ($\epsilon=\{0.0,0.2,0.4,0.6,0.8,1.0\}$) and with a logging policy slightly worse than uniform random defined by $\beta=-1$ (fixed). A larger value of $\epsilon$ introduces a larger entropy for the target policy, making it closer to the logging policy in this setup (an extreme case with $\epsilon=1.0$ produces a uniform random target policy). On the other hand, $\epsilon=0$ produces the optimal, deterministic target policy, which makes OPE harder given $\beta=-1$. The left column of Figure~\ref{fig:varying_eps} suggests that all estimators perform worse for smaller values of $\epsilon$ as expected. IPS and DR perform worse as their variance increases with decreasing $\epsilon$, while DM performs worse as it produces larger bias. The variance of MIPS also increases with decreasing $\epsilon$, but it is often much smaller and robust than those of IPS and DR. Note that, for the uniform random target policy ($\epsilon=1.0$), all estimators are very accurate and there is no significant difference among the estimators.

\paragraph{How does MIPS perform with varying noise levels?}
Next, we explore how the level of noise on the rewards affects the comparison of the estimators.
To this end, we vary the noise level $\sigma \in \{0.5,1.0,1.5,\ldots,4.0\}$ where $\sigma$ is the standard deviation of the Gaussian noise, i.e., $r \sim \mathcal{N} (q(x,a), \sigma^2)$. As stated in the main text, the variance of IPS grows when the reward is noisy. Theorem~\ref{thm:variance_reduction} also implies that the variance reduction of MIPS becomes more appealing with the noisy rewards. Figure~\ref{fig:varying_noise} empirically supports these claims. Specifically, IPS significantly exacerbates its MSE from 0.55 (when $\sigma=0.5$) to 3.22 (when $\sigma=4.0$). MIPS also struggles with noisy rewards, but the improvement of MIPS compared to IPS/DR becomes larger with the added noise. When the noise level is small ($\sigma=0.5$), $\mseratio=2.97$, while $\mseratio=14.98$ when the noise is large ($\sigma=4.0$). Different from IPS, DR, and MIPS, DM is not affected so much by the noise level and becomes increasingly better than IPS and DR in noisy environments. Nontheless, MIPS achieves much smaller MSE than DM even with noisy rewards.

\begin{figure*}[th]
\scalebox{0.95}{
\begin{tabular}{ccc}
\toprule
\textbf{MSE} & \textbf{Squared Bias} & \textbf{Variance} \\ \midrule \midrule
\begin{minipage}{0.33\hsize}
    \begin{center}
        \includegraphics[clip, width=5.8cm]{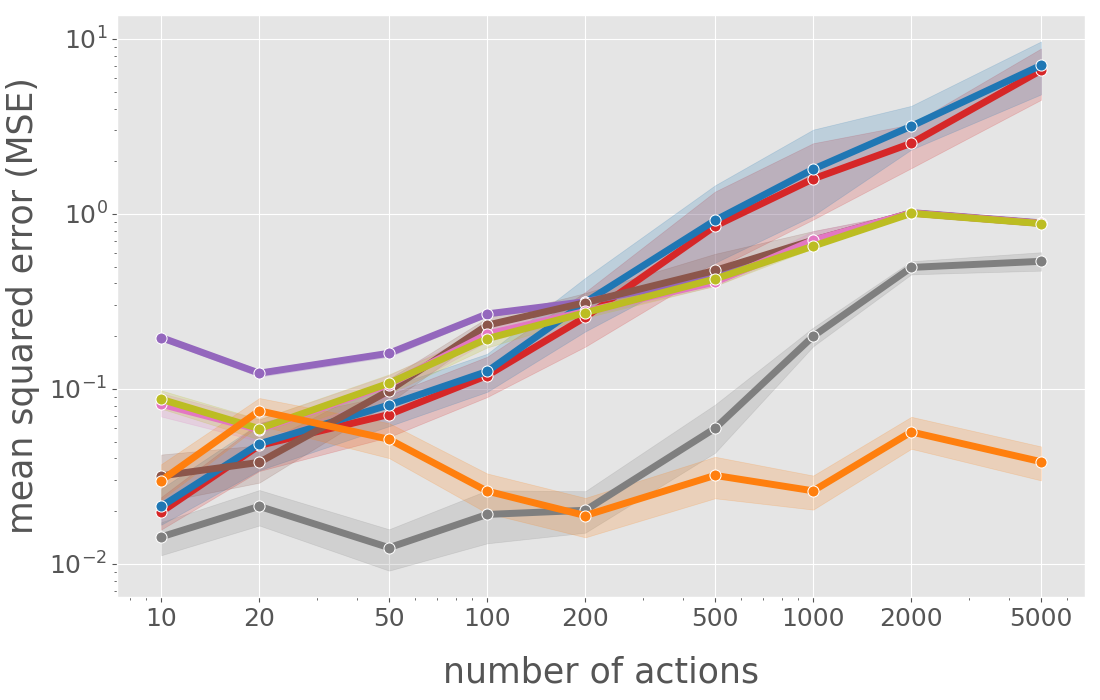}
    \end{center}
\end{minipage}
&
\begin{minipage}{0.33\hsize}
    \begin{center}
        \includegraphics[clip, width=5.8cm]{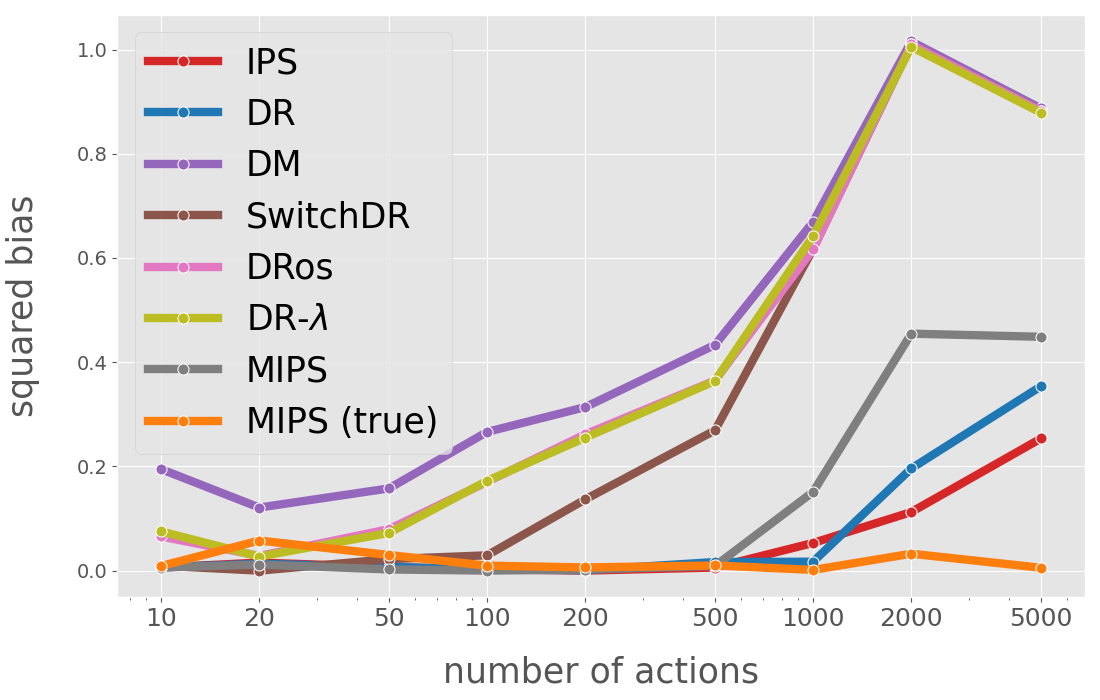}
    \end{center}
\end{minipage}
&
\begin{minipage}{0.33\hsize}
    \begin{center}
        \includegraphics[clip, width=5.8cm]{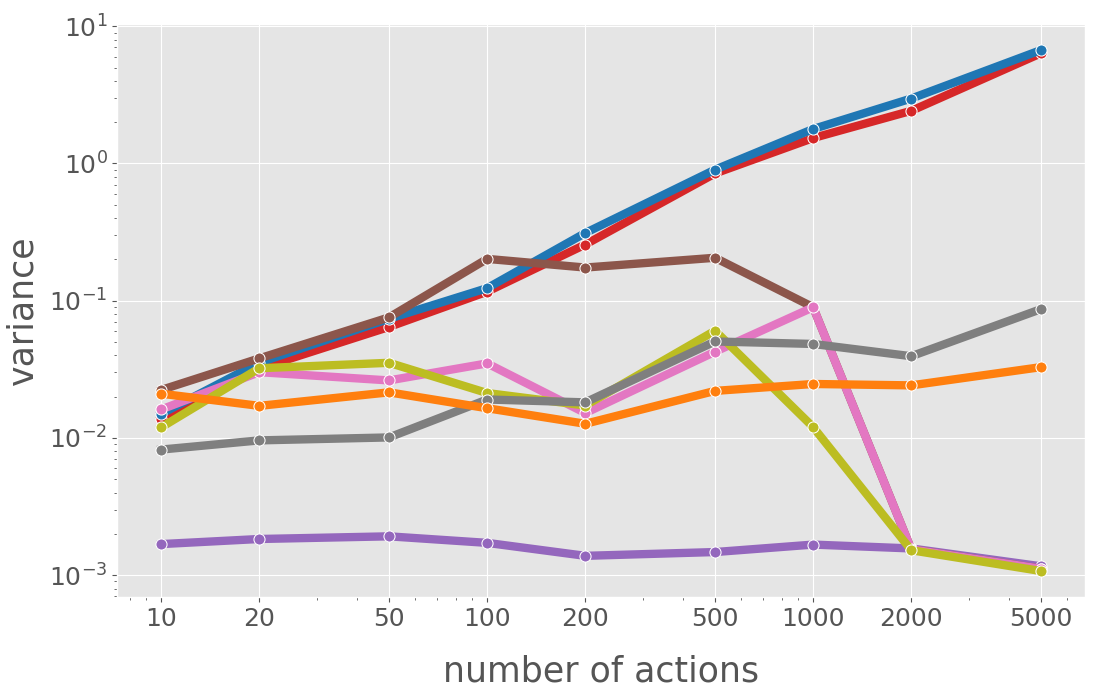}
    \end{center}
\end{minipage}
\\
\multicolumn{3}{c}{
\begin{minipage}{1.0\hsize}
\begin{center}
\caption{MSE, Squared Bias, and Variance with \textbf{varying number of actions}}
\label{fig:beta=-1,eps=0.05_start}
\end{center}
\end{minipage}
}
\\
\begin{minipage}{0.33\hsize}
    \begin{center}
        \includegraphics[clip, width=5.8cm]{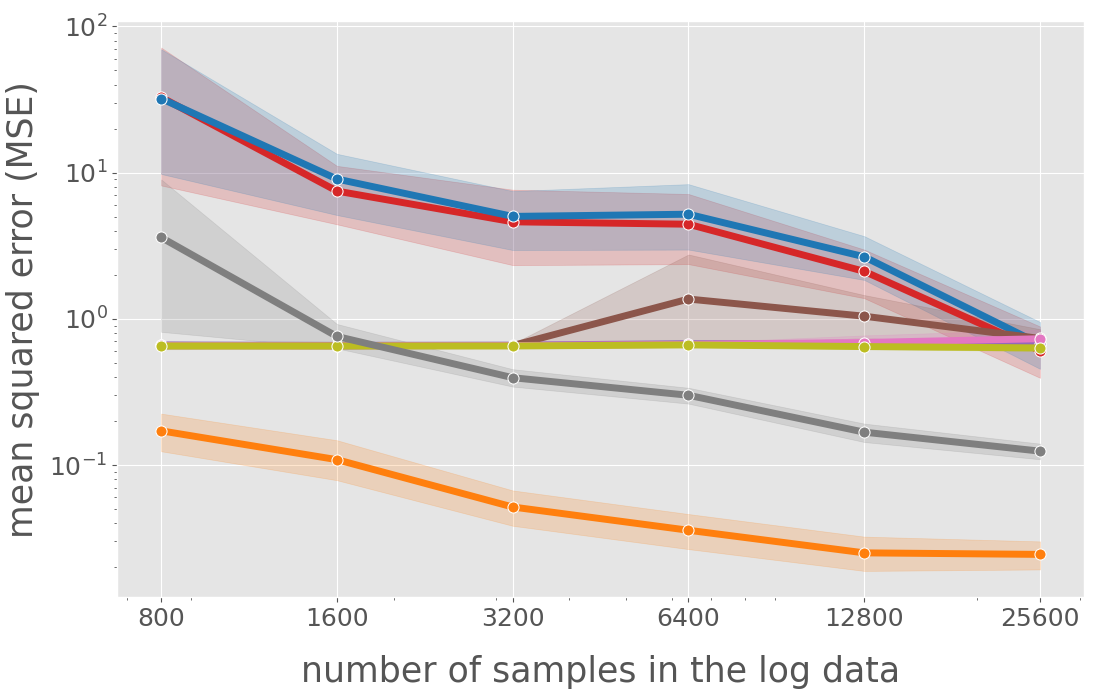}
    \end{center}
\end{minipage}
&
\begin{minipage}{0.33\hsize}
    \begin{center}
        \includegraphics[clip, width=5.8cm]{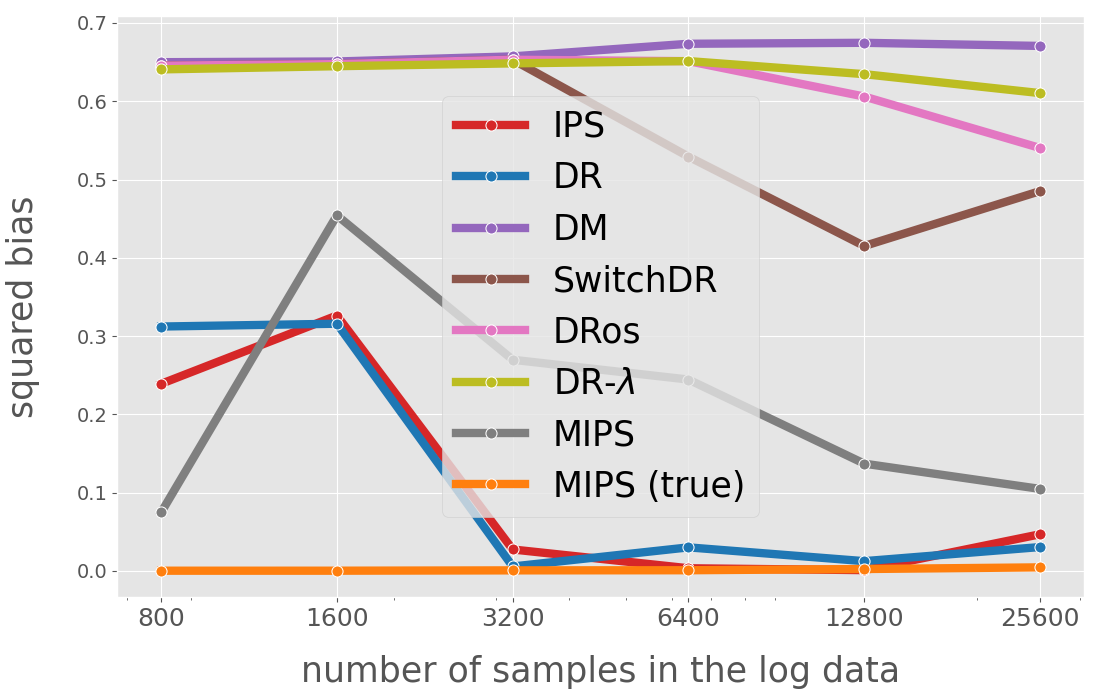}
    \end{center}
\end{minipage}
&
\begin{minipage}{0.33\hsize}
    \begin{center}
        \includegraphics[clip, width=5.8cm]{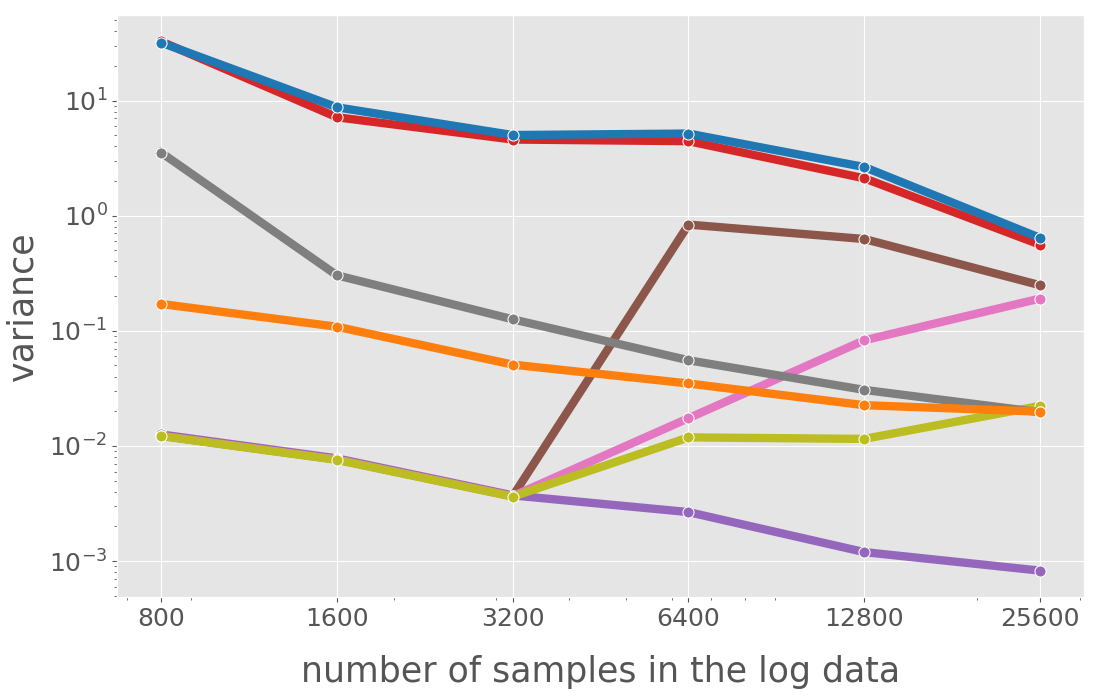}
    \end{center}
\end{minipage}
\\
\multicolumn{3}{c}{
\begin{minipage}{0.95\hsize}
\begin{center}
\caption{MSE, Squared Bias, and Variance with \textbf{varying sample size}}
\end{center}
\end{minipage}
}
\\
\begin{minipage}{0.33\hsize}
    \begin{center}
        \includegraphics[clip, width=5.8cm]{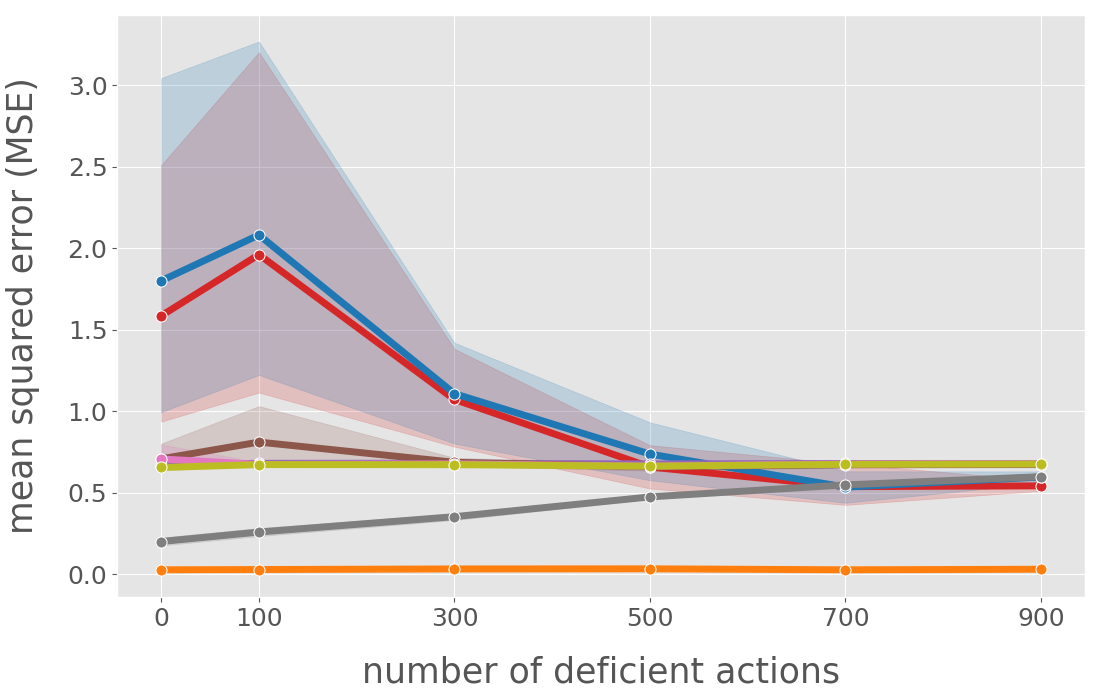}
    \end{center}
\end{minipage}
&
\begin{minipage}{0.33\hsize}
    \begin{center}
        \includegraphics[clip, width=5.8cm]{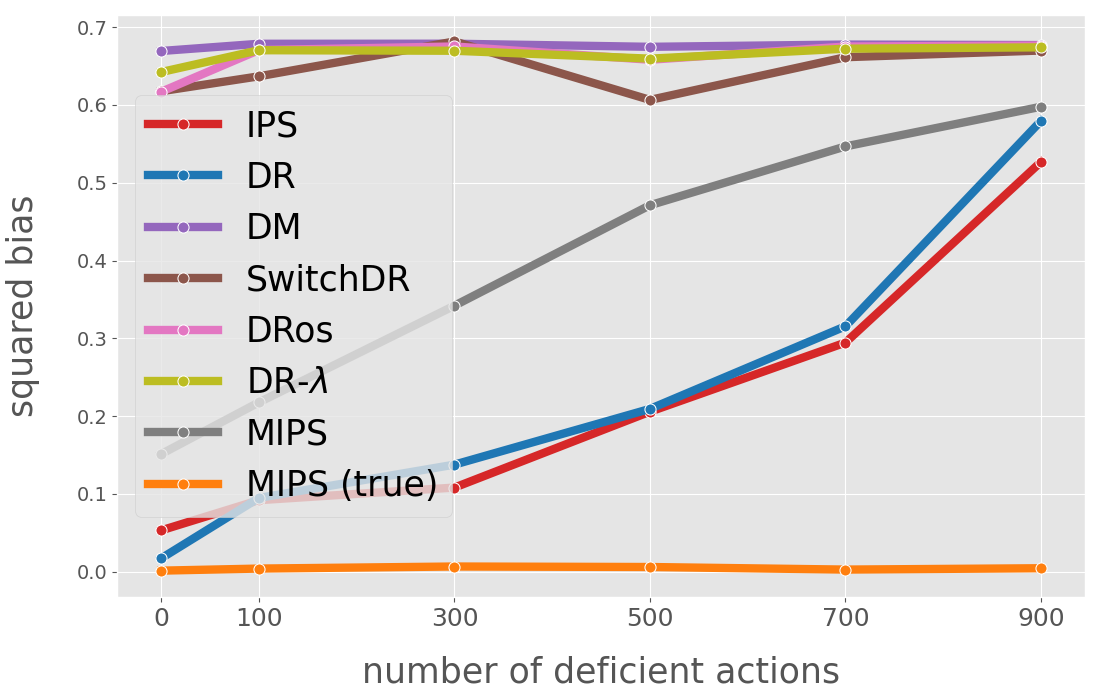}
    \end{center}
\end{minipage}
&
\begin{minipage}{0.33\hsize}
    \begin{center}
        \includegraphics[clip, width=5.8cm]{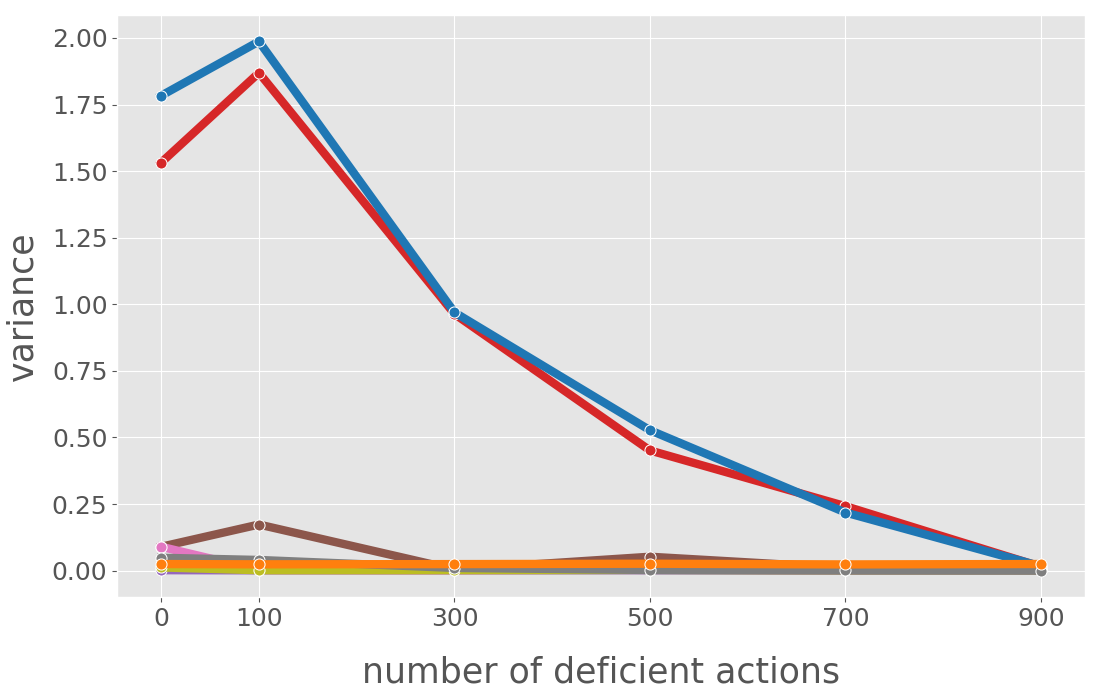}
    \end{center}
\end{minipage}
\\
\multicolumn{3}{c}{
\begin{minipage}{0.95\hsize}
\begin{center}
\caption{MSE, Squared Bias, and Variance \textbf{with varying number of deficient actions}}
\end{center}
\end{minipage}
}
\\ 
\begin{minipage}{0.30\hsize}
    \begin{center}
        \includegraphics[clip, width=5.8cm]{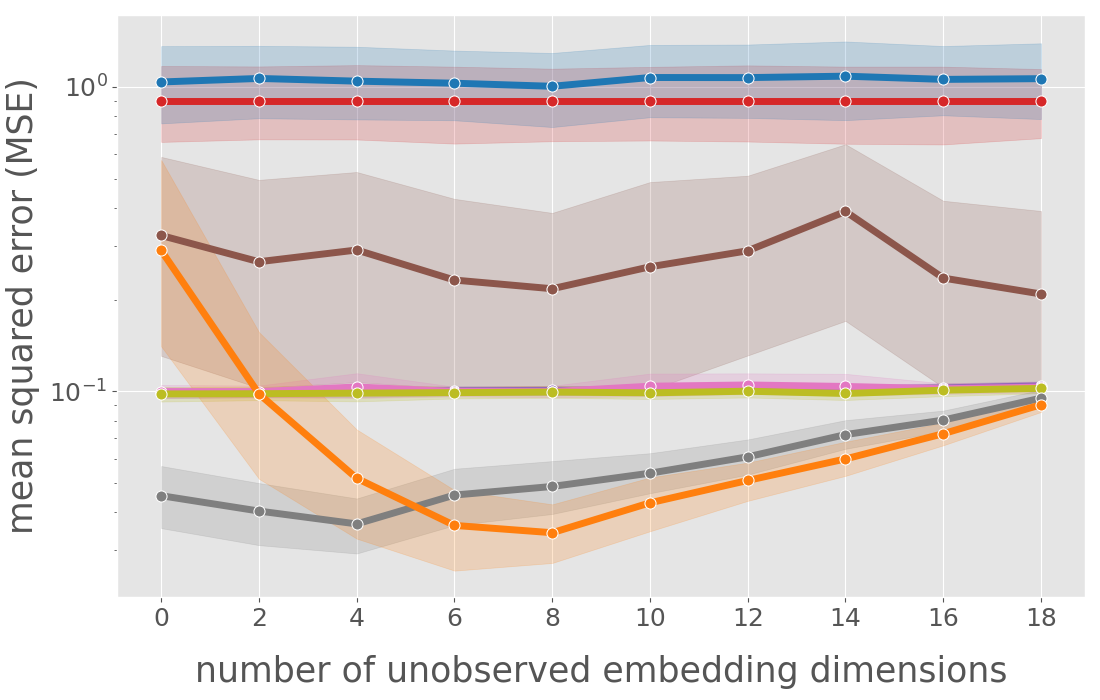}
    \end{center}
\end{minipage}
&
\begin{minipage}{0.30\hsize}
    \begin{center}
        \includegraphics[clip, width=5.8cm]{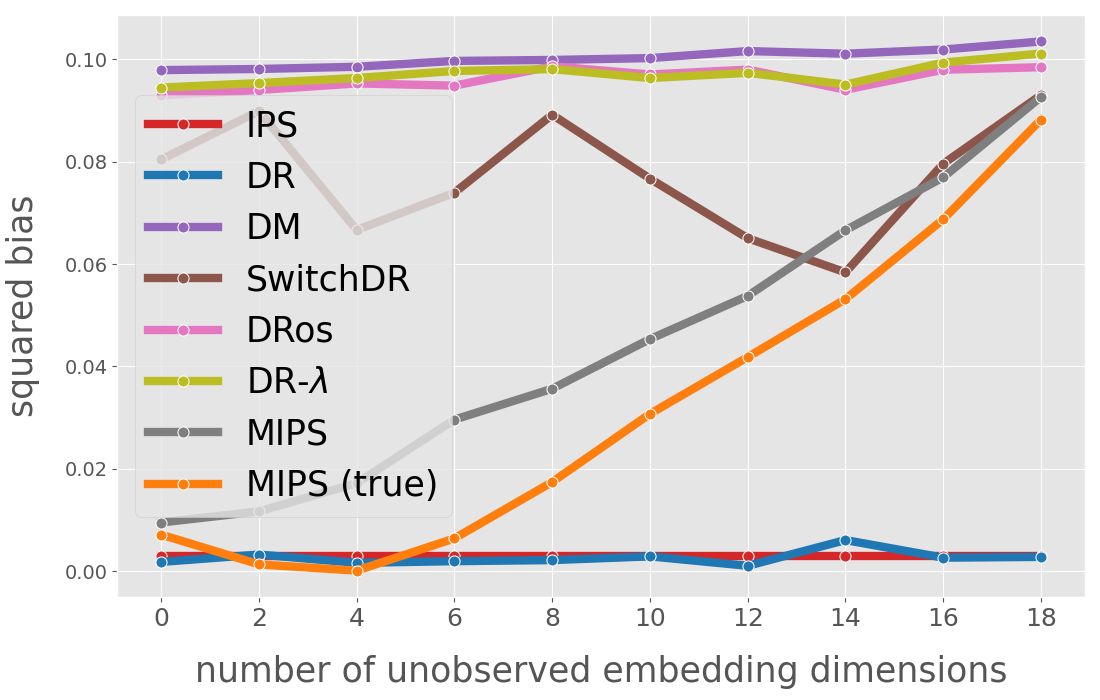}
    \end{center}
\end{minipage}
&
\begin{minipage}{0.30\hsize}
    \begin{center}
        \includegraphics[clip, width=5.8cm]{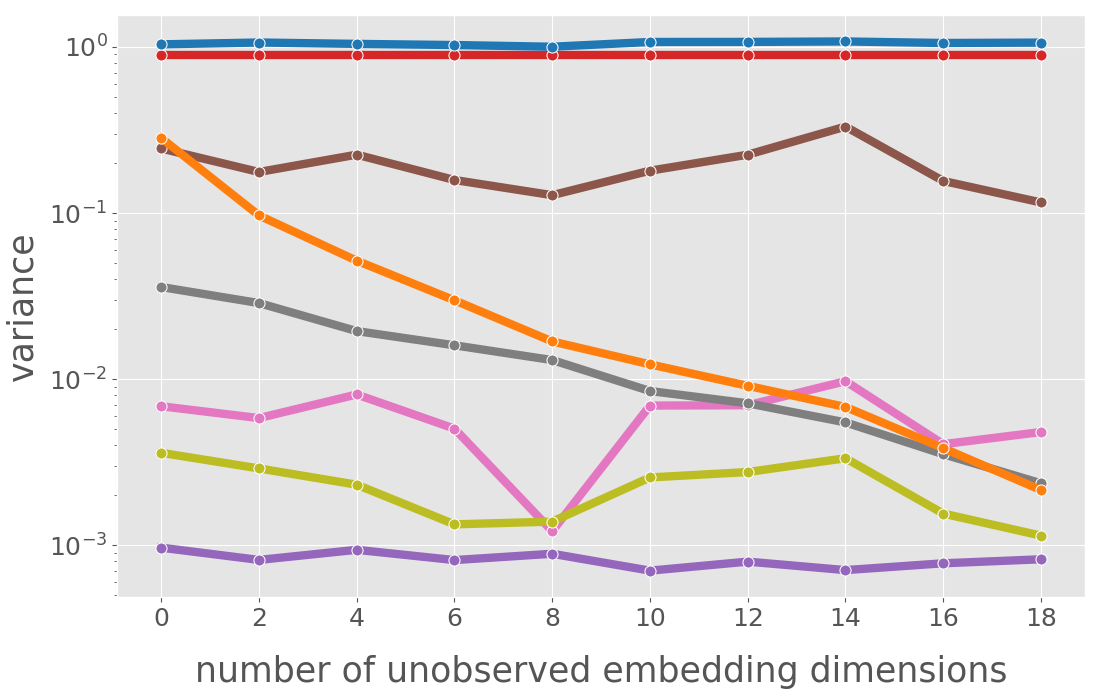}
    \end{center}
\end{minipage}
\\
\multicolumn{3}{c}{
\begin{minipage}{0.95\hsize}
\begin{center}
\caption{MSE, Squared Bias, and Variance with \textbf{varying number of unobserved dimensions in action embeddings}}
\label{fig:beta=-1,eps=0.05_end}
\end{center}
\end{minipage}
}
\\ 
\bottomrule
\end{tabular}
}
\vskip 0.1in
\raggedright
\fontsize{9pt}{9pt}\selectfont \textit{Note}:
We set $\beta=-1$ and $\epsilon=0.05$, which produce \textbf{logging policy slightly worse than uniform random} and \textbf{near-optimal/near-deterministic target policy}.
The results are averaged over 100 different sets of synthetic logged data replicated with different random seeds.
The shaded regions in the MSE plots represent the 95\% confidence intervals estimated with bootstrap.
The y-axis of MSE and Variance plots (the left and right columns) is reported on log-scale.
\end{figure*}

\begin{figure*}[th]
\scalebox{0.95}{
\begin{tabular}{ccc}
\toprule
\textbf{MSE} & \textbf{Squared Bias} & \textbf{Variance} \\ \midrule \midrule
\begin{minipage}{0.33\hsize}
    \begin{center}
        \includegraphics[clip, width=5.8cm]{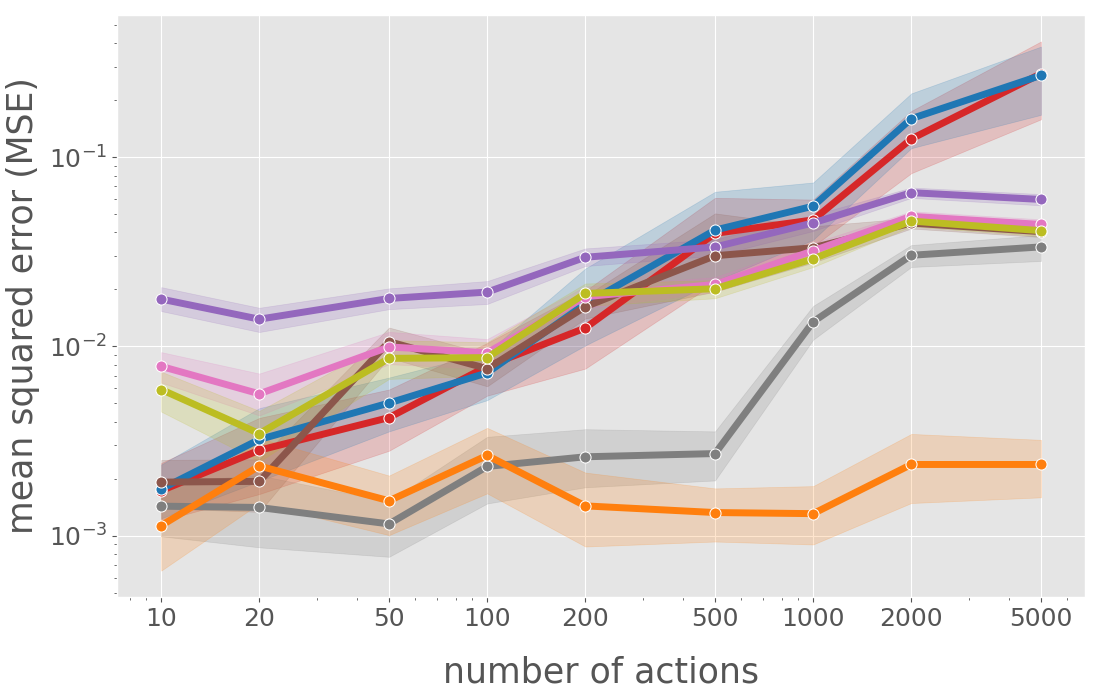}
    \end{center}
\end{minipage}
&
\begin{minipage}{0.33\hsize}
    \begin{center}
        \includegraphics[clip, width=5.8cm]{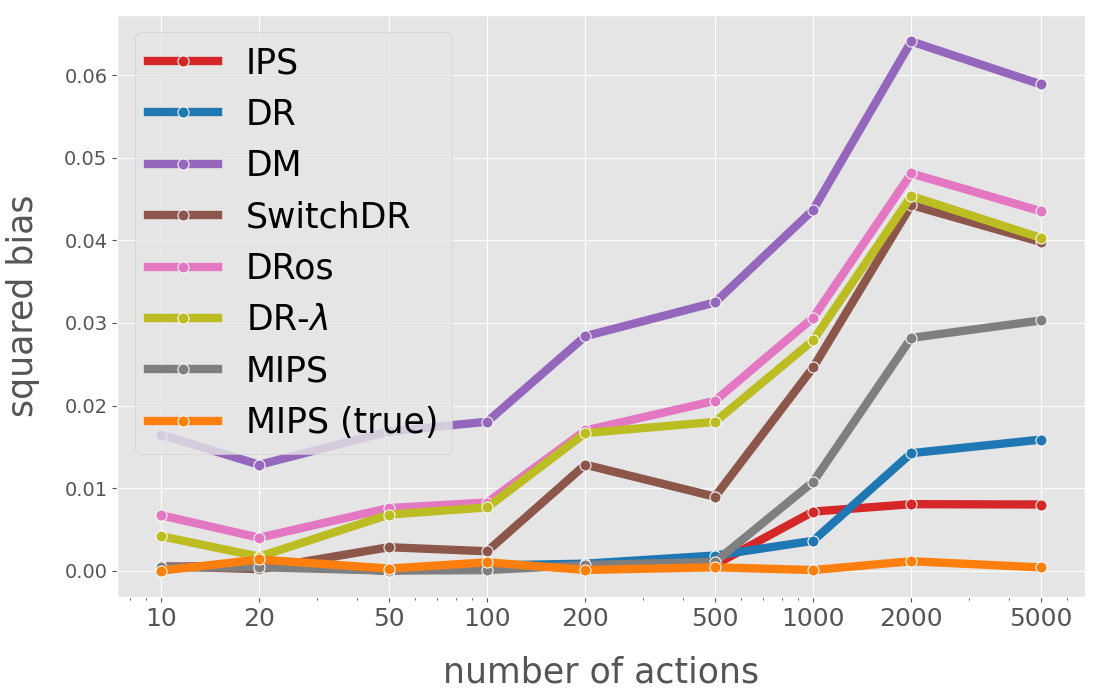}
    \end{center}
\end{minipage}
&
\begin{minipage}{0.33\hsize}
    \begin{center}
        \includegraphics[clip, width=5.8cm]{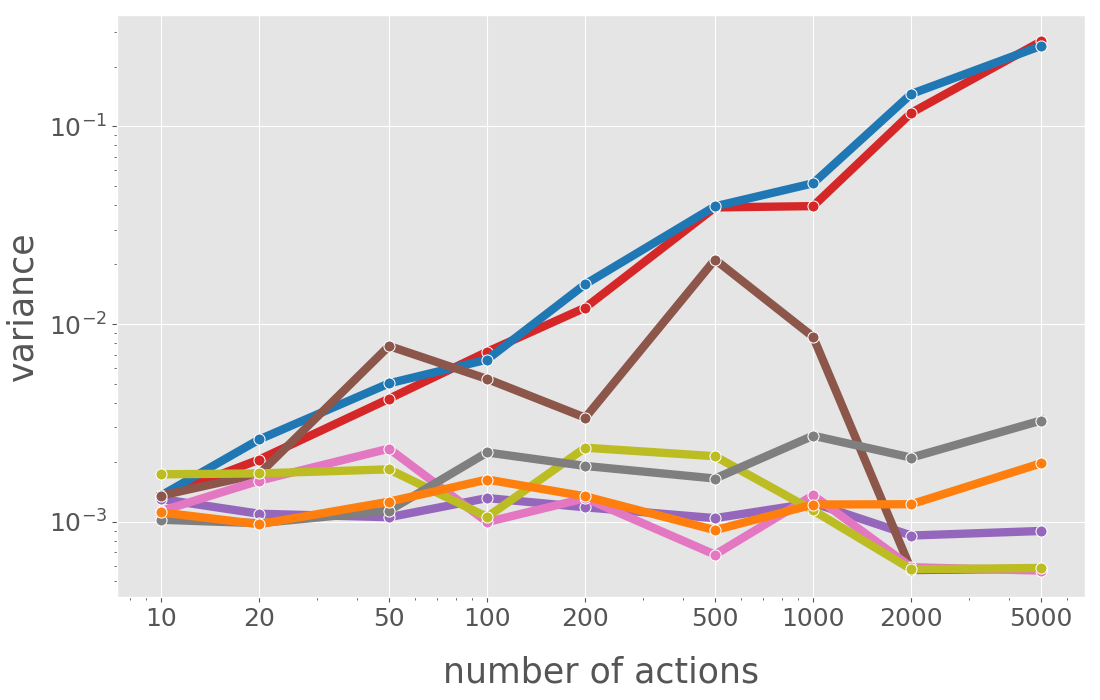}
    \end{center}
\end{minipage}
\\
\multicolumn{3}{c}{
\begin{minipage}{1.0\hsize}
\begin{center}
\caption{MSE, Squared Bias, and Variance with \textbf{varying number of actions}}
\label{fig:beta=-1,eps=0.8_start}
\end{center}
\end{minipage}
}
\\
\begin{minipage}{0.33\hsize}
    \begin{center}
        \includegraphics[clip, width=5.8cm]{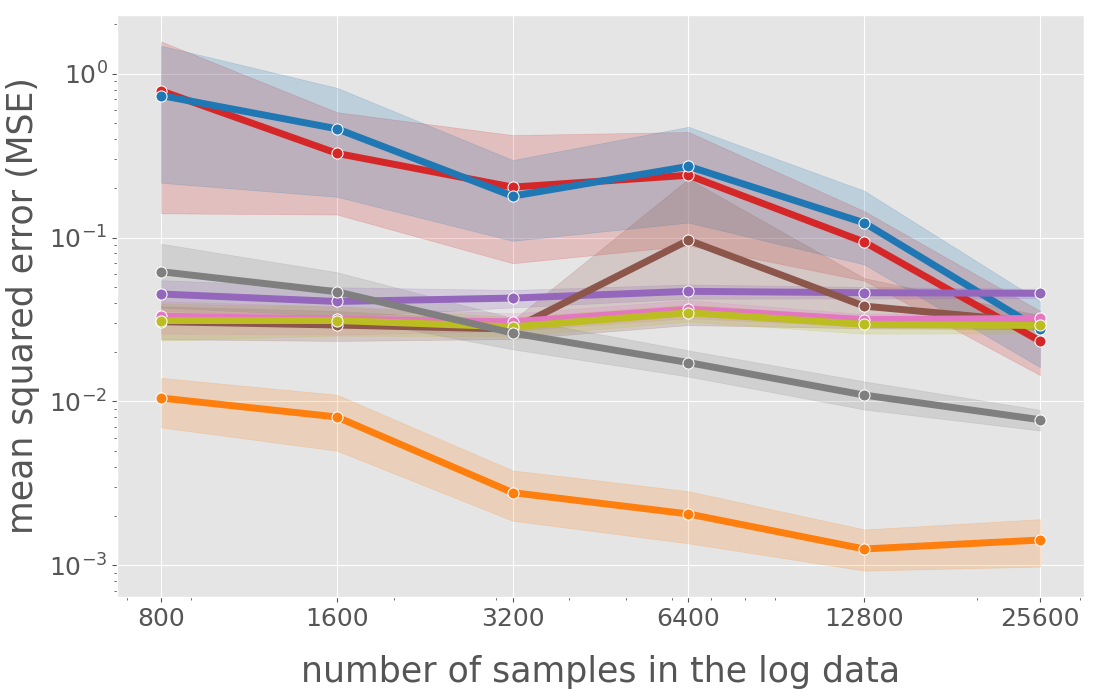}
    \end{center}
\end{minipage}
&
\begin{minipage}{0.33\hsize}
    \begin{center}
        \includegraphics[clip, width=5.8cm]{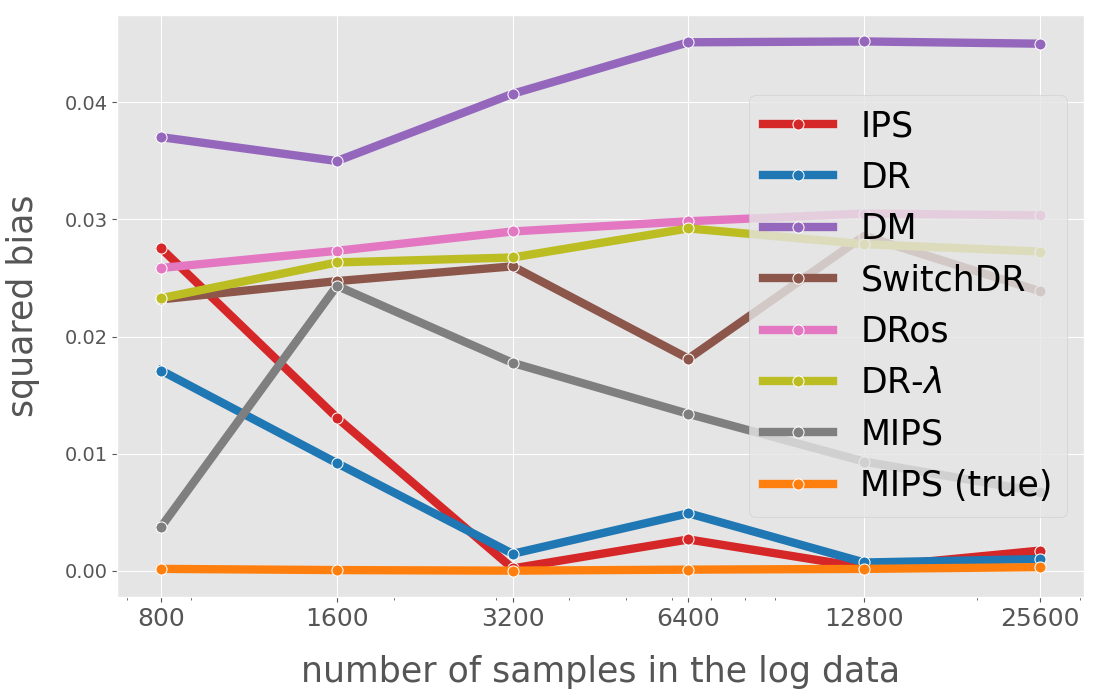}
    \end{center}
\end{minipage}
&
\begin{minipage}{0.33\hsize}
    \begin{center}
        \includegraphics[clip, width=5.8cm]{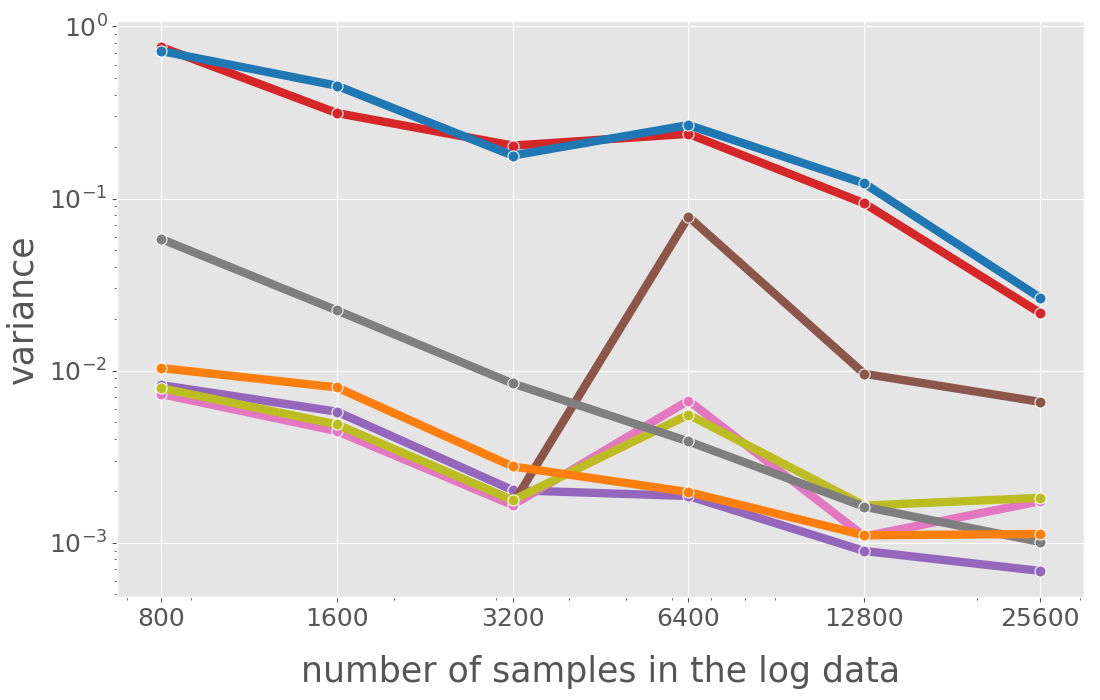}
    \end{center}
\end{minipage}
\\
\multicolumn{3}{c}{
\begin{minipage}{0.95\hsize}
\begin{center}
\caption{MSE, Squared Bias, and Variance with \textbf{varying sample size}}
\end{center}
\end{minipage}
}
\\
\begin{minipage}{0.33\hsize}
    \begin{center}
        \includegraphics[clip, width=5.8cm]{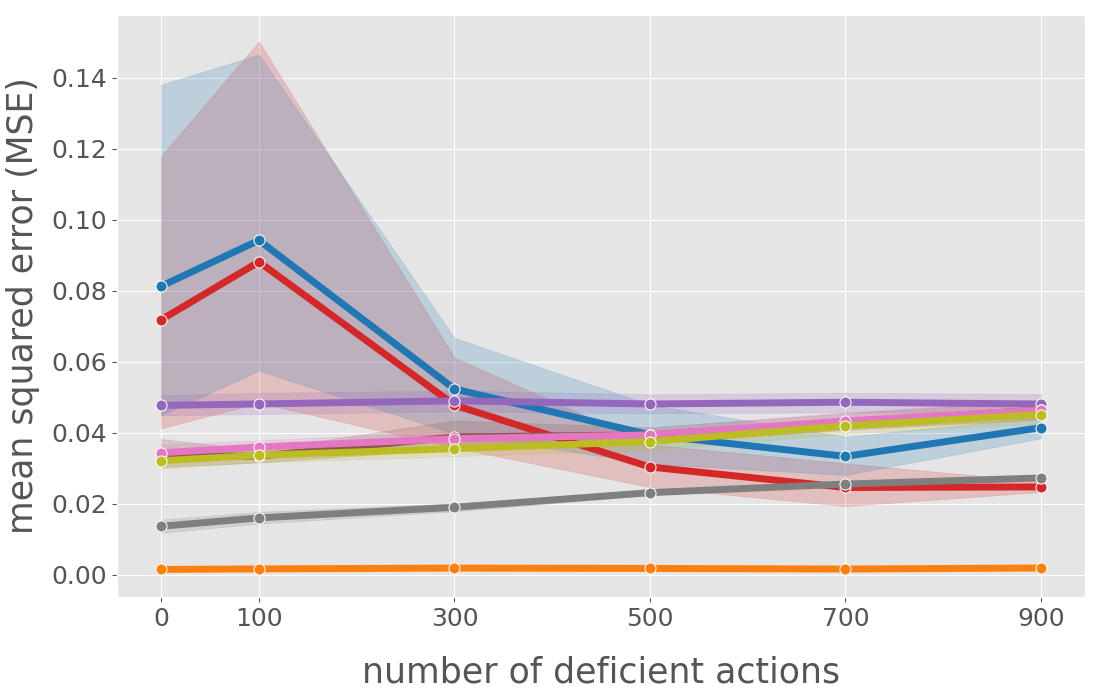}
    \end{center}
\end{minipage}
&
\begin{minipage}{0.33\hsize}
    \begin{center}
        \includegraphics[clip, width=5.8cm]{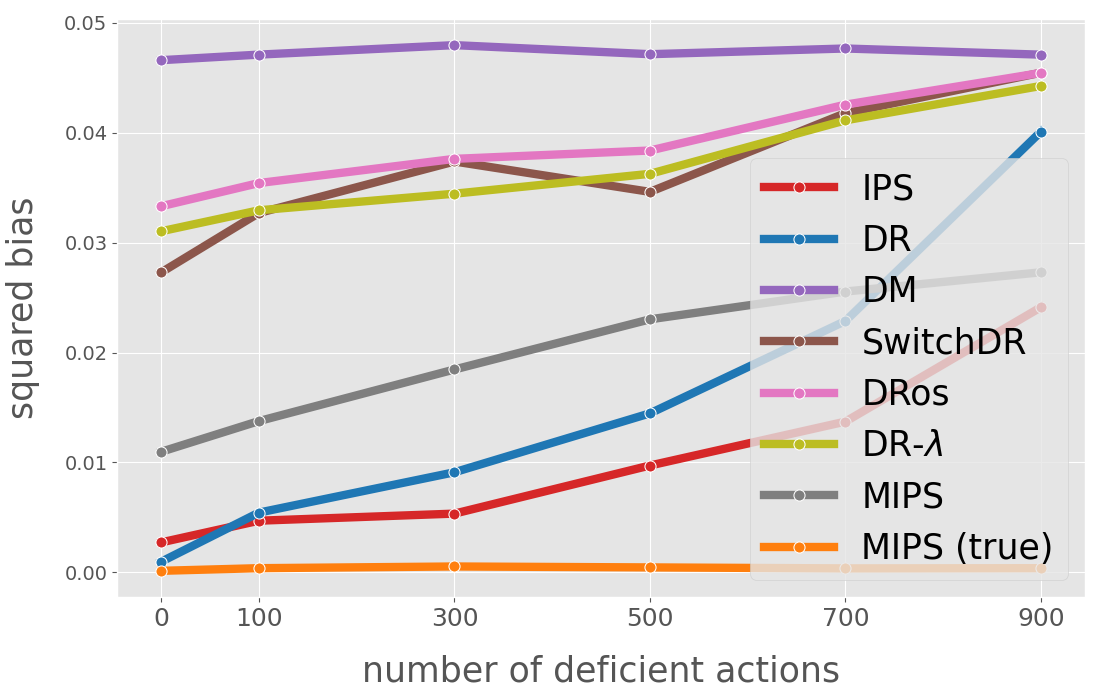}
    \end{center}
\end{minipage}
&
\begin{minipage}{0.33\hsize}
    \begin{center}
        \includegraphics[clip, width=5.8cm]{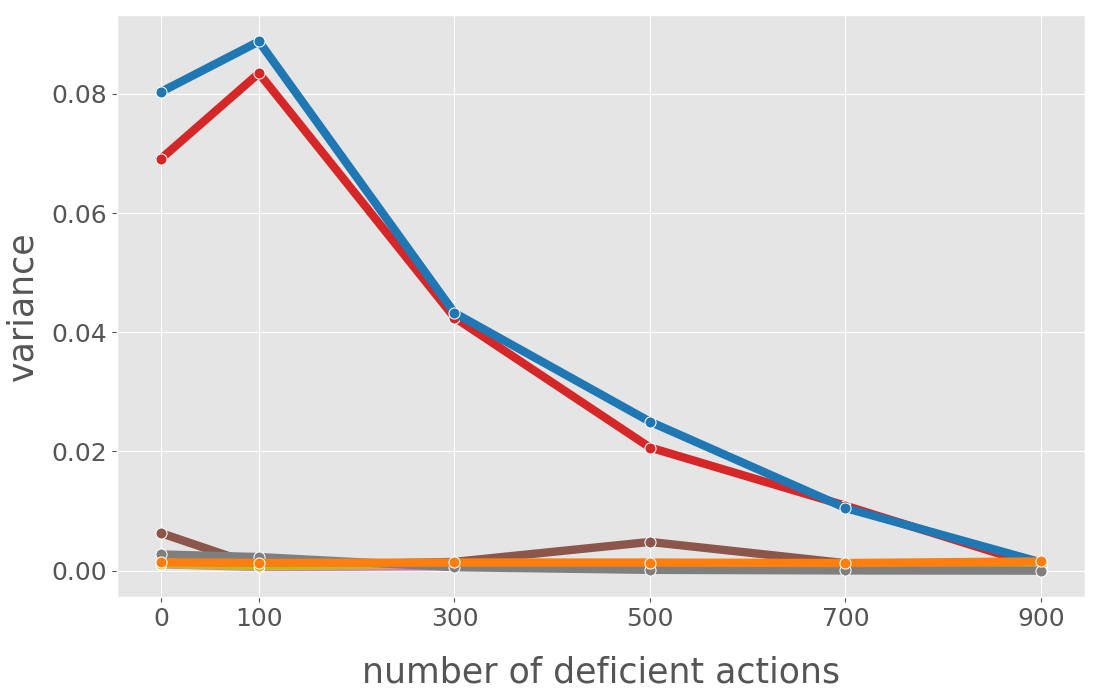}
    \end{center}
\end{minipage}
\\
\multicolumn{3}{c}{
\begin{minipage}{0.95\hsize}
\begin{center}
\caption{MSE, Squared Bias, and Variance \textbf{with varying number of deficient actions}}
\end{center}
\end{minipage}
}
\\ 
\begin{minipage}{0.30\hsize}
    \begin{center}
        \includegraphics[clip, width=5.8cm]{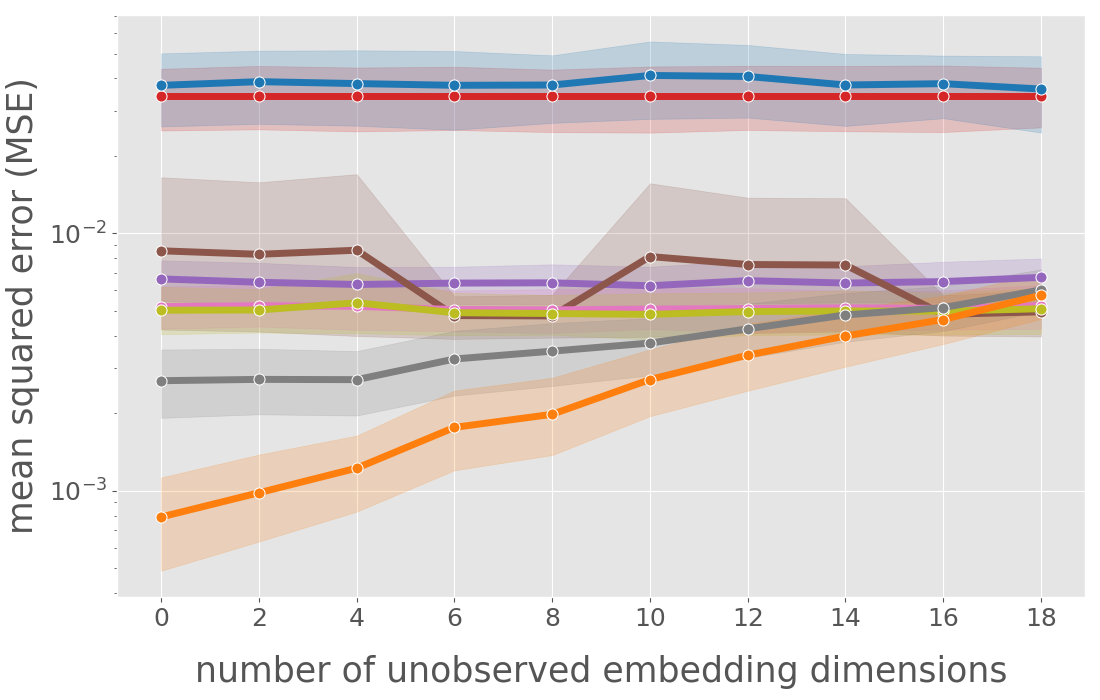}
    \end{center}
\end{minipage}
&
\begin{minipage}{0.30\hsize}
    \begin{center}
        \includegraphics[clip, width=5.8cm]{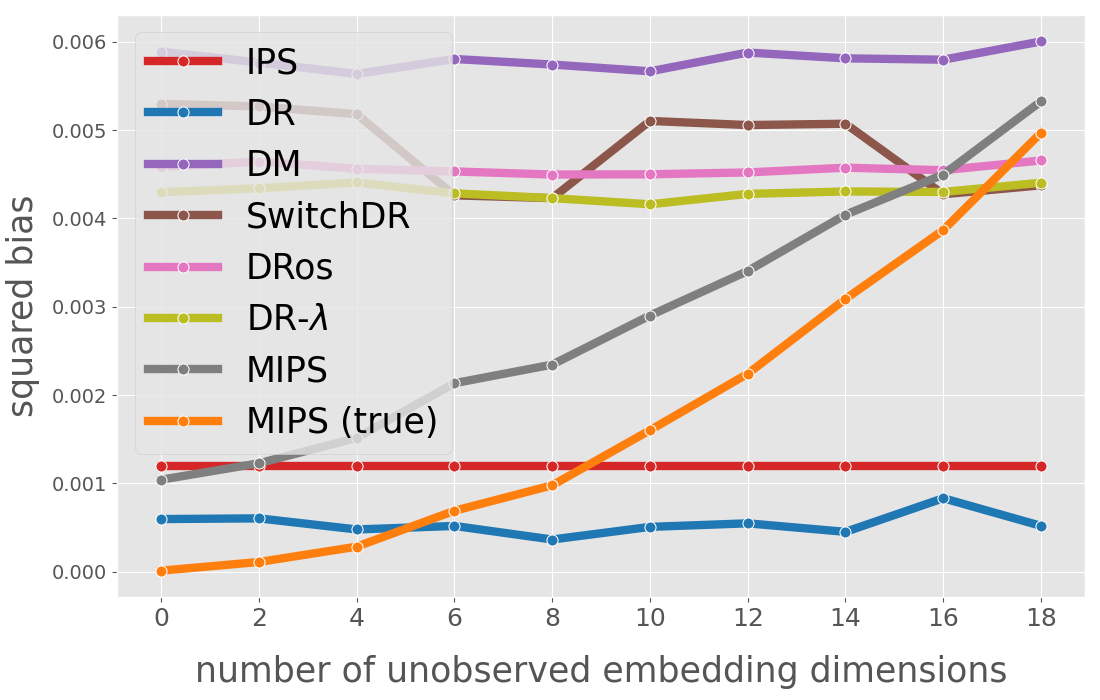}
    \end{center}
\end{minipage}
&
\begin{minipage}{0.30\hsize}
    \begin{center}
        \includegraphics[clip, width=5.8cm]{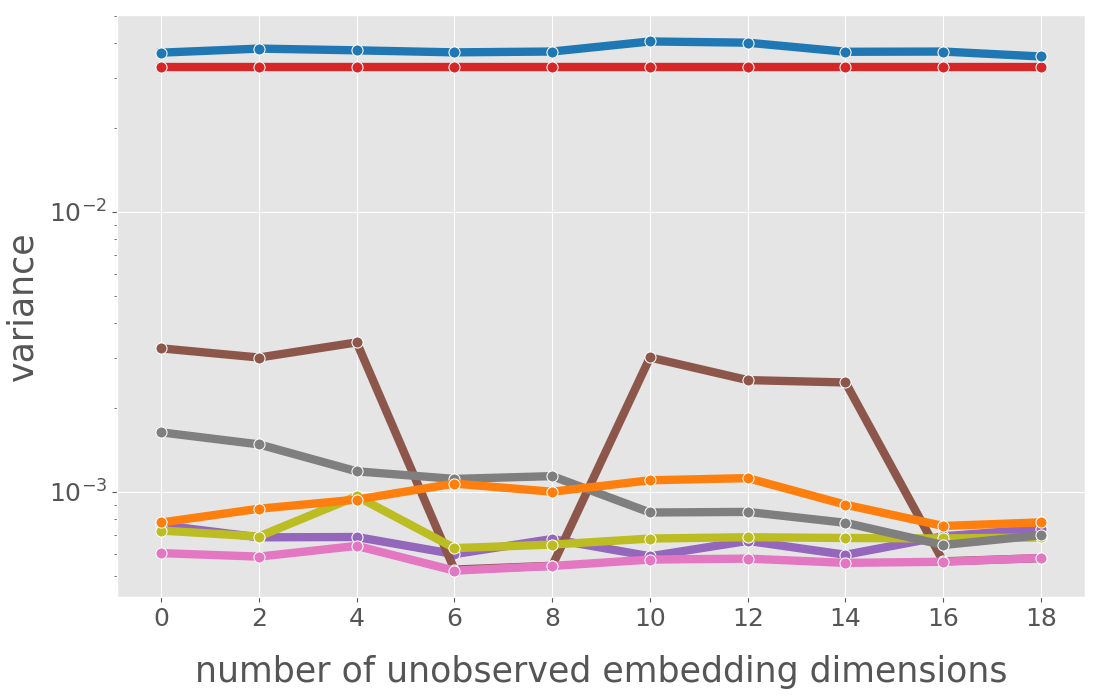}
    \end{center}
\end{minipage}
\\
\multicolumn{3}{c}{
\begin{minipage}{0.95\hsize}
\begin{center}
\caption{MSE, Squared Bias, and Variance with \textbf{varying number of unobserved dimensions in action embeddings}}
\label{fig:beta=-1,eps=0.8_end}
\end{center}
\end{minipage}
}
\\ 
\bottomrule
\end{tabular}
}
\vskip 0.1in
\raggedright
\fontsize{9pt}{9pt}\selectfont \textit{Note}:
We set $\beta=-1$ and $\epsilon=0.8$, which produce \textbf{logging policy slightly worse than uniform random} and \textbf{near-uniform target policy}.
The results are averaged over 100 different sets of synthetic logged data replicated with different random seeds.
The shaded regions in the MSE plots represent the 95\% confidence intervals estimated with bootstrap.
The y-axis of MSE and Variance plots (the left and right columns) is reported on log-scale.
\end{figure*}

\begin{figure*}[th]
\scalebox{0.95}{
\begin{tabular}{ccc}
\toprule
\textbf{MSE} & \textbf{Squared Bias} & \textbf{Variance} \\ \midrule \midrule
\begin{minipage}{0.33\hsize}
    \begin{center}
        \includegraphics[clip, width=5.8cm]{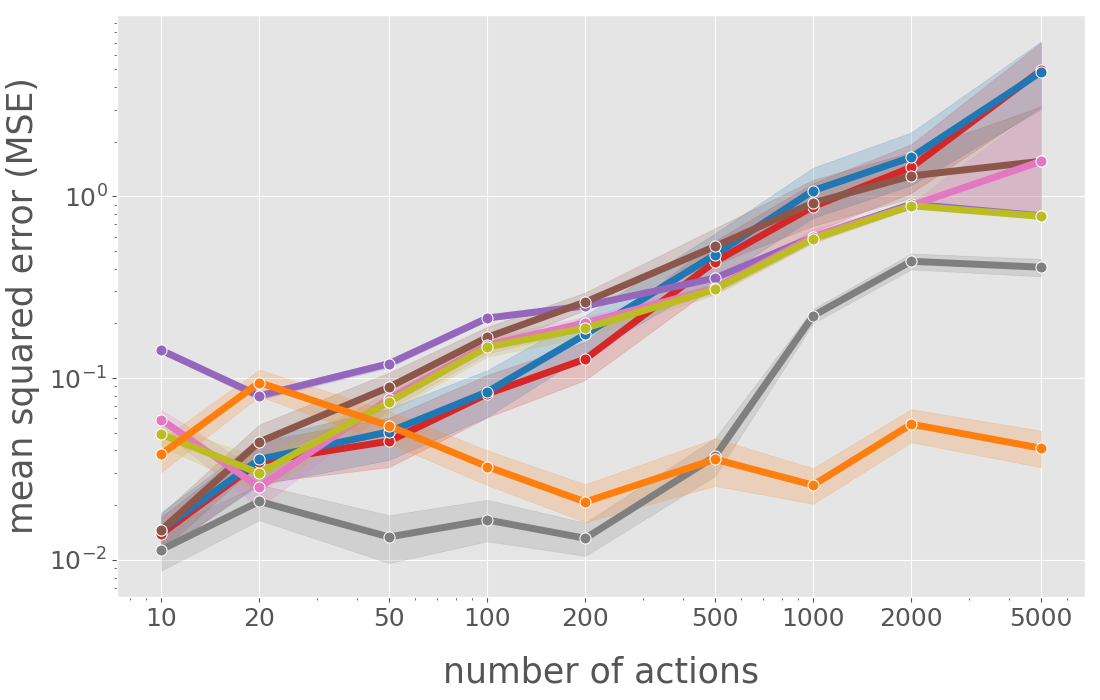}
    \end{center}
\end{minipage}
&
\begin{minipage}{0.33\hsize}
    \begin{center}
        \includegraphics[clip, width=5.8cm]{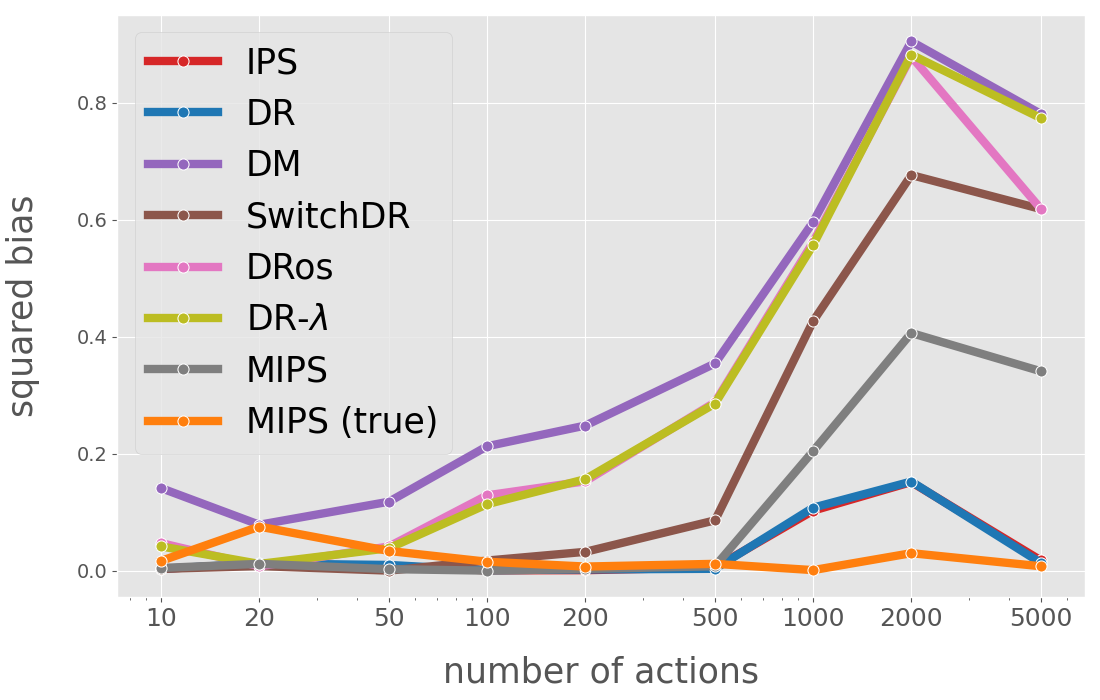}
    \end{center}
\end{minipage}
&
\begin{minipage}{0.33\hsize}
    \begin{center}
        \includegraphics[clip, width=5.8cm]{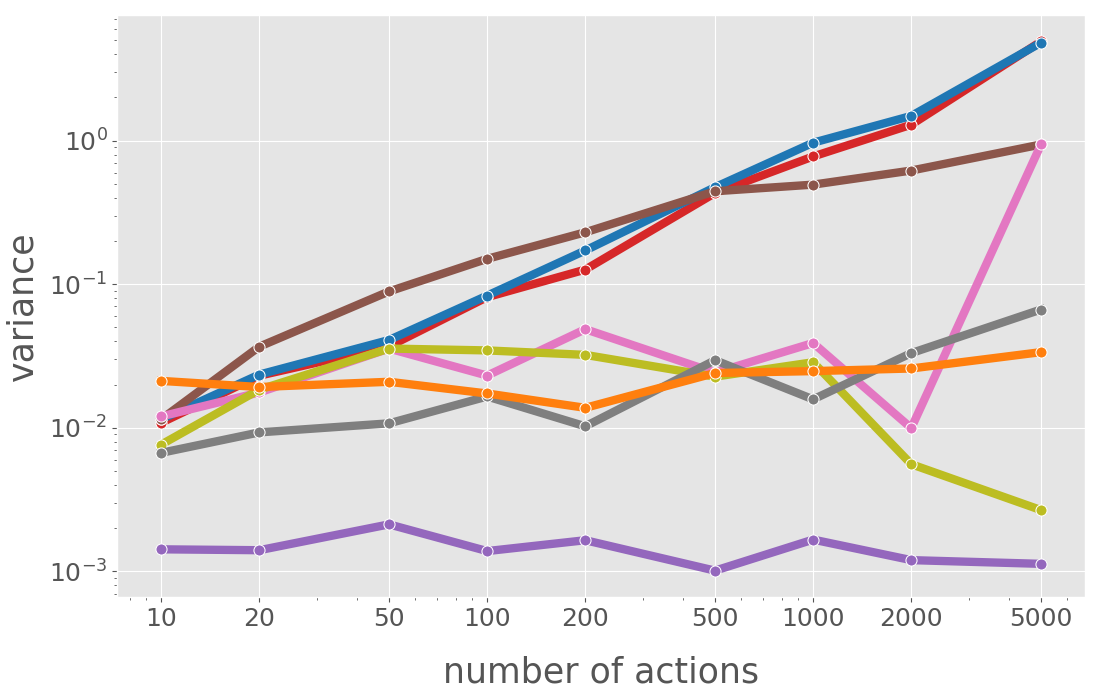}
    \end{center}
\end{minipage}
\\
\multicolumn{3}{c}{
\begin{minipage}{1.0\hsize}
\begin{center}
\caption{MSE, Squared Bias, and Variance with \textbf{varying number of actions}}
\label{fig:beta=0,eps=0.05_start}
\end{center}
\end{minipage}
}
\\
\begin{minipage}{0.33\hsize}
    \begin{center}
        \includegraphics[clip, width=5.8cm]{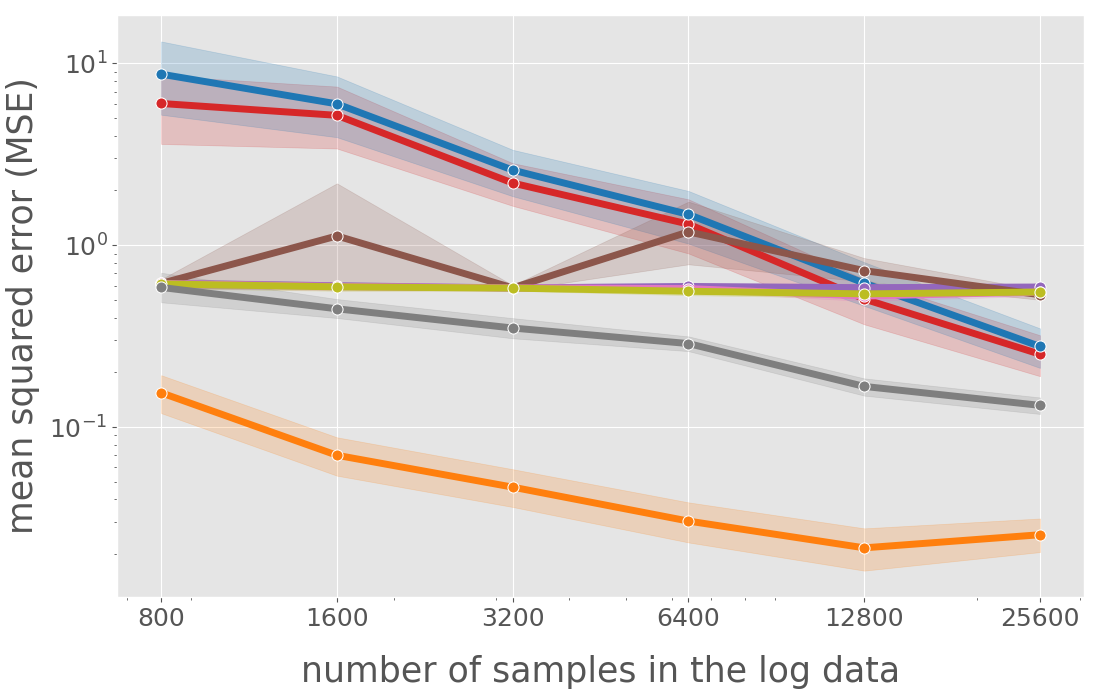}
    \end{center}
\end{minipage}
&
\begin{minipage}{0.33\hsize}
    \begin{center}
        \includegraphics[clip, width=5.8cm]{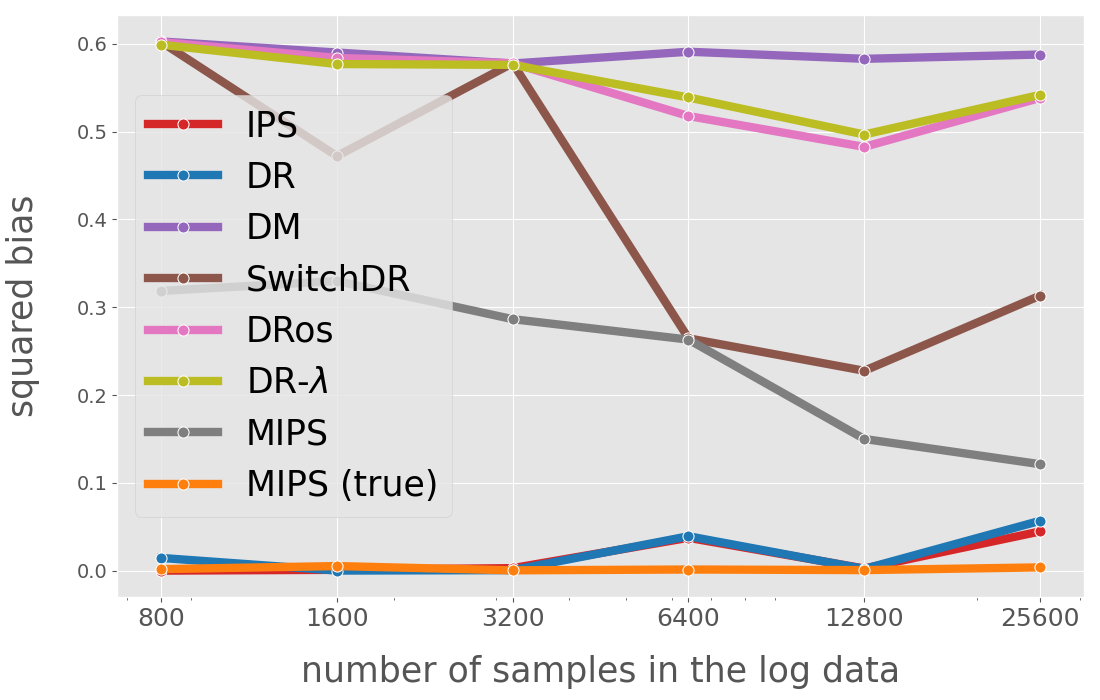}
    \end{center}
\end{minipage}
&
\begin{minipage}{0.33\hsize}
    \begin{center}
        \includegraphics[clip, width=5.8cm]{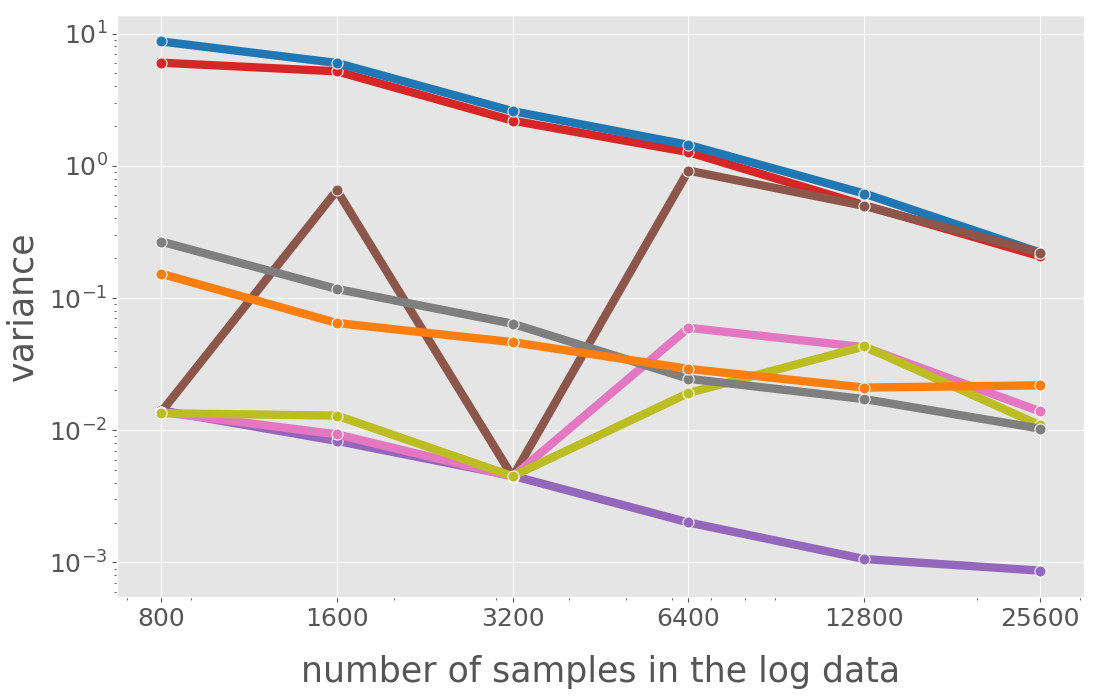}
    \end{center}
\end{minipage}
\\
\multicolumn{3}{c}{
\begin{minipage}{0.95\hsize}
\begin{center}
\caption{MSE, Squared Bias, and Variance with \textbf{varying sample size}}
\end{center}
\end{minipage}
}
\\
\begin{minipage}{0.33\hsize}
    \begin{center}
        \includegraphics[clip, width=5.8cm]{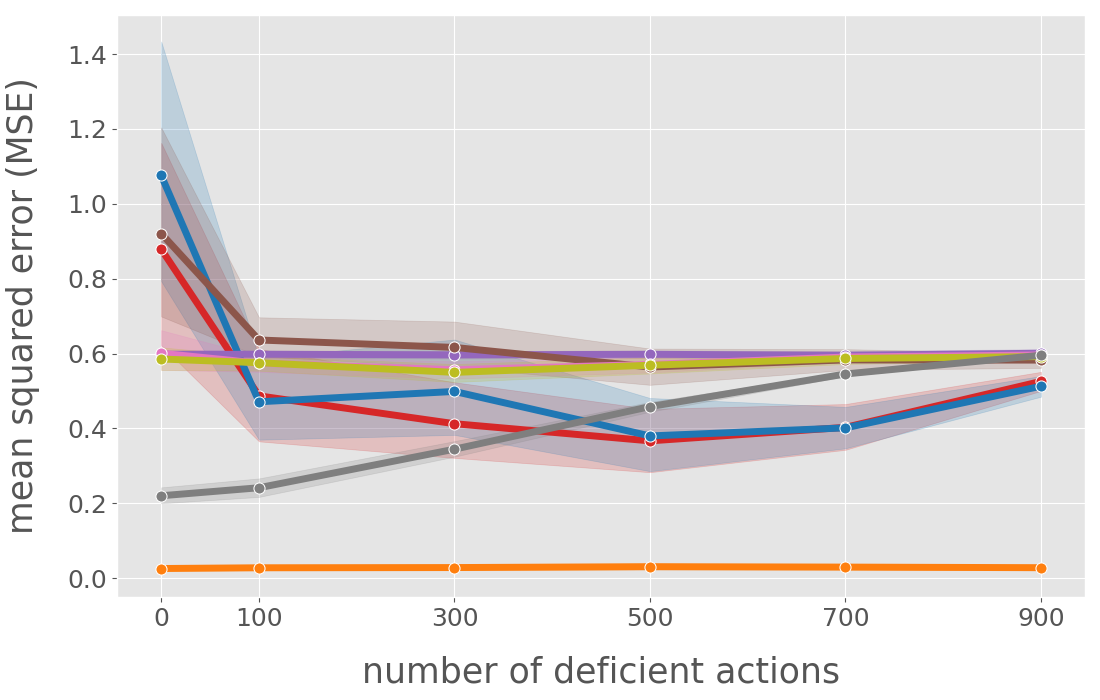}
    \end{center}
\end{minipage}
&
\begin{minipage}{0.33\hsize}
    \begin{center}
        \includegraphics[clip, width=5.8cm]{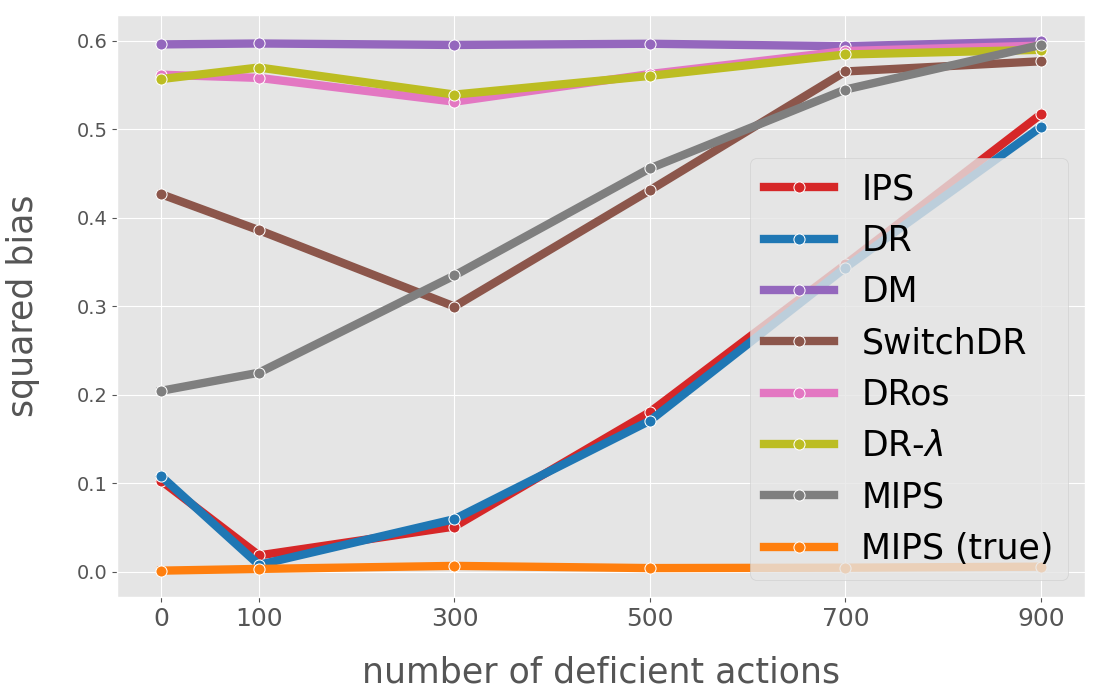}
    \end{center}
\end{minipage}
&
\begin{minipage}{0.33\hsize}
    \begin{center}
        \includegraphics[clip, width=5.8cm]{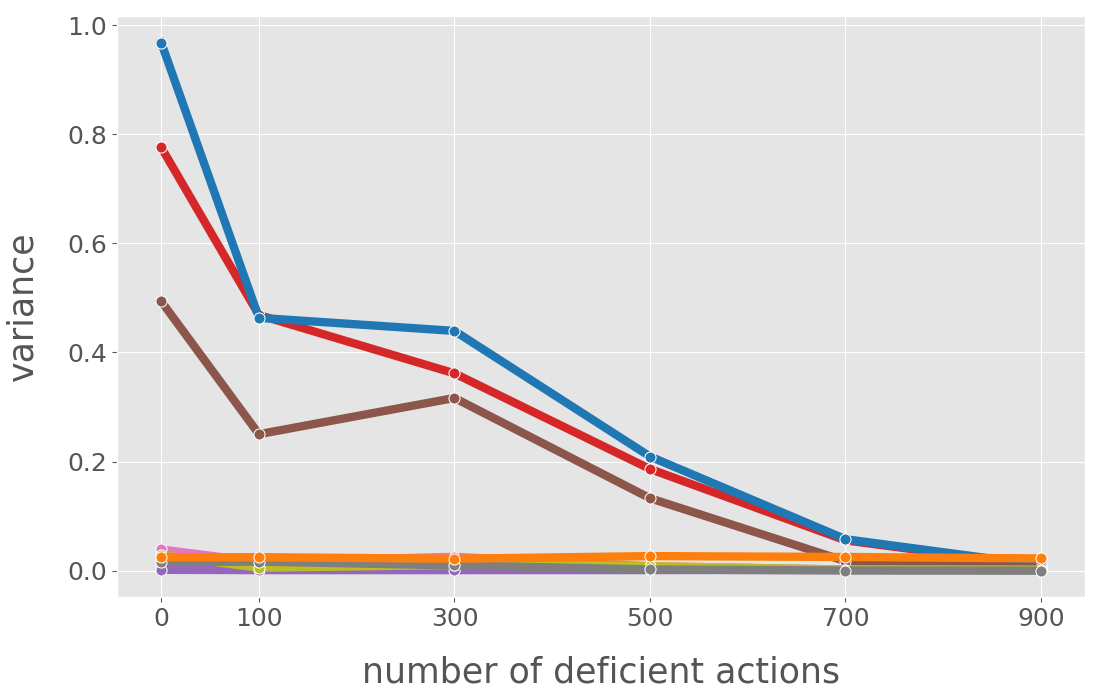}
    \end{center}
\end{minipage}
\\
\multicolumn{3}{c}{
\begin{minipage}{0.95\hsize}
\begin{center}
\caption{MSE, Squared Bias, and Variance \textbf{with varying number of deficient actions}}
\end{center}
\end{minipage}
}
\\ 
\begin{minipage}{0.30\hsize}
    \begin{center}
        \includegraphics[clip, width=5.8cm]{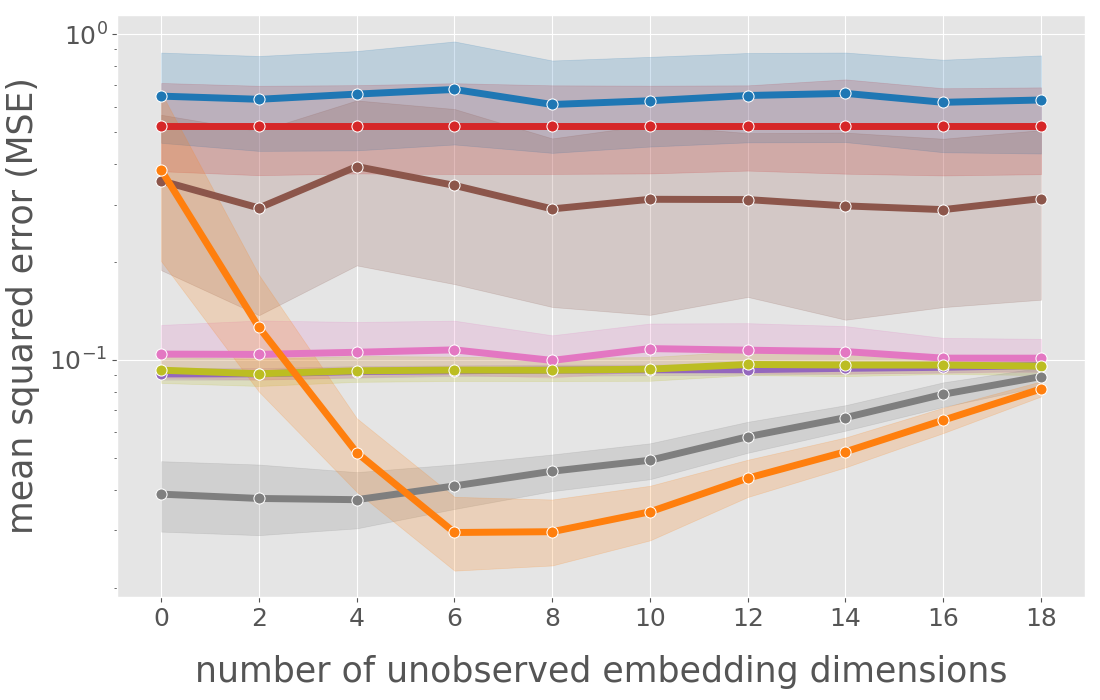}
    \end{center}
\end{minipage}
&
\begin{minipage}{0.30\hsize}
    \begin{center}
        \includegraphics[clip, width=5.8cm]{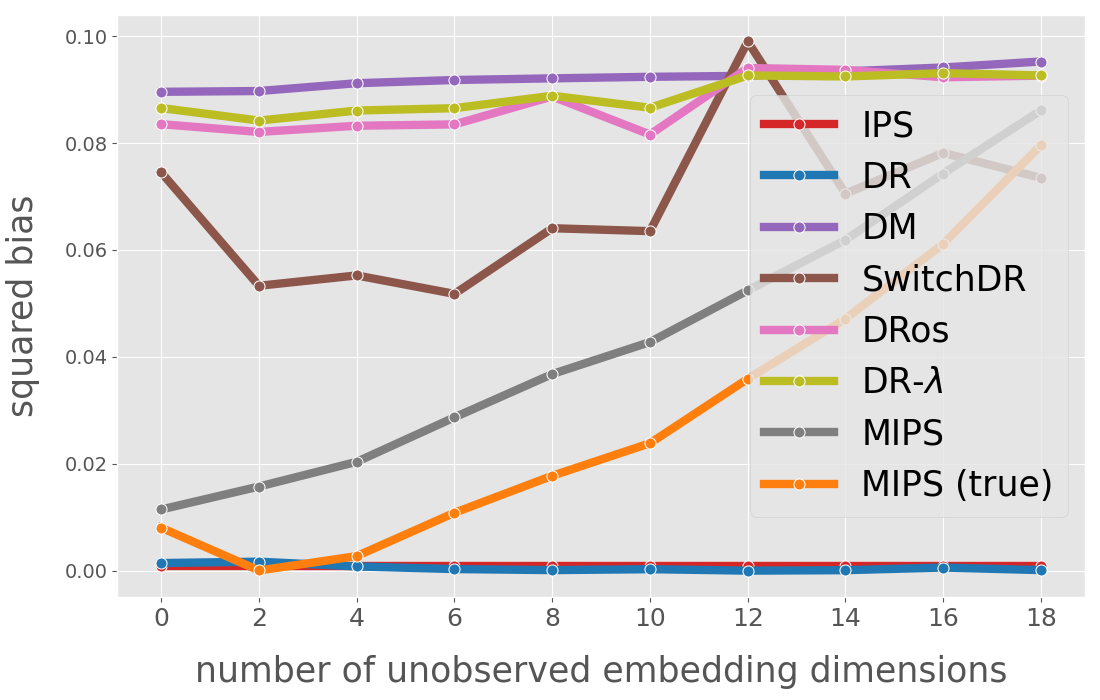}
    \end{center}
\end{minipage}
&
\begin{minipage}{0.30\hsize}
    \begin{center}
        \includegraphics[clip, width=5.8cm]{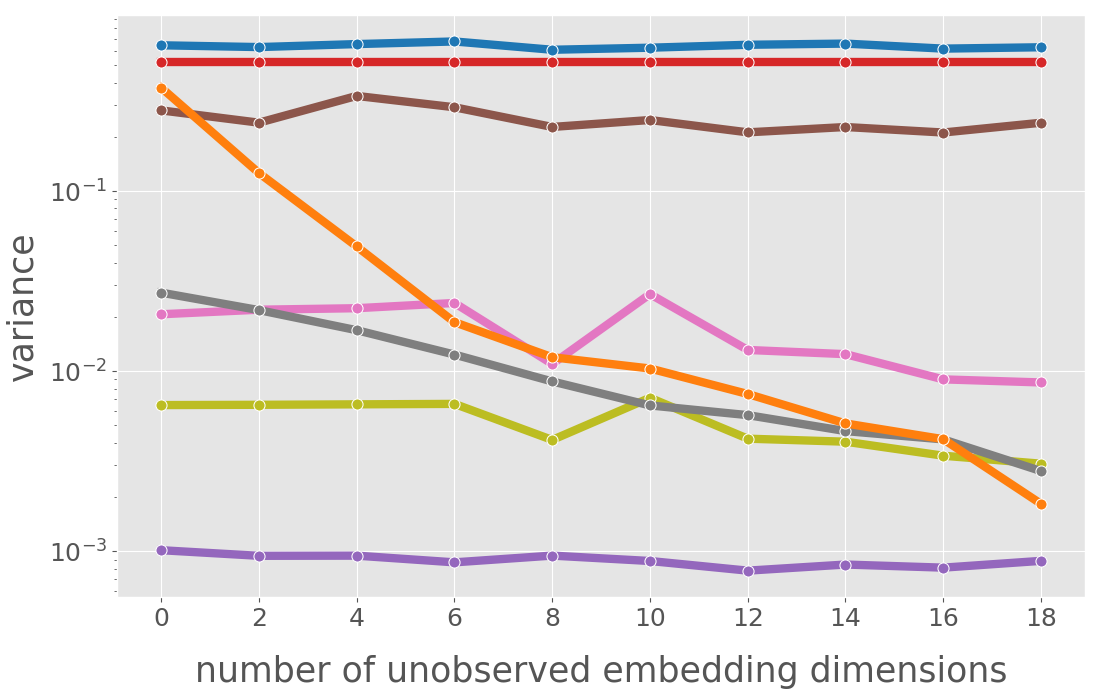}
    \end{center}
\end{minipage}
\\
\multicolumn{3}{c}{
\begin{minipage}{0.95\hsize}
\begin{center}
\caption{MSE, Squared Bias, and Variance with \textbf{varying number of unobserved dimensions in action embeddings}}
\label{fig:beta=0,eps=0.05_end}
\end{center}
\end{minipage}
}
\\ 
\bottomrule
\end{tabular}
}
\vskip 0.1in
\raggedright
\fontsize{9pt}{9pt}\selectfont \textit{Note}:
We set $\beta=0$ and $\epsilon=0.05$, which produce \textbf{uniform random logging policy} and \textbf{near-optimal/near-deterministic target policy}.
The results are averaged over 100 different sets of synthetic logged data replicated with different random seeds.
The shaded regions in the MSE plots represent the 95\% confidence intervals estimated with bootstrap.
The y-axis of MSE and Variance plots (the left and right columns) is reported on log-scale.
\end{figure*}

\begin{figure*}[th]
\scalebox{0.95}{
\begin{tabular}{ccc}
\toprule
\textbf{MSE} & \textbf{Squared Bias} & \textbf{Variance} \\ \midrule \midrule
\begin{minipage}{0.33\hsize}
    \begin{center}
        \includegraphics[clip, width=5.8cm]{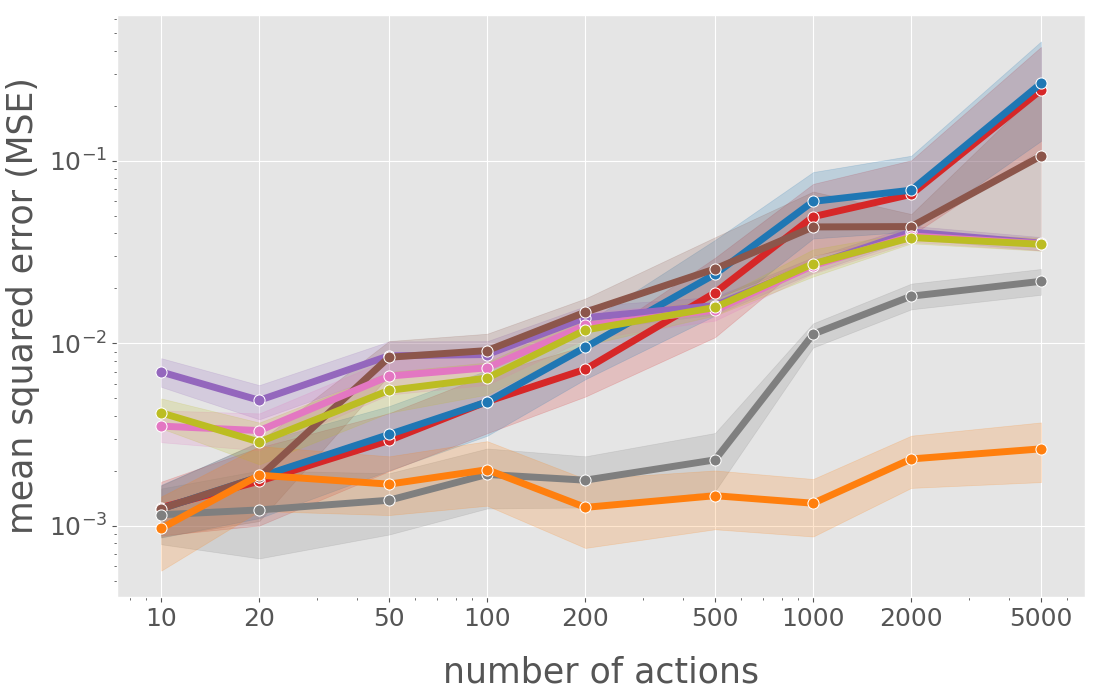}
    \end{center}
\end{minipage}
&
\begin{minipage}{0.33\hsize}
    \begin{center}
        \includegraphics[clip, width=5.8cm]{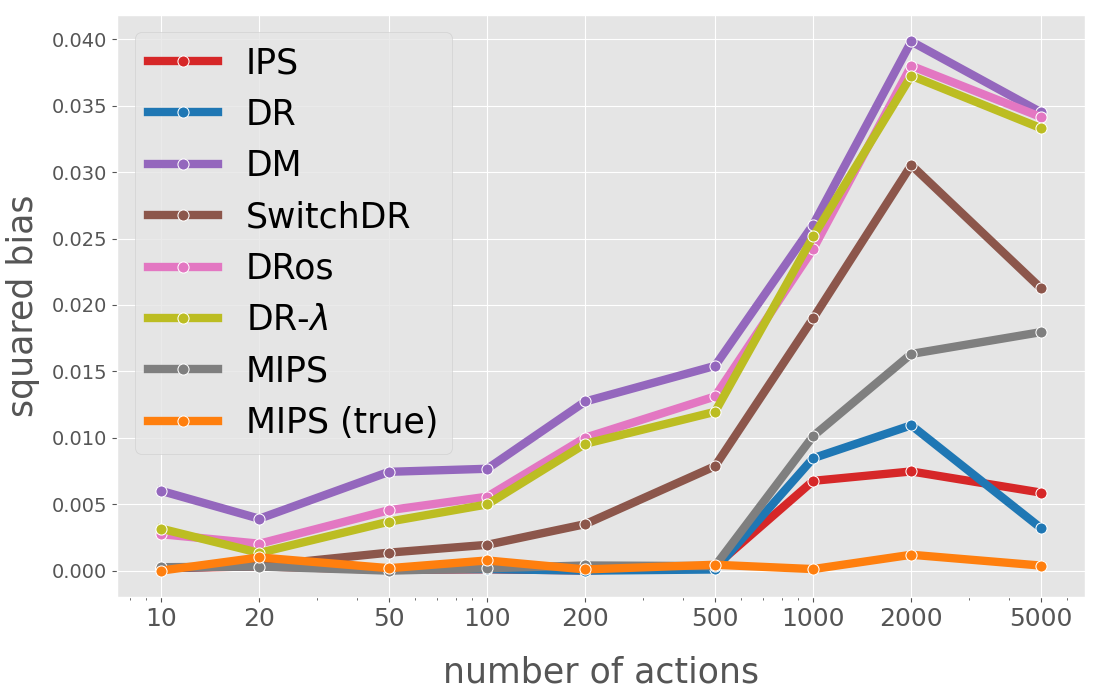}
    \end{center}
\end{minipage}
&
\begin{minipage}{0.33\hsize}
    \begin{center}
        \includegraphics[clip, width=5.8cm]{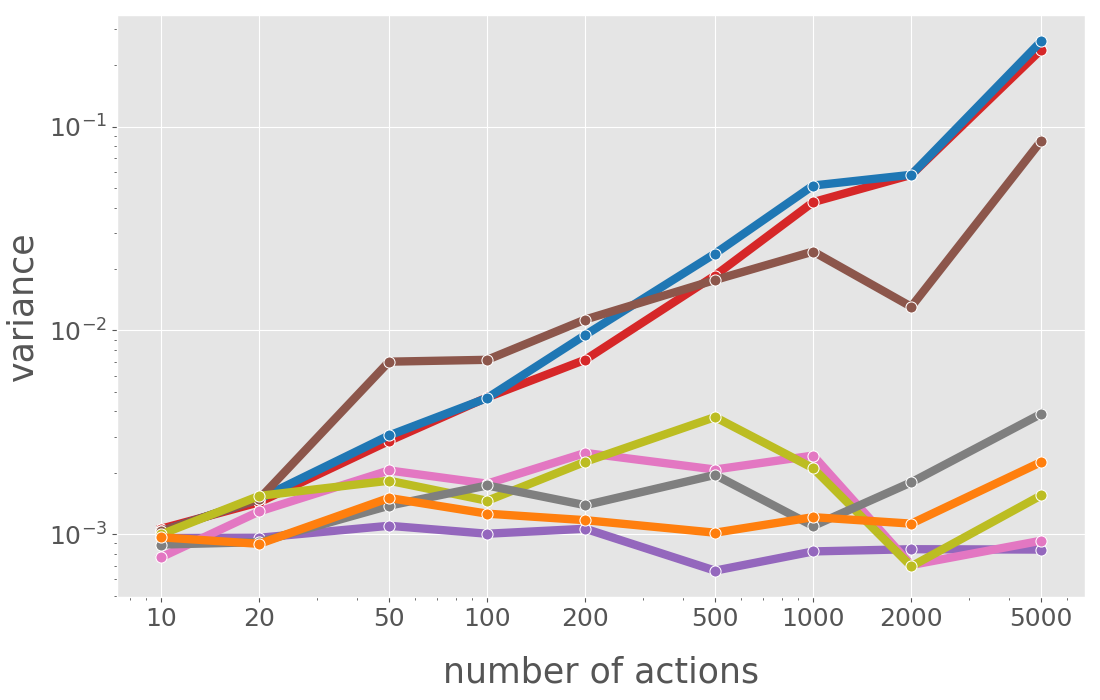}
    \end{center}
\end{minipage}
\\
\multicolumn{3}{c}{
\begin{minipage}{1.0\hsize}
\begin{center}
\caption{MSE, Squared Bias, and Variance with \textbf{varying number of actions}}
\label{fig:beta=0,eps=0.8_start}
\end{center}
\end{minipage}
}
\\
\begin{minipage}{0.33\hsize}
    \begin{center}
        \includegraphics[clip, width=5.8cm]{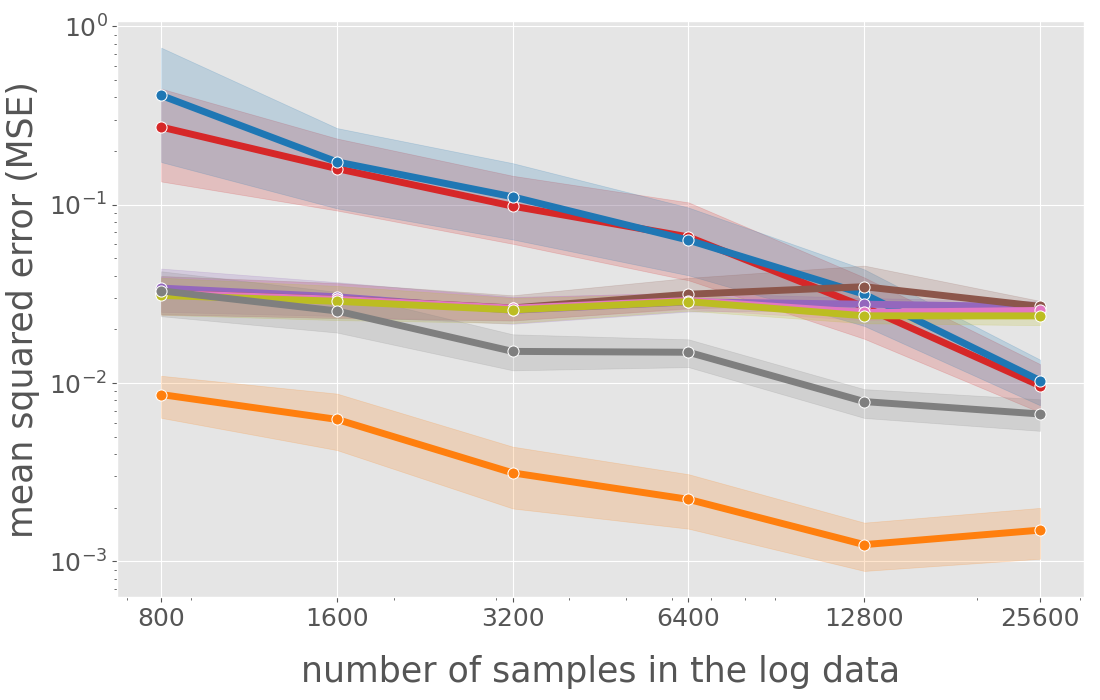}
    \end{center}
\end{minipage}
&
\begin{minipage}{0.33\hsize}
    \begin{center}
        \includegraphics[clip, width=5.8cm]{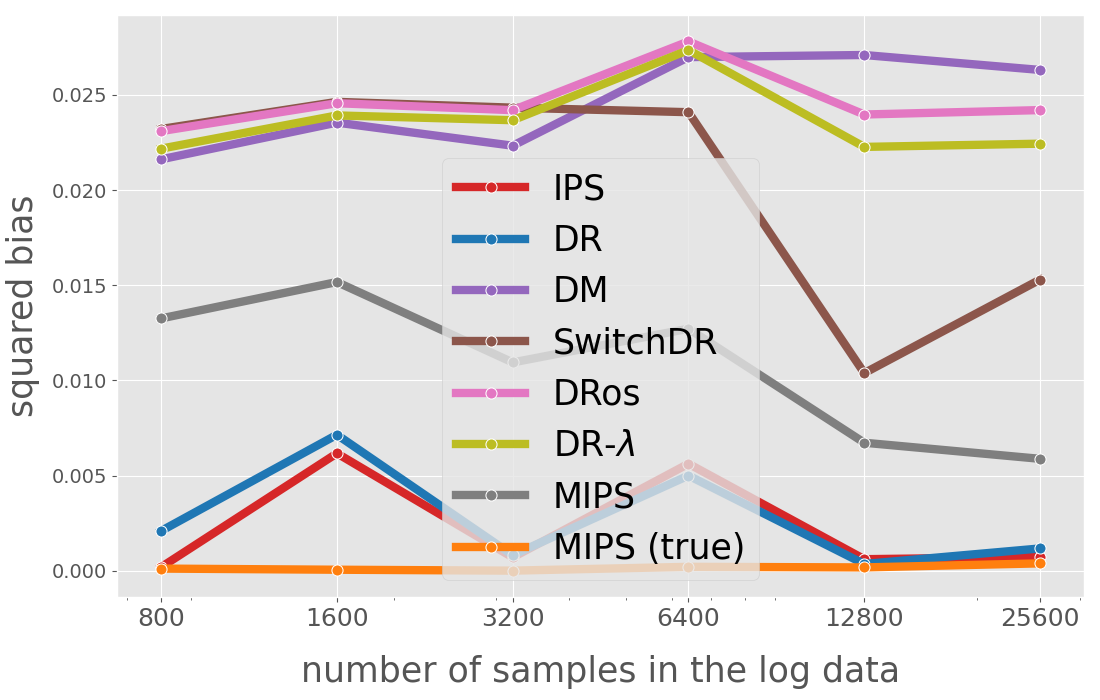}
    \end{center}
\end{minipage}
&
\begin{minipage}{0.33\hsize}
    \begin{center}
        \includegraphics[clip, width=5.8cm]{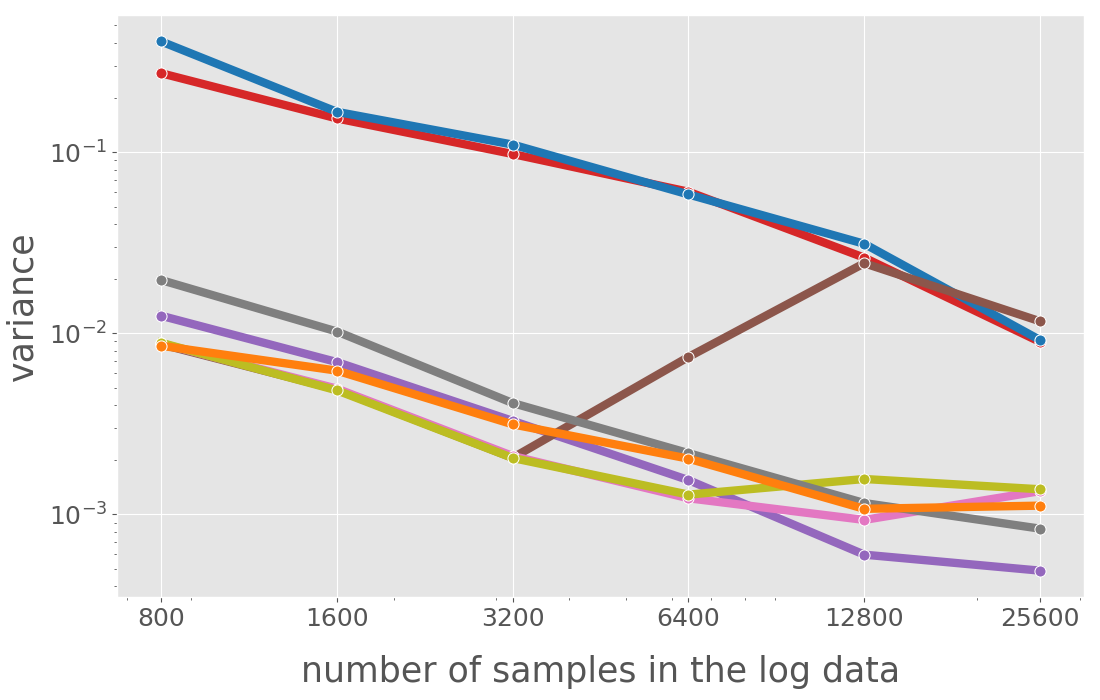}
    \end{center}
\end{minipage}
\\
\multicolumn{3}{c}{
\begin{minipage}{0.95\hsize}
\begin{center}
\caption{MSE, Squared Bias, and Variance with \textbf{varying sample size}}
\end{center}
\end{minipage}
}
\\
\begin{minipage}{0.33\hsize}
    \begin{center}
        \includegraphics[clip, width=5.8cm]{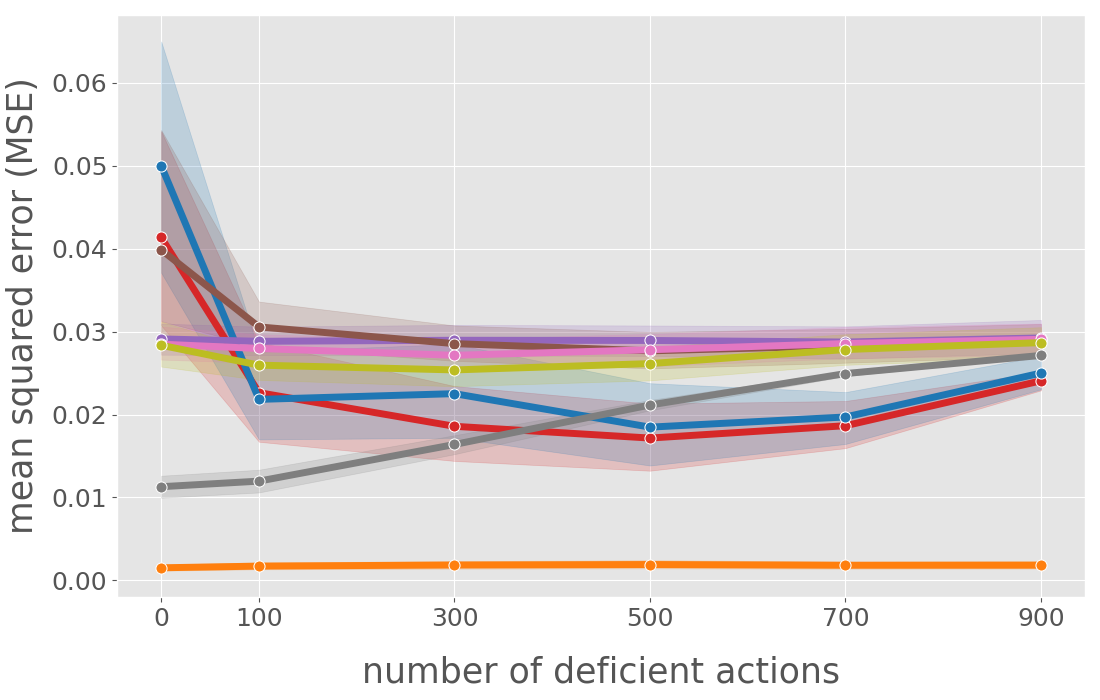}
    \end{center}
\end{minipage}
&
\begin{minipage}{0.33\hsize}
    \begin{center}
        \includegraphics[clip, width=5.8cm]{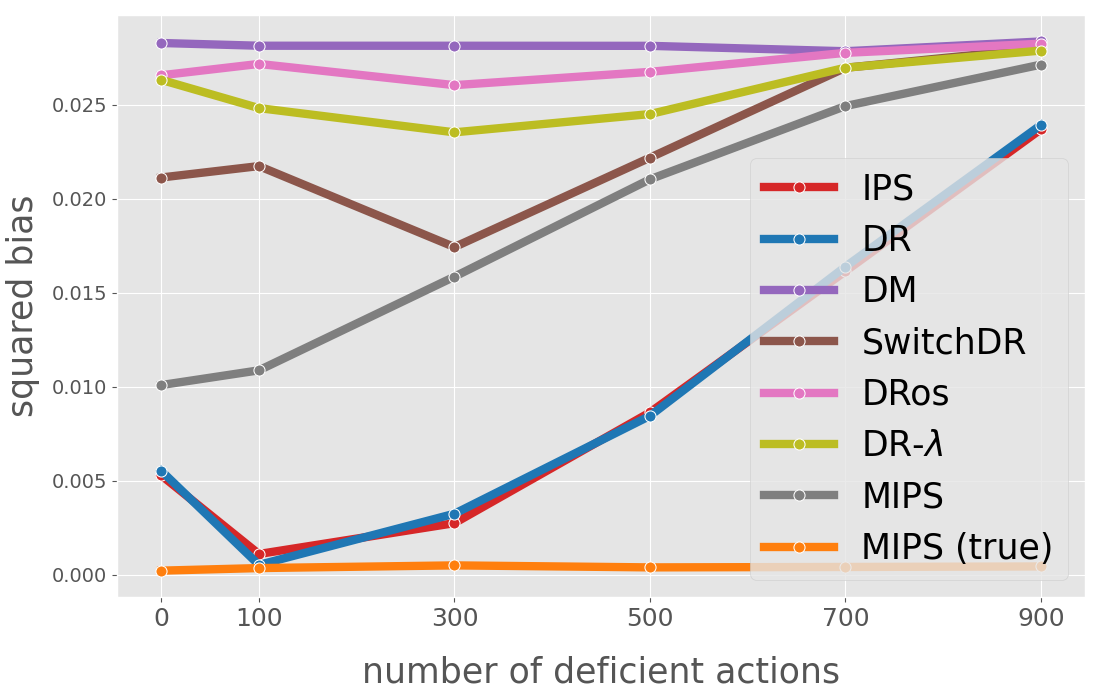}
    \end{center}
\end{minipage}
&
\begin{minipage}{0.33\hsize}
    \begin{center}
        \includegraphics[clip, width=5.8cm]{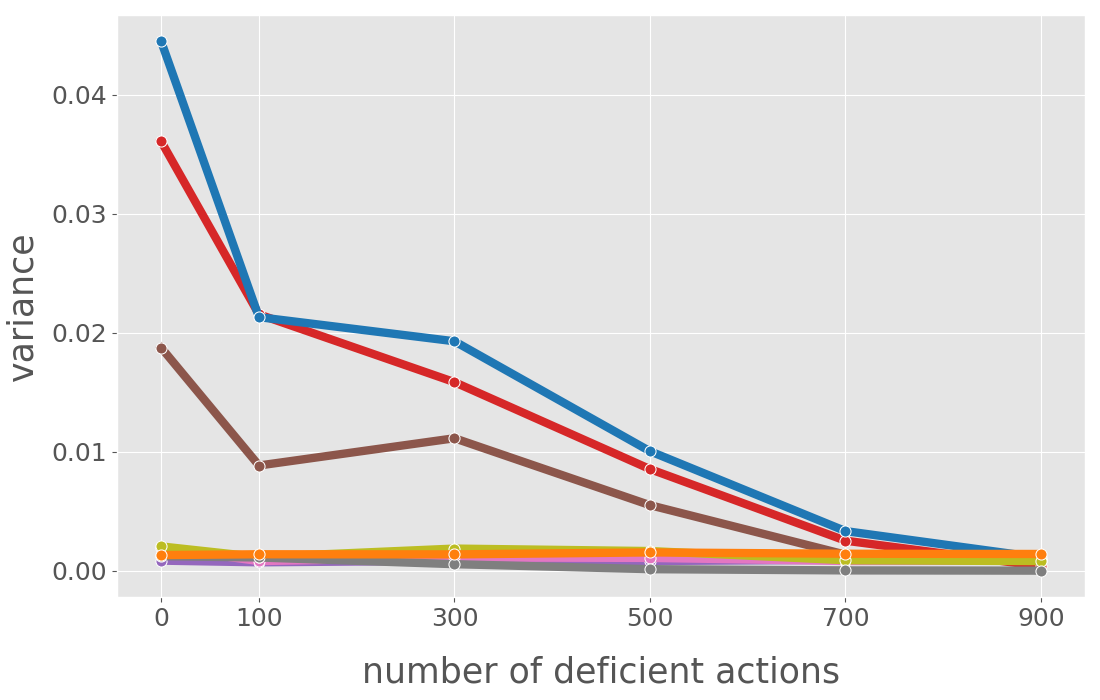}
    \end{center}
\end{minipage}
\\
\multicolumn{3}{c}{
\begin{minipage}{0.95\hsize}
\begin{center}
\caption{MSE, Squared Bias, and Variance \textbf{with varying number of deficient actions}}
\end{center}
\end{minipage}
}
\\ 
\begin{minipage}{0.30\hsize}
    \begin{center}
        \includegraphics[clip, width=5.8cm]{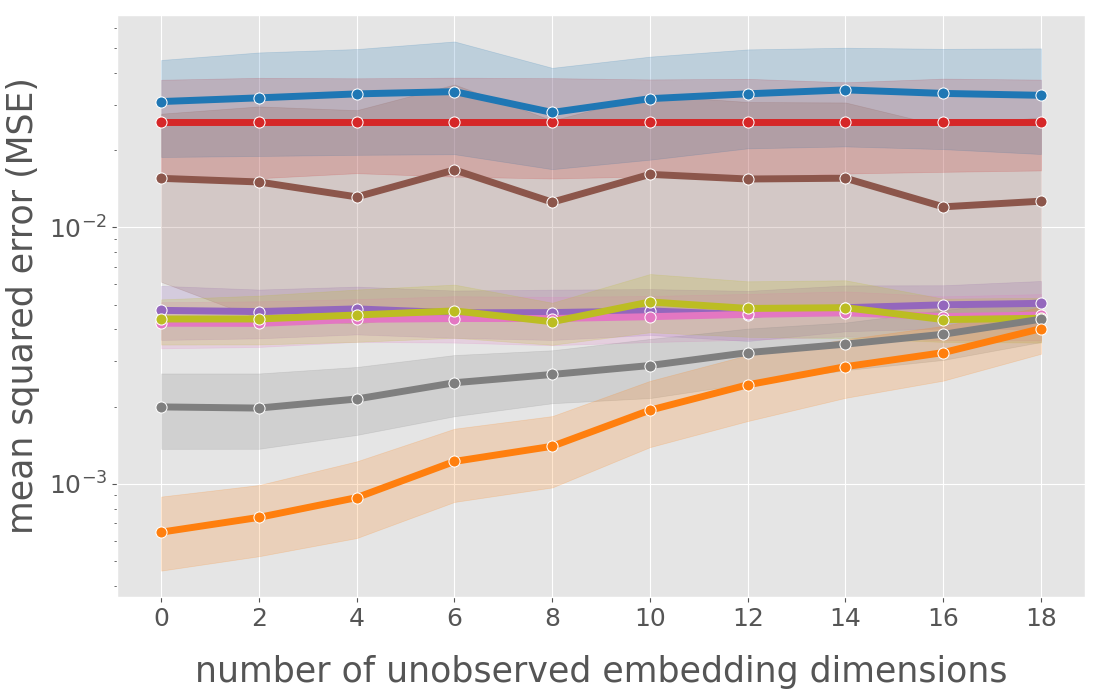}
    \end{center}
\end{minipage}
&
\begin{minipage}{0.30\hsize}
    \begin{center}
        \includegraphics[clip, width=5.8cm]{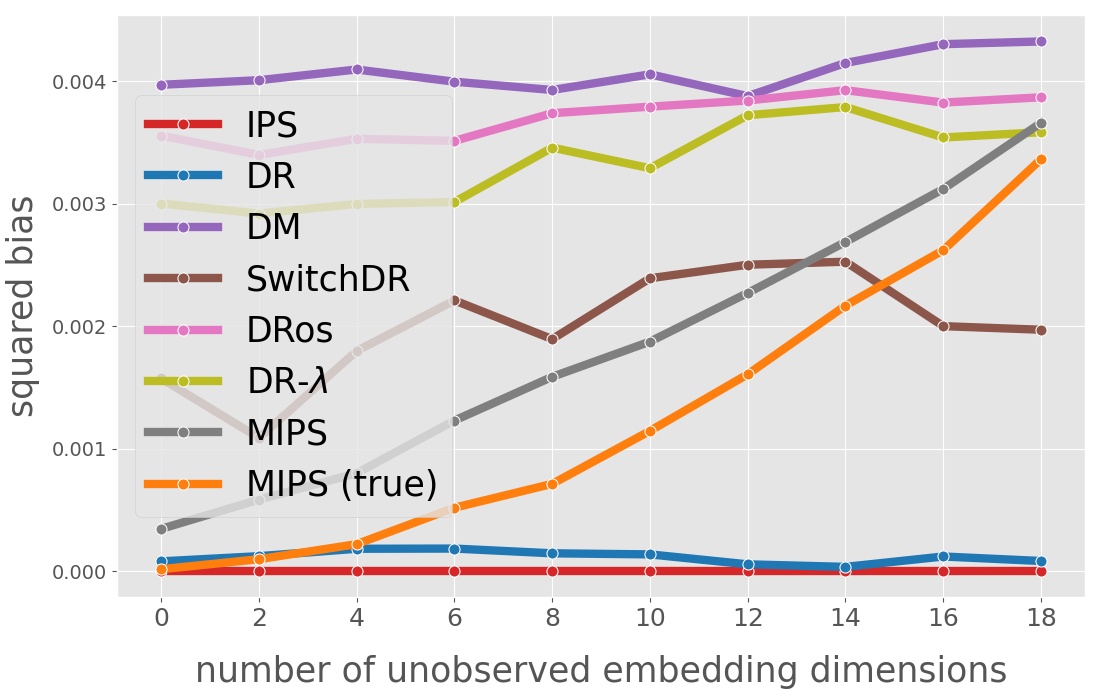}
    \end{center}
\end{minipage}
&
\begin{minipage}{0.30\hsize}
    \begin{center}
        \includegraphics[clip, width=5.8cm]{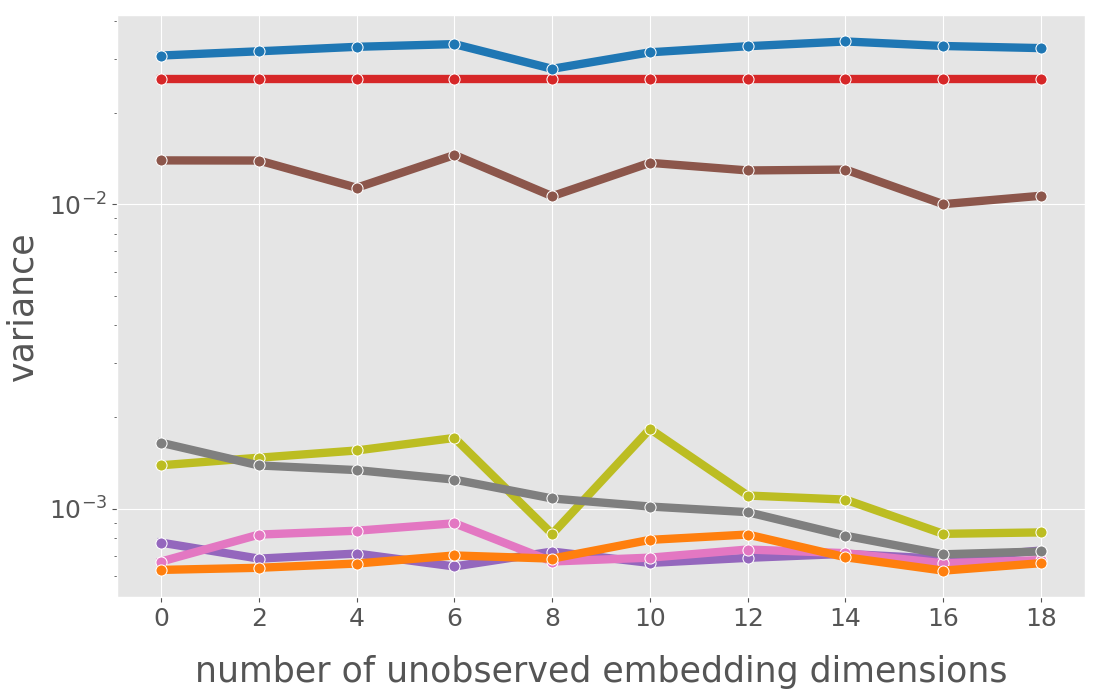}
    \end{center}
\end{minipage}
\\
\multicolumn{3}{c}{
\begin{minipage}{0.95\hsize}
\begin{center}
\caption{MSE, Squared Bias, and Variance with \textbf{varying number of unobserved dimensions in action embeddings}}
\label{fig:beta=0,eps=0.8_end}
\end{center}
\end{minipage}
}
\\ 
\bottomrule
\end{tabular}
}
\vskip 0.1in
\raggedright
\fontsize{9pt}{9pt}\selectfont \textit{Note}:
We set $\beta=0$ and $\epsilon=0.8$, which produce \textbf{uniform random logging policy} and \textbf{near-uniform target policy}.
The results are averaged over 100 different sets of synthetic logged data replicated with different random seeds.
The shaded regions in the MSE plots represent the 95\% confidence intervals estimated with bootstrap.
The y-axis of MSE and Variance plots (the left and right columns) is reported on log-scale.
\end{figure*}

\begin{figure*}[th]
\scalebox{0.95}{
\begin{tabular}{ccc}
\toprule
\textbf{MSE} & \textbf{Squared Bias} & \textbf{Variance} \\ \midrule \midrule
\begin{minipage}{0.33\hsize}
    \begin{center}
        \includegraphics[clip, width=5.8cm]{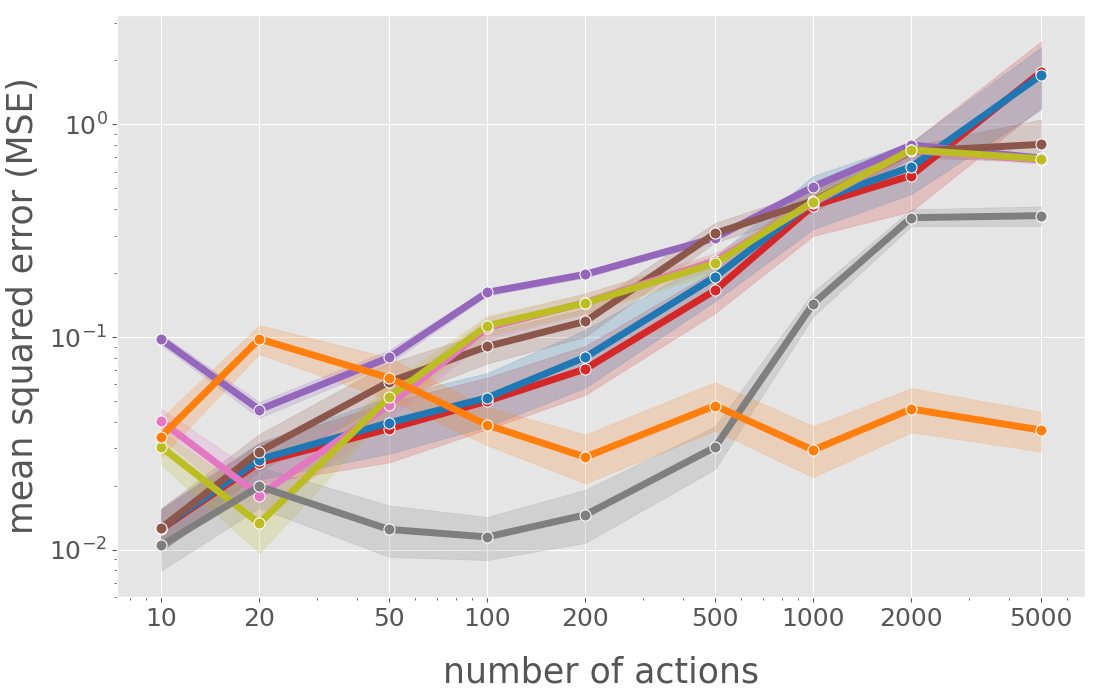}
    \end{center}
\end{minipage}
&
\begin{minipage}{0.33\hsize}
    \begin{center}
        \includegraphics[clip, width=5.8cm]{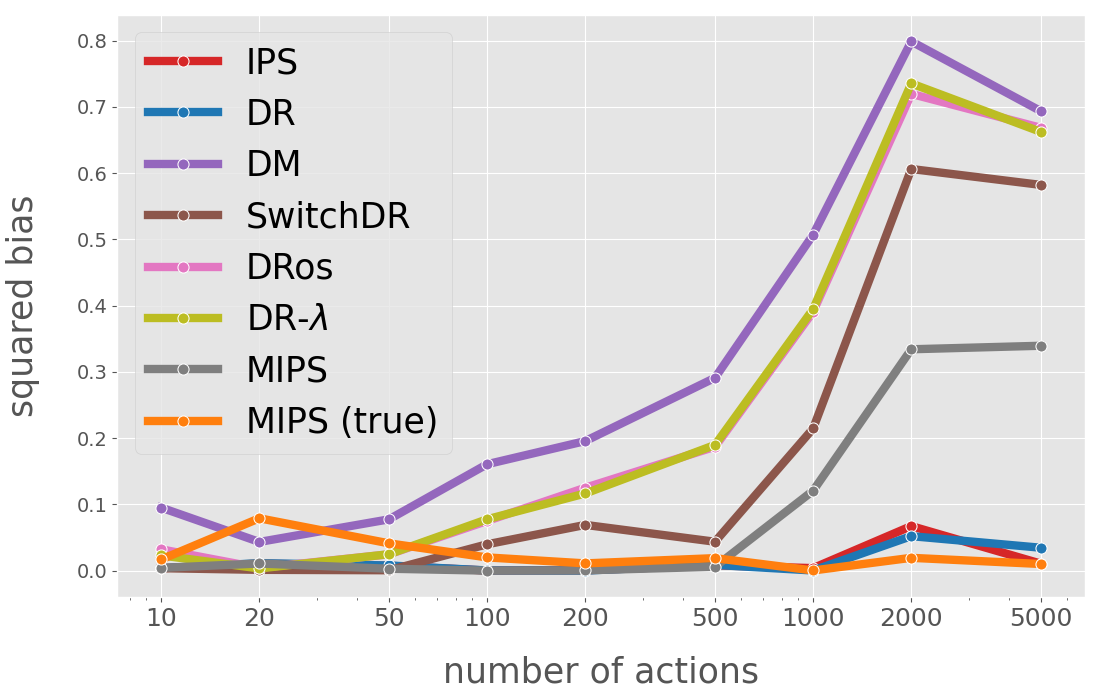}
    \end{center}
\end{minipage}
&
\begin{minipage}{0.33\hsize}
    \begin{center}
        \includegraphics[clip, width=5.8cm]{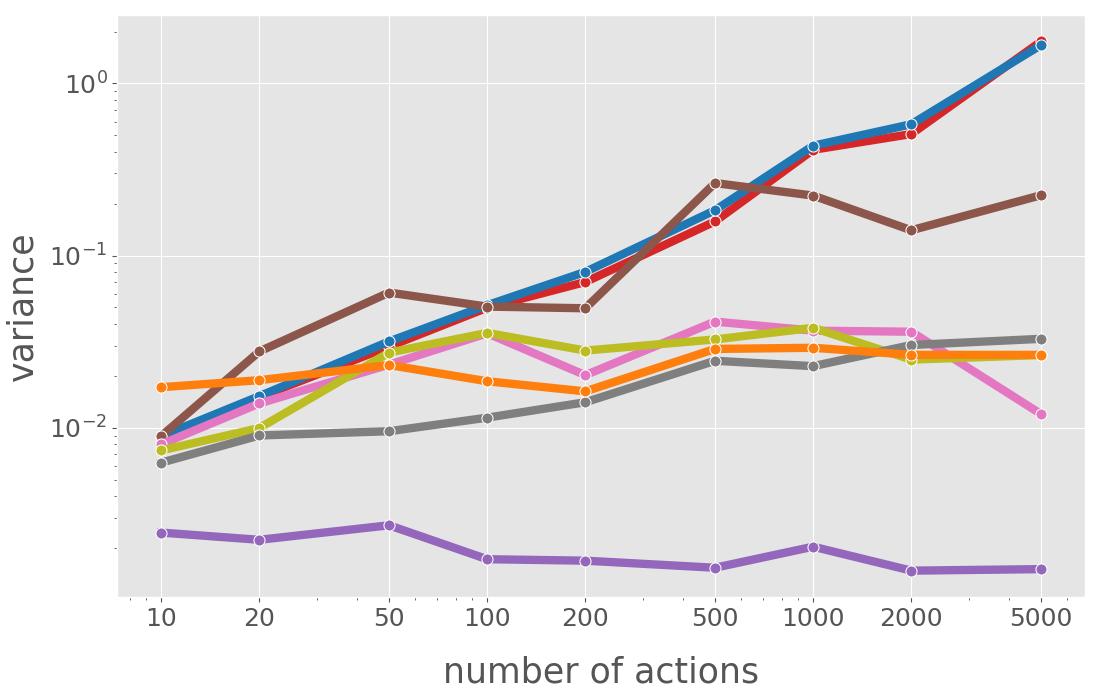}
    \end{center}
\end{minipage}
\\
\multicolumn{3}{c}{
\begin{minipage}{1.0\hsize}
\begin{center}
\caption{MSE, Squared Bias, and Variance with \textbf{varying number of actions}}
\label{fig:beta=1,eps=0.05_start}
\end{center}
\end{minipage}
}
\\
\begin{minipage}{0.33\hsize}
    \begin{center}
        \includegraphics[clip, width=5.8cm]{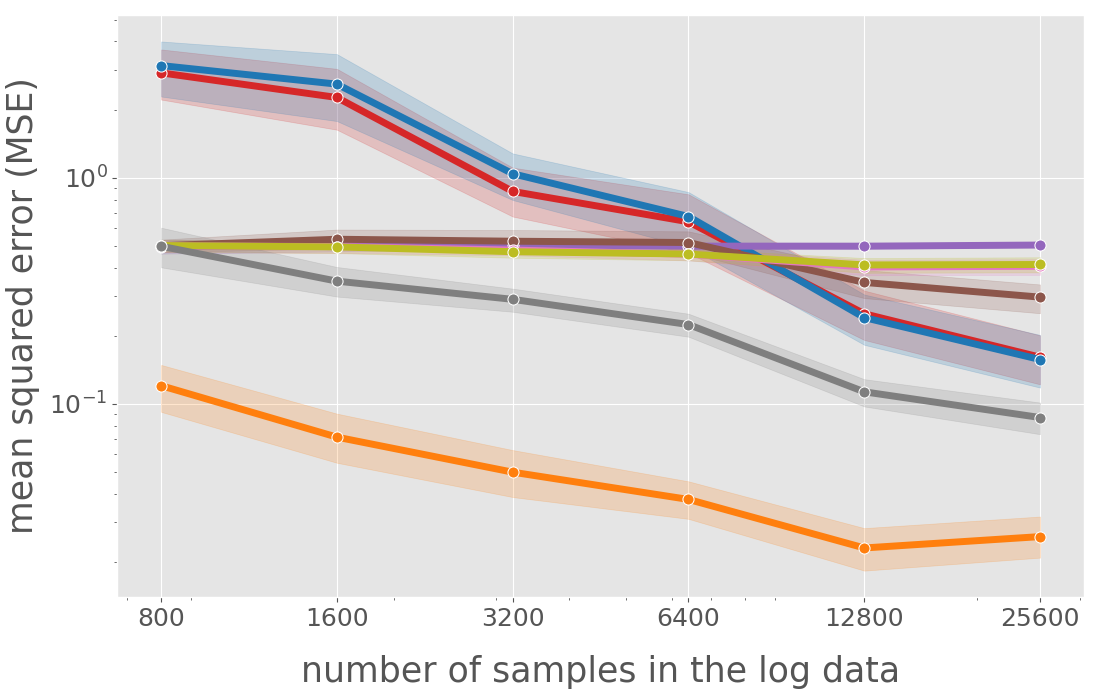}
    \end{center}
\end{minipage}
&
\begin{minipage}{0.33\hsize}
    \begin{center}
        \includegraphics[clip, width=5.8cm]{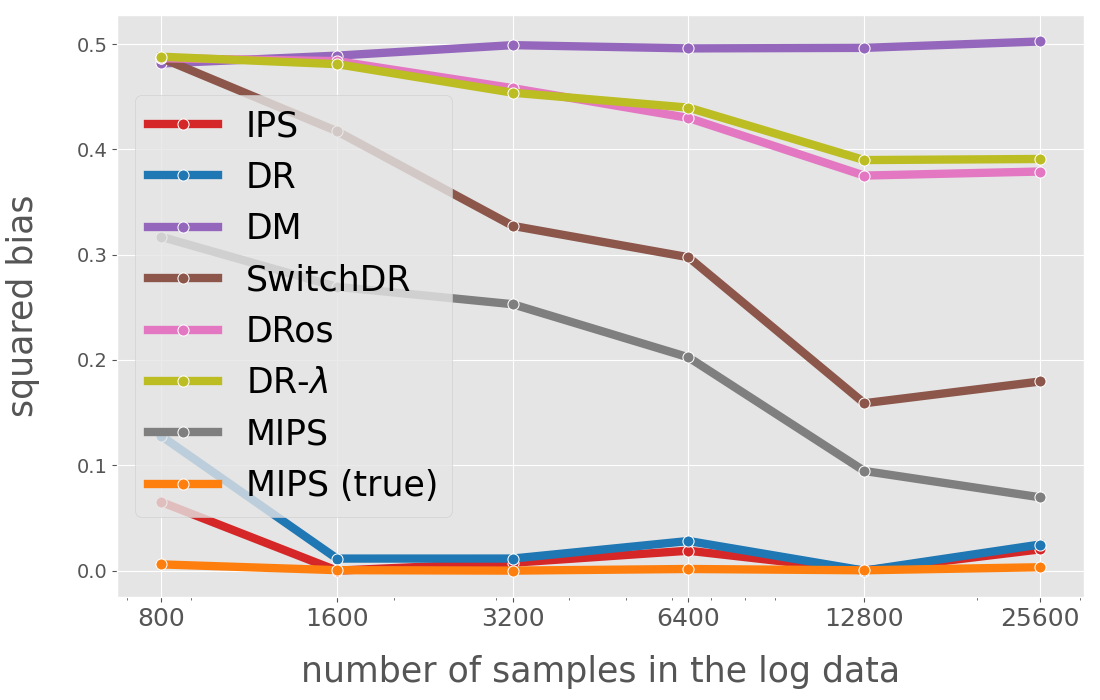}
    \end{center}
\end{minipage}
&
\begin{minipage}{0.33\hsize}
    \begin{center}
        \includegraphics[clip, width=5.8cm]{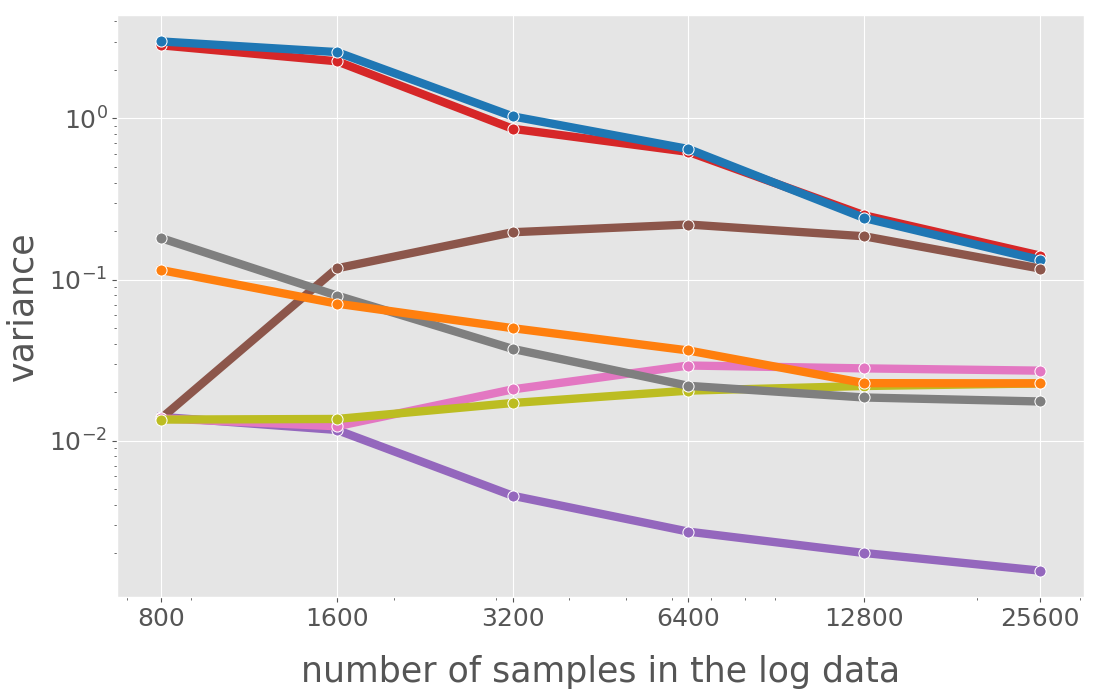}
    \end{center}
\end{minipage}
\\
\multicolumn{3}{c}{
\begin{minipage}{0.95\hsize}
\begin{center}
\caption{MSE, Squared Bias, and Variance with \textbf{varying sample size}}
\end{center}
\end{minipage}
}
\\
\begin{minipage}{0.33\hsize}
    \begin{center}
        \includegraphics[clip, width=5.8cm]{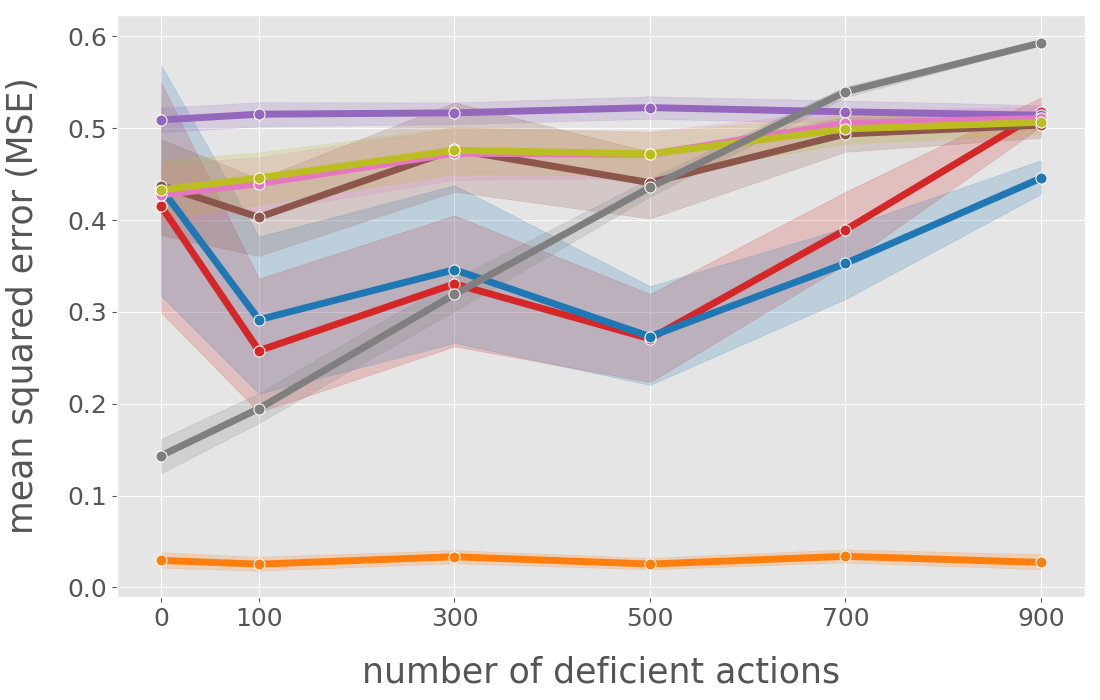}
    \end{center}
\end{minipage}
&
\begin{minipage}{0.33\hsize}
    \begin{center}
        \includegraphics[clip, width=5.8cm]{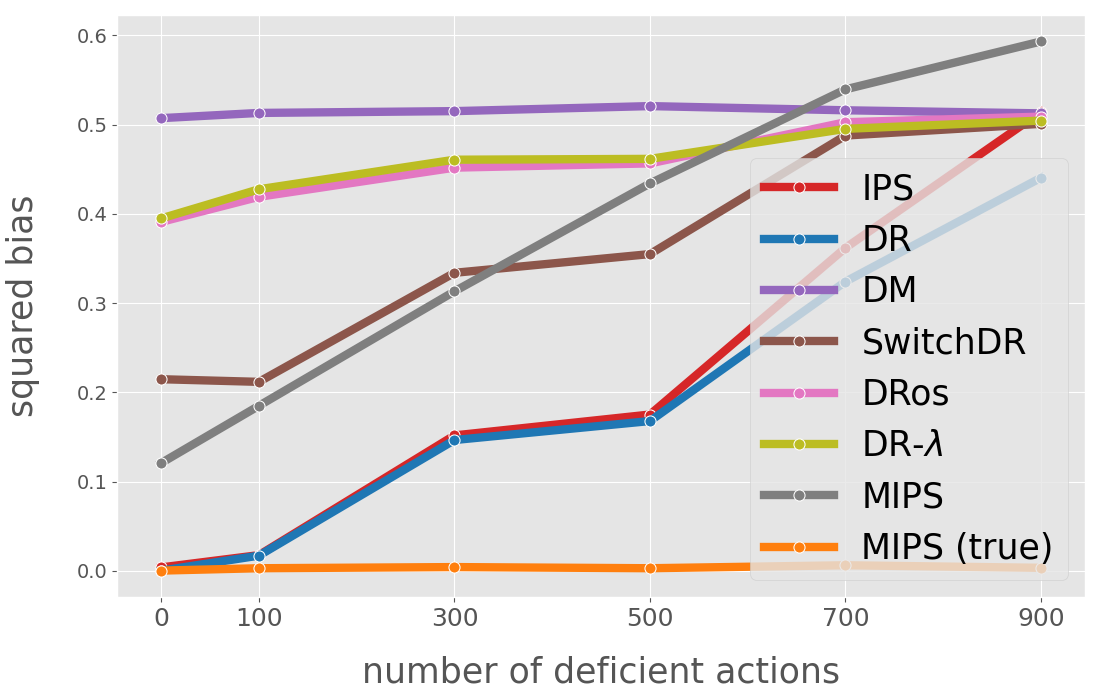}
    \end{center}
\end{minipage}
&
\begin{minipage}{0.33\hsize}
    \begin{center}
        \includegraphics[clip, width=5.8cm]{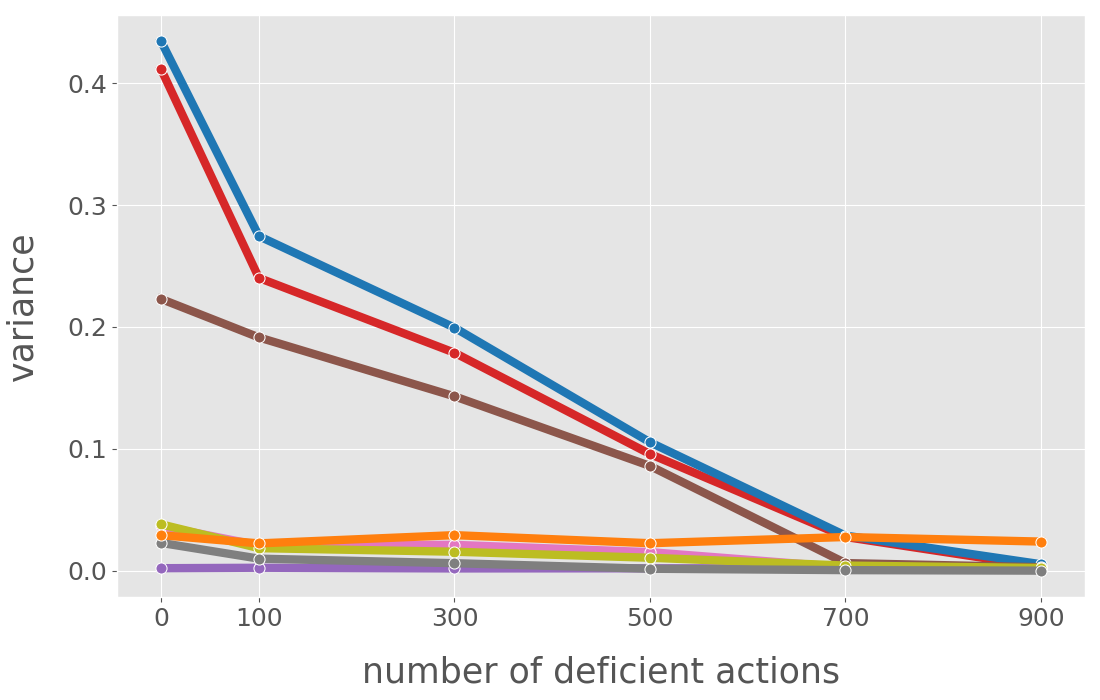}
    \end{center}
\end{minipage}
\\
\multicolumn{3}{c}{
\begin{minipage}{0.95\hsize}
\begin{center}
\caption{MSE, Squared Bias, and Variance \textbf{with varying number of deficient actions}}
\end{center}
\end{minipage}
}
\\ 
\begin{minipage}{0.30\hsize}
    \begin{center}
        \includegraphics[clip, width=5.8cm]{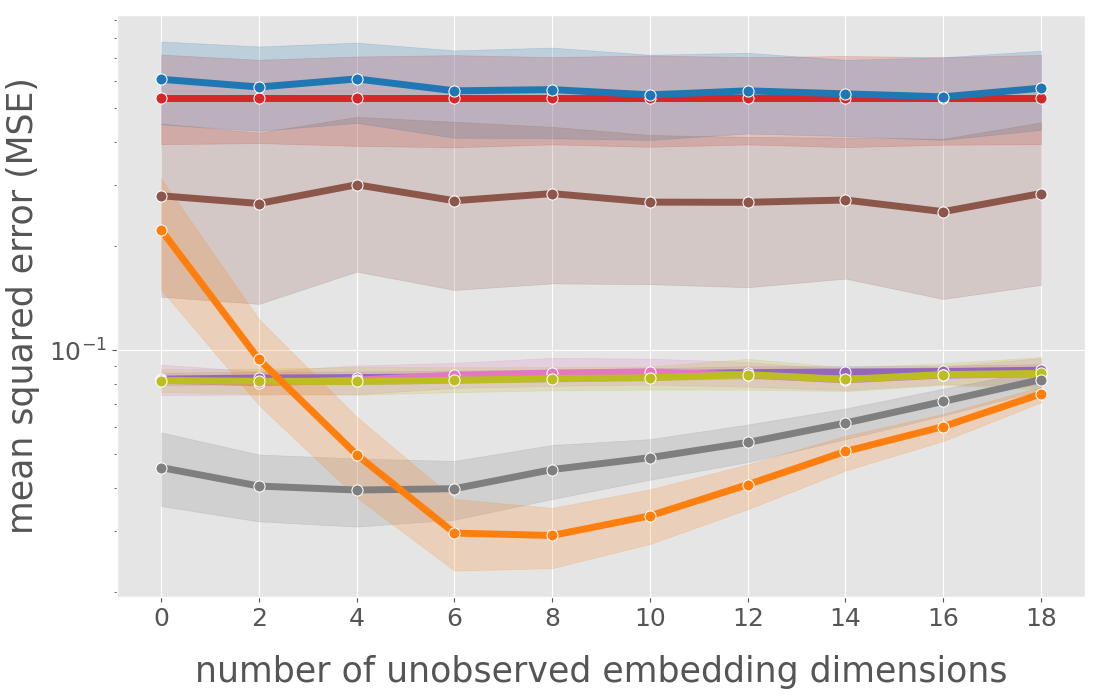}
    \end{center}
\end{minipage}
&
\begin{minipage}{0.30\hsize}
    \begin{center}
        \includegraphics[clip, width=5.8cm]{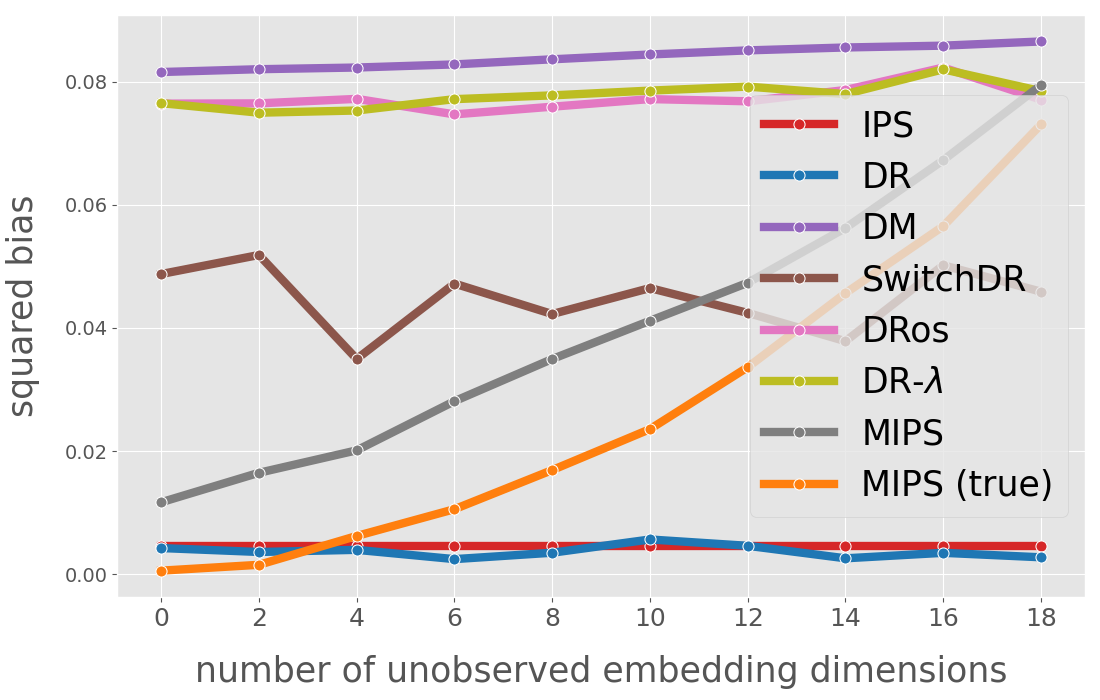}
    \end{center}
\end{minipage}
&
\begin{minipage}{0.30\hsize}
    \begin{center}
        \includegraphics[clip, width=5.8cm]{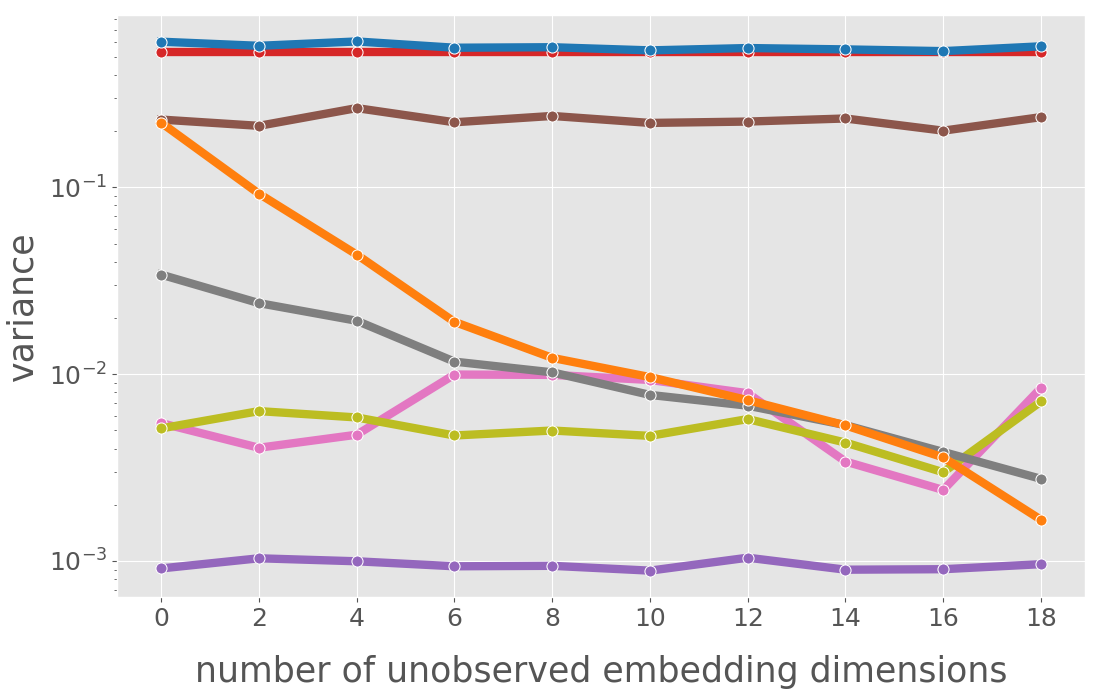}
    \end{center}
\end{minipage}
\\
\multicolumn{3}{c}{
\begin{minipage}{0.95\hsize}
\begin{center}
\caption{MSE, Squared Bias, and Variance with \textbf{varying number of unobserved dimensions in action embeddings}}
\label{fig:beta=1,eps=0.05_end}
\end{center}
\end{minipage}
}
\\ 
\bottomrule
\end{tabular}
}
\vskip 0.1in
\raggedright
\fontsize{9pt}{9pt}\selectfont \textit{Note}:
We set $\beta=1$ and $\epsilon=0.05$, which produce \textbf{logging policy slightly better than uniform random} and \textbf{near-optimal/near-deterministic target policy}.
The results are averaged over 100 different sets of synthetic logged data replicated with different random seeds.
The shaded regions in the MSE plots represent the 95\% confidence intervals estimated with bootstrap.
The y-axis of MSE and Variance plots (the left and right columns) is reported on log-scale.
\end{figure*}

\begin{figure*}[th]
\scalebox{0.95}{
\begin{tabular}{ccc}
\toprule
\textbf{MSE} & \textbf{Squared Bias} & \textbf{Variance} \\ \midrule \midrule
\begin{minipage}{0.33\hsize}
    \begin{center}
        \includegraphics[clip, width=5.8cm]{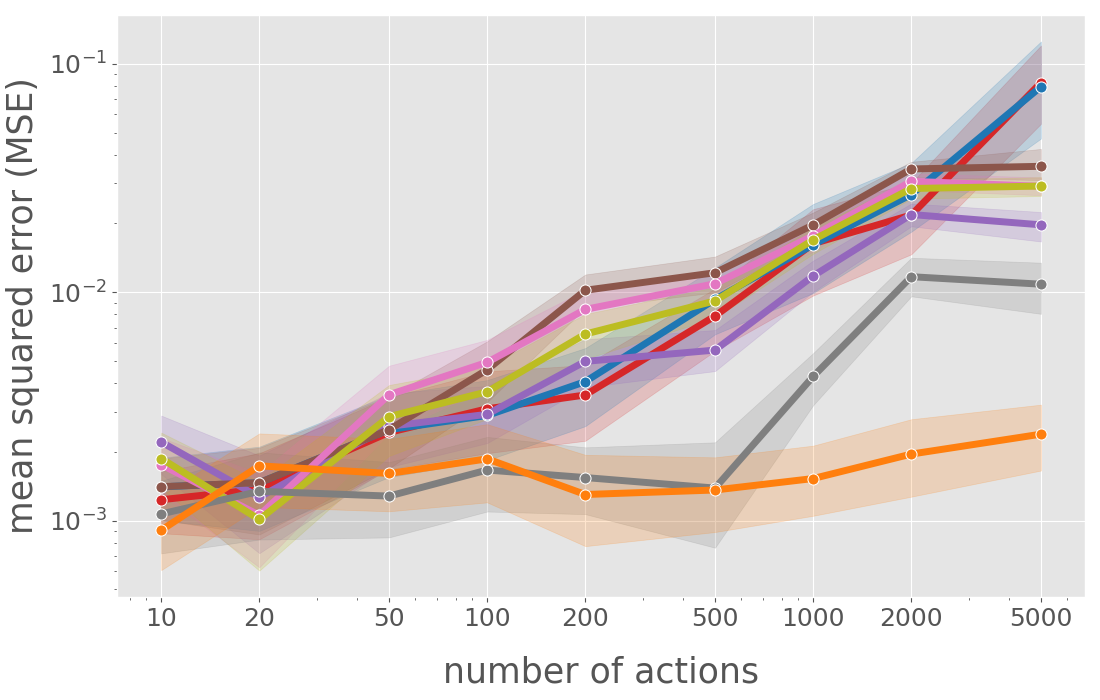}
    \end{center}
\end{minipage}
&
\begin{minipage}{0.33\hsize}
    \begin{center}
        \includegraphics[clip, width=5.8cm]{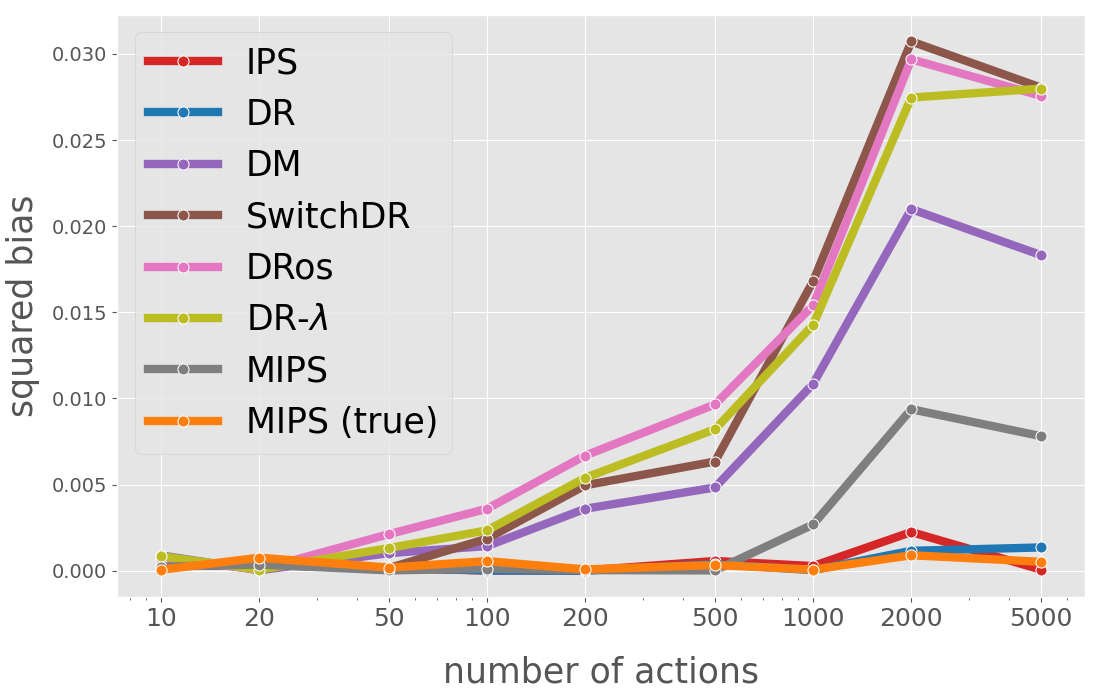}
    \end{center}
\end{minipage}
&
\begin{minipage}{0.33\hsize}
    \begin{center}
        \includegraphics[clip, width=5.8cm]{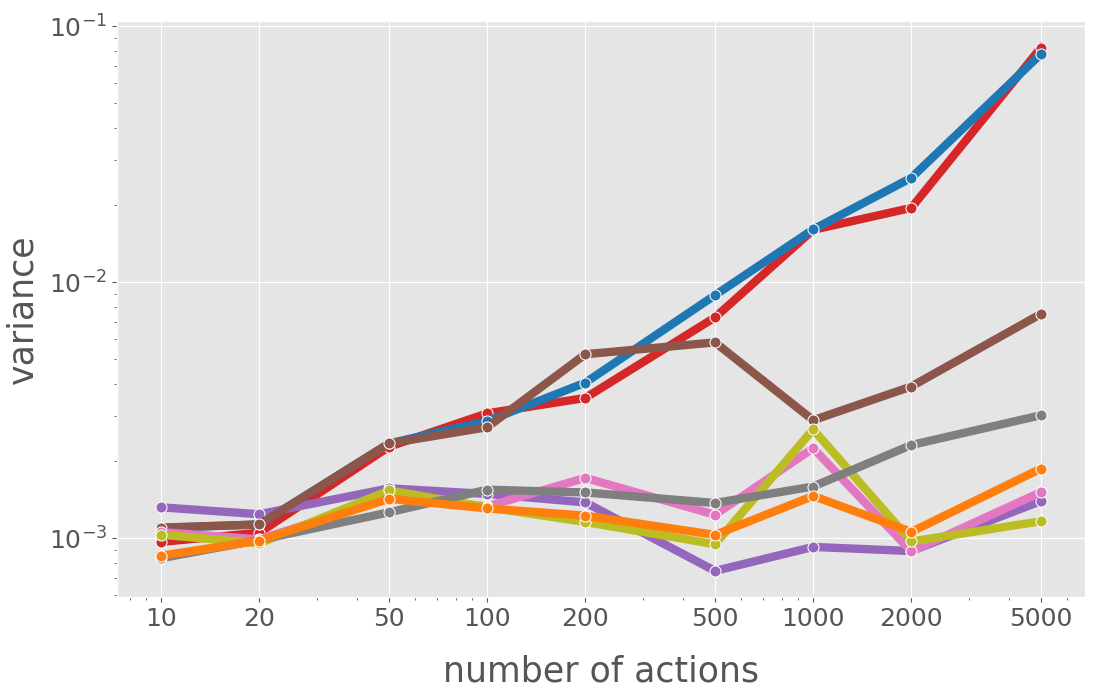}
    \end{center}
\end{minipage}
\\
\multicolumn{3}{c}{
\begin{minipage}{1.0\hsize}
\begin{center}
\caption{MSE, Squared Bias, and Variance with \textbf{varying number of actions}}
\label{fig:beta=1,eps=0.8_start}
\end{center}
\end{minipage}
}
\\
\begin{minipage}{0.33\hsize}
    \begin{center}
        \includegraphics[clip, width=5.8cm]{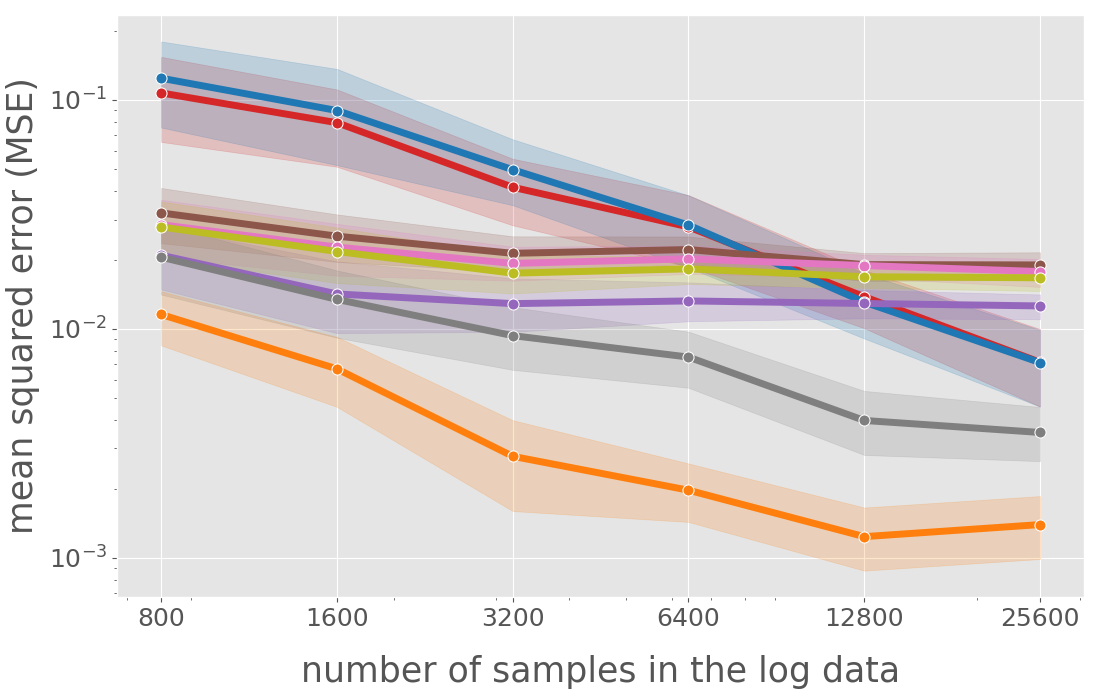}
    \end{center}
\end{minipage}
&
\begin{minipage}{0.33\hsize}
    \begin{center}
        \includegraphics[clip, width=5.8cm]{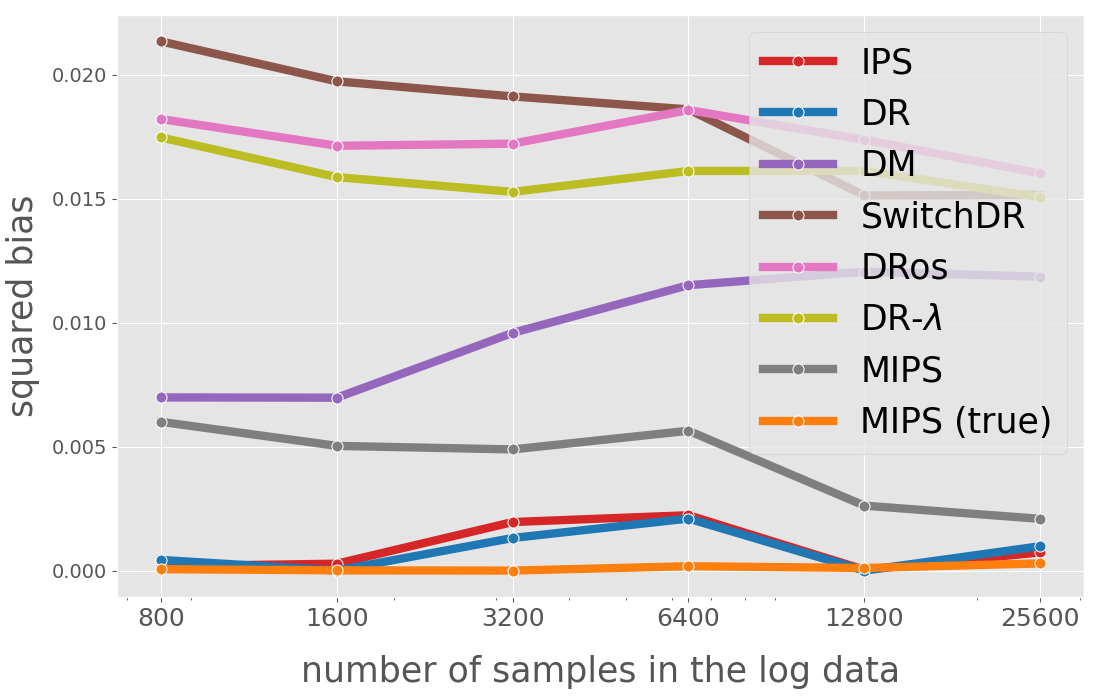}
    \end{center}
\end{minipage}
&
\begin{minipage}{0.33\hsize}
    \begin{center}
        \includegraphics[clip, width=5.8cm]{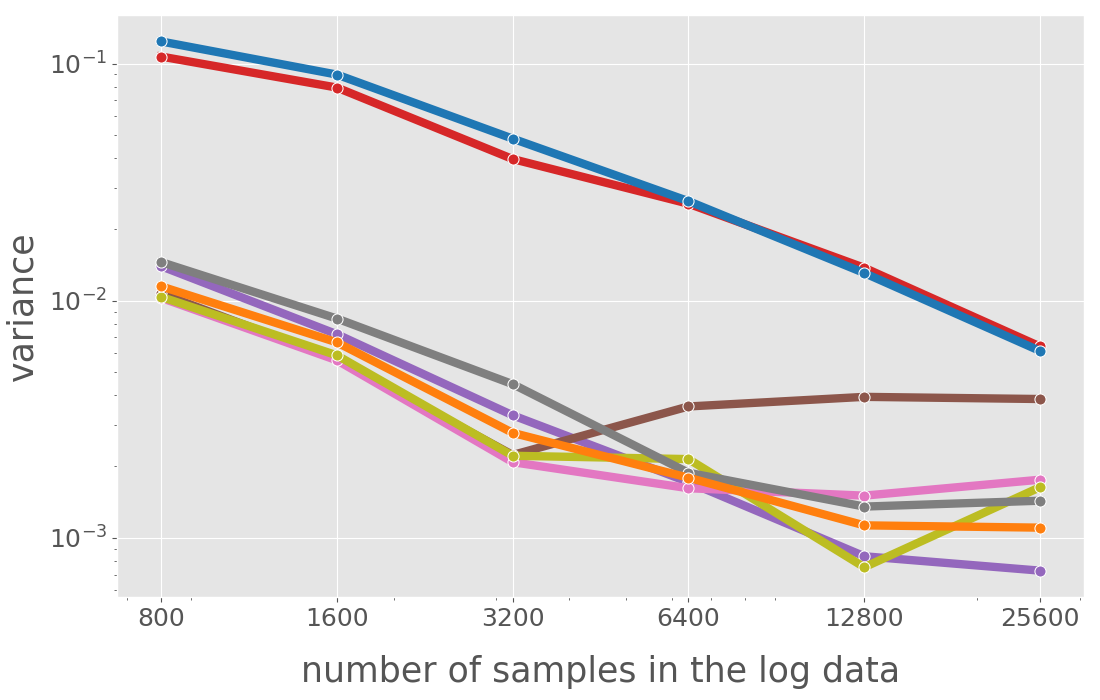}
    \end{center}
\end{minipage}
\\
\multicolumn{3}{c}{
\begin{minipage}{0.95\hsize}
\begin{center}
\caption{MSE, Squared Bias, and Variance with \textbf{varying sample size}}
\end{center}
\end{minipage}
}
\\
\begin{minipage}{0.33\hsize}
    \begin{center}
        \includegraphics[clip, width=5.8cm]{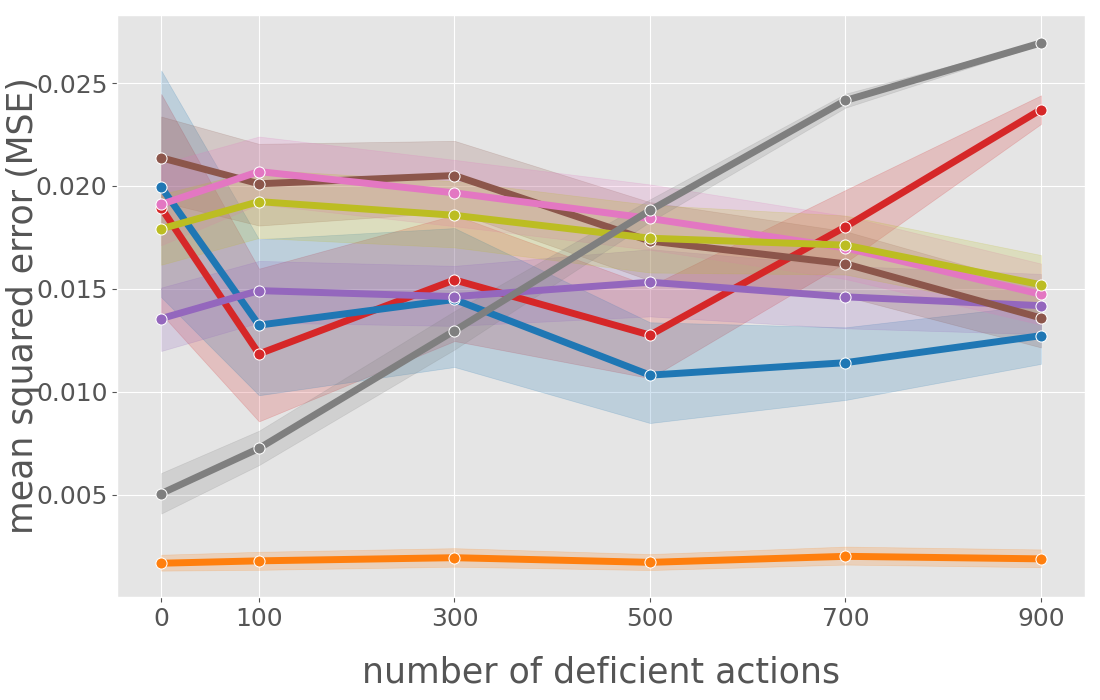}
    \end{center}
\end{minipage}
&
\begin{minipage}{0.33\hsize}
    \begin{center}
        \includegraphics[clip, width=5.8cm]{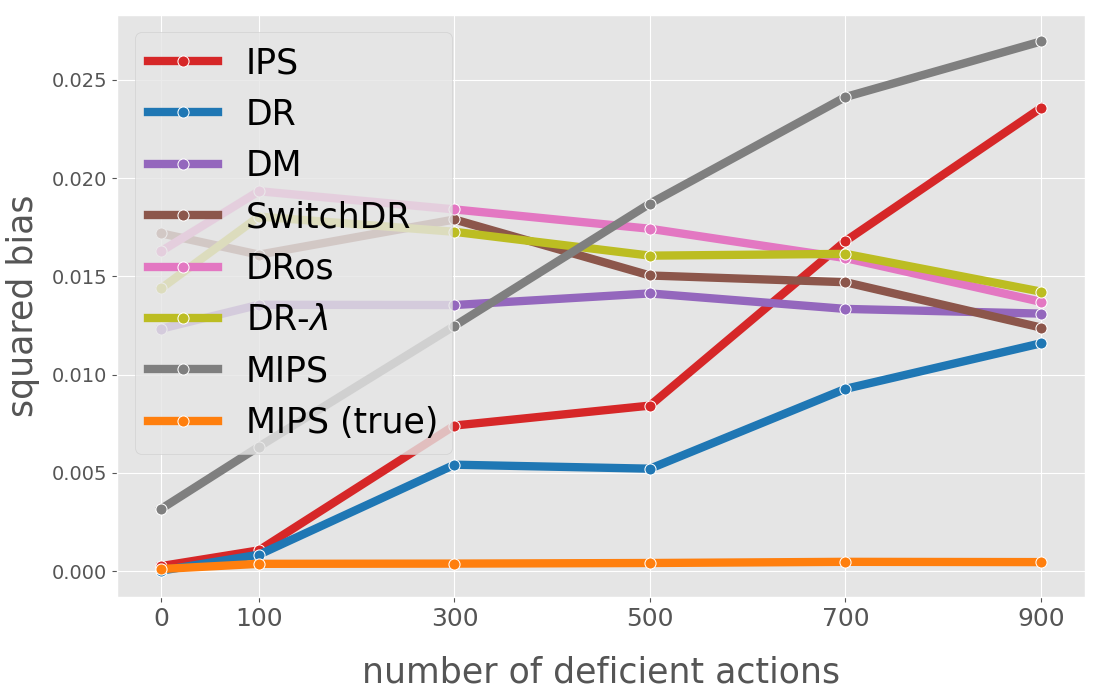}
    \end{center}
\end{minipage}
&
\begin{minipage}{0.33\hsize}
    \begin{center}
        \includegraphics[clip, width=5.8cm]{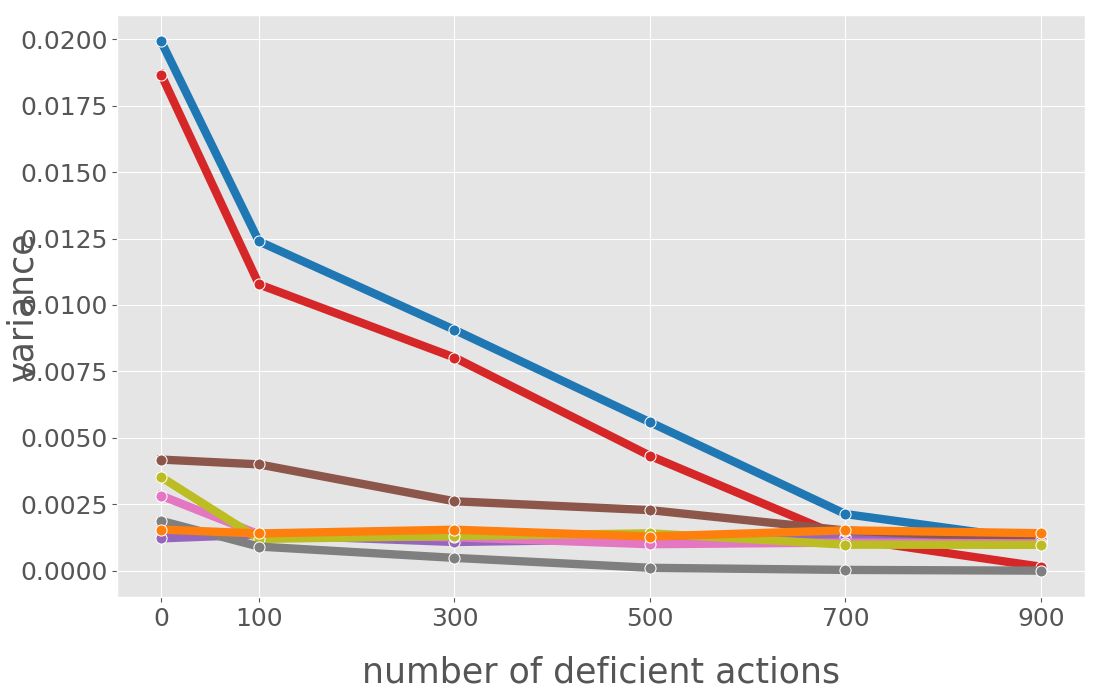}
    \end{center}
\end{minipage}
\\
\multicolumn{3}{c}{
\begin{minipage}{0.95\hsize}
\begin{center}
\caption{MSE, Squared Bias, and Variance \textbf{with varying number of deficient actions}}
\end{center}
\end{minipage}
}
\\ 
\begin{minipage}{0.30\hsize}
    \begin{center}
        \includegraphics[clip, width=5.8cm]{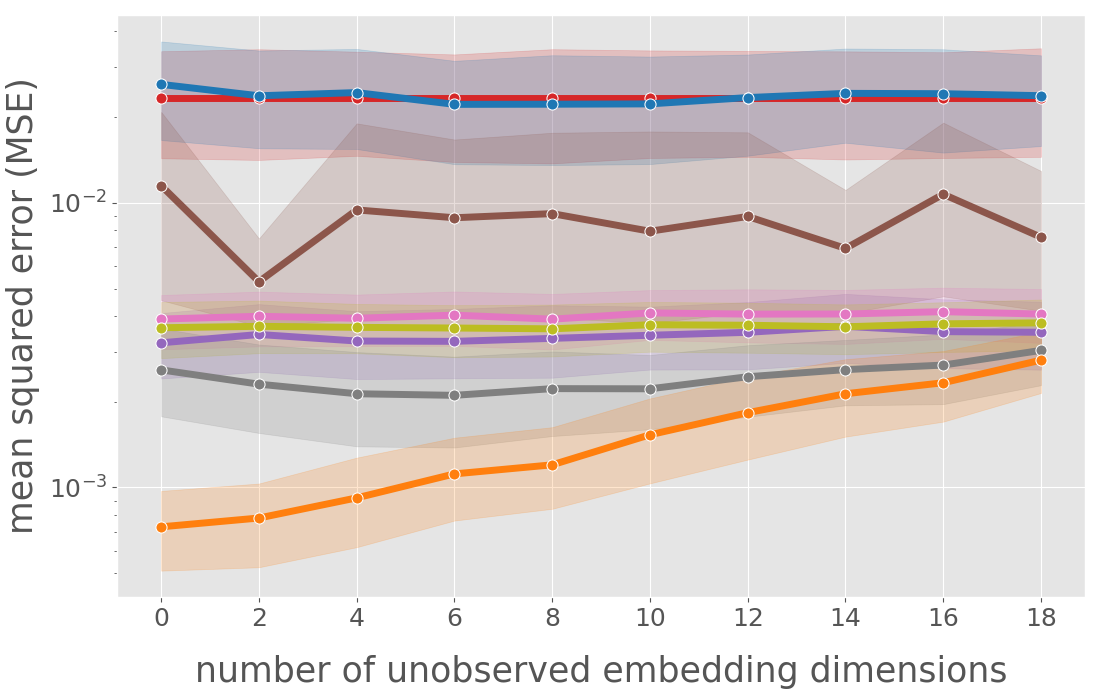}
    \end{center}
\end{minipage}
&
\begin{minipage}{0.30\hsize}
    \begin{center}
        \includegraphics[clip, width=5.8cm]{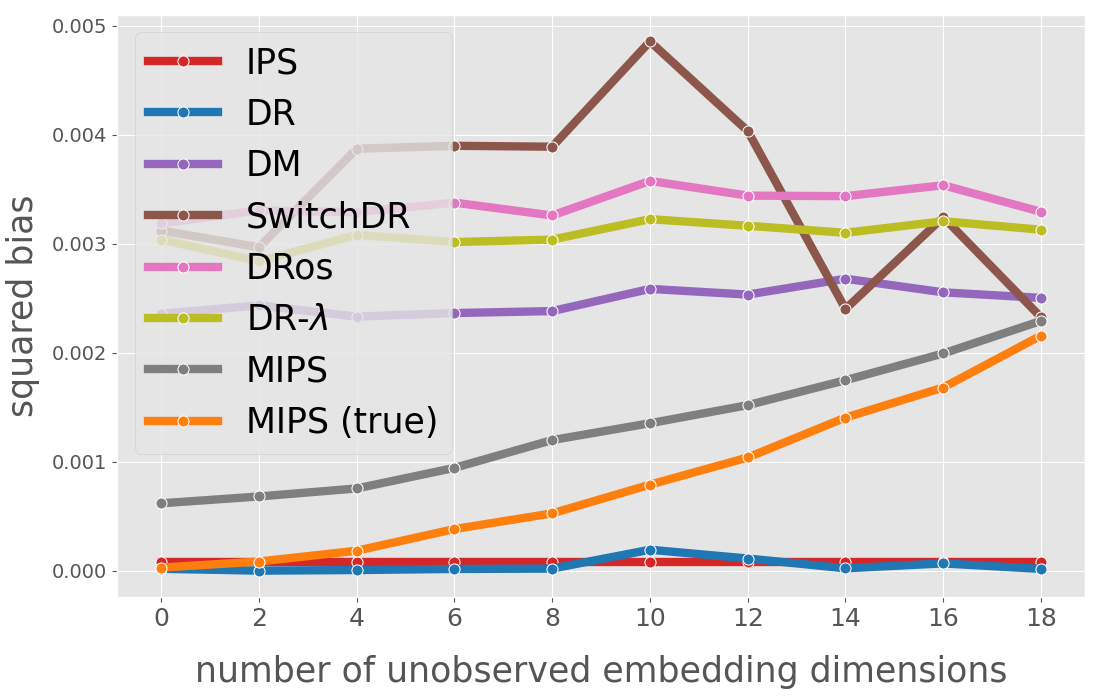}
    \end{center}
\end{minipage}
&
\begin{minipage}{0.30\hsize}
    \begin{center}
        \includegraphics[clip, width=5.8cm]{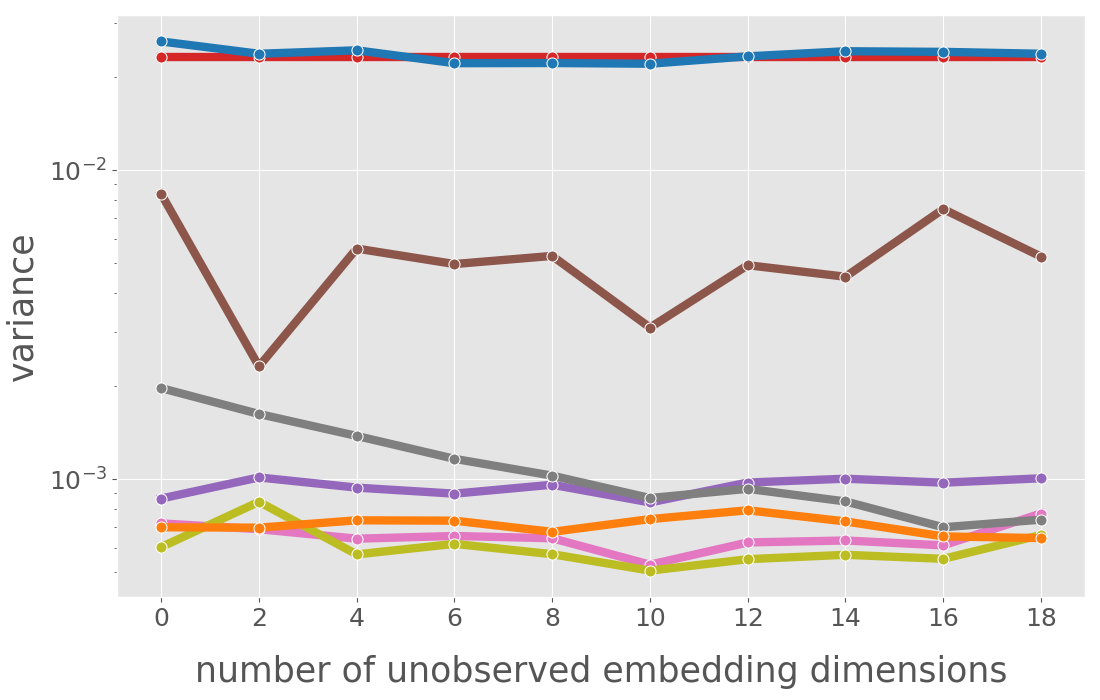}
    \end{center}
\end{minipage}
\\
\multicolumn{3}{c}{
\begin{minipage}{0.95\hsize}
\begin{center}
\caption{MSE, Squared Bias, and Variance with \textbf{varying number of unobserved dimensions in action embeddings}}
\label{fig:beta=1,eps=0.8_end}
\end{center}
\end{minipage}
}
\\ 
\bottomrule
\end{tabular}
}
\vskip 0.1in
\raggedright
\fontsize{9pt}{9pt}\selectfont \textit{Note}:
We set $\beta=1$ and $\epsilon=0.8$, which produce \textbf{logging policy slightly better than uniform random} and \textbf{near-uniform target policy}.
The results are averaged over 100 different sets of synthetic logged data replicated with different random seeds.
The shaded regions in the MSE plots represent the 95\% confidence intervals estimated with bootstrap.
The y-axis of MSE and Variance plots (the left and right columns) is reported on log-scale.
\end{figure*}

\paragraph{Comparison with additional baselines across additional experimental conditions.}

We include additional baselines (Switch-DR, DRos, and DR-$\lambda$) described in Appendix~\ref{app:baselines} to the empirical evaluations. Their built-in hyperparameters are tuned with SLOPE++ proposed by \citet{tucker2021improved}, which slightly improves the original SLOPE of \citet{su2020adaptive}. We use implementations of these advanced estimators provided by OBP (version 0.5.5). We evaluate the four research questions addressed in the main text with six different pairs of $(\beta,\epsilon)$. 
Figures~\ref{fig:beta=-1,eps=0.05_start}-\ref{fig:beta=-1,eps=0.05_end} report the results with $\beta=-1$ and $\epsilon=0.05$. 
Figures~\ref{fig:beta=-1,eps=0.8_start}-\ref{fig:beta=-1,eps=0.8_end} report the results with $\beta=-1$ and $\epsilon=0.8$. 
Figures~\ref{fig:beta=0,eps=0.05_start}-\ref{fig:beta=0,eps=0.05_end} report the results with $\beta=0$ and $\epsilon=0.05$. 
Figures~\ref{fig:beta=0,eps=0.8_start}-\ref{fig:beta=0,eps=0.8_end} report the results with $\beta=0$ and $\epsilon=0.8$. 
Figures~\ref{fig:beta=1,eps=0.05_start}-\ref{fig:beta=1,eps=0.05_end} report the results with $\beta=1$ and $\epsilon=0.05$. 
Figures~\ref{fig:beta=1,eps=0.8_start}-\ref{fig:beta=1,eps=0.8_end} report the results with $\beta=1$ and $\epsilon=0.8$.

In general, we observe results similar to those reported in the main text. Specifically, MIPS works better than all existing estimators, including the advanced ones, in a range of situations, in particular for small data and large action spaces. This result suggests that even the recent state-of-the-art estimators fail to deal with large action spaces. Regarding the additional baselines, Switch-DR, DRos, and DR-$\lambda$ work similarly to DM. These estimators fail to improve their variance with the growing sample sizes and become worse than IPS and DR in large sample regimes. This observation suggests that SLOPE++ avoids huge importance weights and favors low variance, but highly biased estimators in our setup. We indeed also tested the More Robust Doubly Robust (MRDR) estimator~\cite{farajtabar2018more}, but find that MRDR suffers from its growing variance with a growing number of actions and works similarly to IPS and DR.

\clearpage
\begin{algorithm}[t]
\caption{An Experimental Procedure to Evaluate an OPE Estimator with Real-World Bandit Data}
\begin{algorithmic}[1]
\REQUIRE an estimator to be evaluated $\hat{V}$, target policy and corresponding logged bandit data $(\pi, \calD)$, logging policy and corresponding logged bandit data $(\pi_0, \calD_0)$, sample size in OPE $n$, number of random seeds $T$
\ENSURE empirical CDF of the squared error ($\hat{F}_Z$)
\STATE $\mathcal{Z} \leftarrow \emptyset$ (initialize set of results)
\FOR{$t = 1,2, \ldots, T$} \vspace{0.02in}
    \STATE $ \calD_{0,t}^* \leftarrow \mathrm{Bootstrap} (\calD_0;n) $ \hfill \textit{\small // randomly sample size $n$ of bootstrapped samples}  \vspace{0.02in}
    \STATE $ z^{\prime} \leftarrow \big(V_{\mathrm{on}} (\pi;\calD) - \hat{V}(\pi; \calD_{0,t}^*) \big)^2 / \big(V_{\mathrm{on}} (\pi;\calD) - \hat{V}_{\mathrm{IPS}}(\pi; \calD_{0,t}^*) \big)^2$ \hfill \textit{\small // calculate the relative SE of $\hat{V}$ w.r.t IPS}  \vspace{0.05in}
    \STATE $ \mathcal{Z} \leftarrow \mathcal{Z} \cup \{z^{\prime}\} $ \hfill \textit{\small // store the result} \vspace{0.02in}
\ENDFOR
\STATE Estimate CDF of relative SE ($F_Z$) based on $\mathcal{Z}$ (Eq.~\ref{eq:empirical_cdf})
\end{algorithmic}
\label{algo:evaluation_of_ope}
\end{algorithm}

\begin{figure*}[t]
\vspace{0.1in}
\scalebox{0.95}{
\begin{tabular}{ccc}
\begin{minipage}{0.33\hsize}
    \begin{center}
        \includegraphics[clip, width=6.2cm]{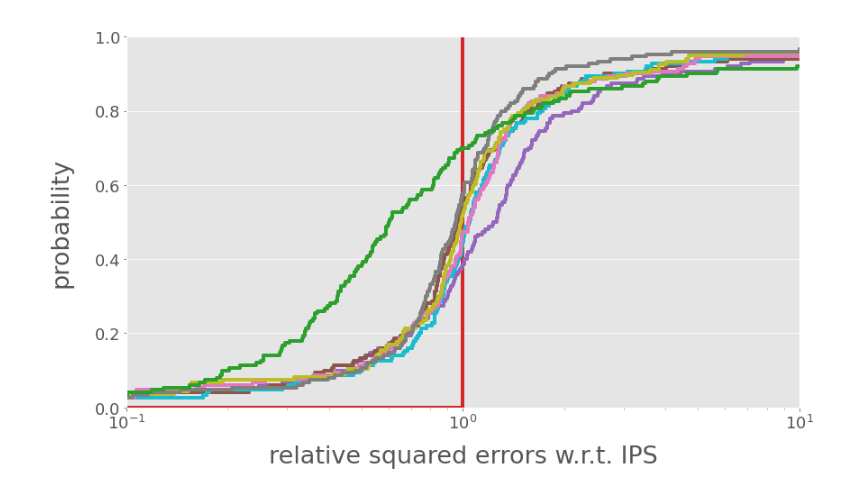}
    \end{center}
\end{minipage}
&
\begin{minipage}{0.33\hsize}
    \begin{center}
        \includegraphics[clip, width=6.2cm]{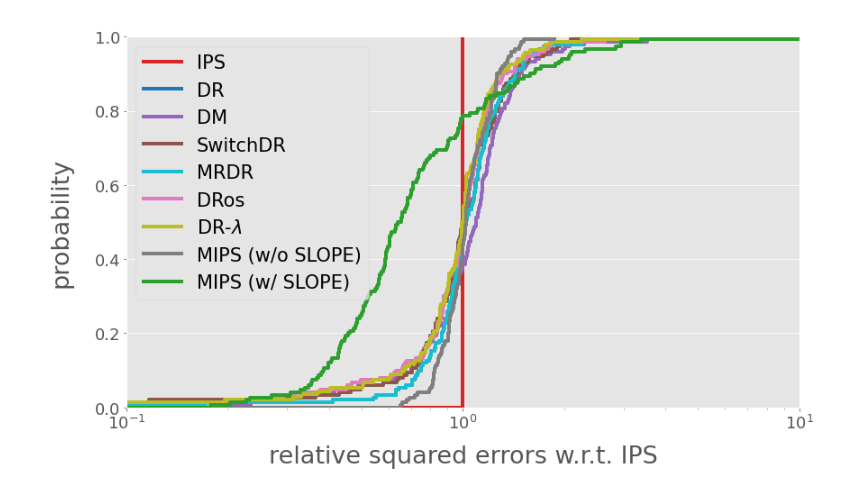}
    \end{center}
\end{minipage}
&
\begin{minipage}{0.33\hsize}
    \begin{center}
        \includegraphics[clip, width=6.2cm]{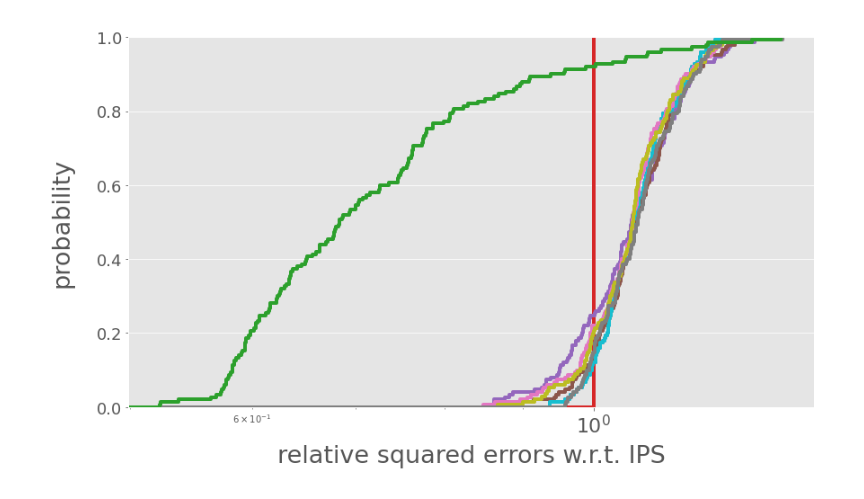}
    \end{center}
\end{minipage}
\\
\multicolumn{3}{c}{
\begin{minipage}{1.0\hsize}
\begin{center}
\caption{CDF of squared errors relative to IPS with different sample sizes (From left to right, $n=10000,50000,500000$). CDFs are estimated with 150 different sets of bootstrapped logged bandit data. Note that the x-axis is reported on a log-scale.}
\label{fig:real_additional}
\end{center}
\end{minipage}
}
\\
\end{tabular}
}
\end{figure*}

\subsection{Experimental Procedure to Evaluate OPE Estimators on Real-World Bandit Data} \label{app:procedure_with_obd}
Following~\citet{saito2020open,saito2021evaluating}, we empirically evaluate the accuracy of the estimators by leveraging two sources of logged bandit data collected by running two different policies denoted as $\pi$ (regarded as target policy) and $\pi_0$ (regarded as logging policy).
We let $\calD$ denote a logged bandit dataset collected by $\pi$ and $\calD_0$ denote that collected by $\pi_0$.
We then apply the following procedure to evaluate the accuracy of an OPE estimator $\hat{V}$.
\begin{enumerate}
    \item Perform bootstrap sampling on $\calD_0$ and construct $\calD_0^* := \{ (x_i^*, a_i^*, r_i^*) \}_{i=1}^{n}$, which consists of size $n$ of independently resampled data with replacement.
    \item Estimate the policy value of $\pi$ using $\calD_0^*$ and OPE estimator $\hat{V}$. We represent a policy value estimated by $\hat{V}$ as $\hat{V} (\pi; \calD_0^*)$.
    \item Evaluate the estimation accuracy of $\hat{V}$ with the following \textit{relative squared error w.r.t IPS} (rel-SE):
    \begin{align*}
        \textit{rel-SE} (\hat{V}; \calD_0^*) := \big(V_{\mathrm{on}} (\pi;\calD) - \hat{V}(\pi; \calD_{0}^*) \big)^2 / \big(V_{\mathrm{on}} (\pi;\calD) - \hat{V}_{\mathrm{IPS}}(\pi; \calD_{0}^*) \big)^2,
    \end{align*}
    where $\hat{V}_{\mathrm{on}} (\pi; \calD) := |\calD|^{-1} \sum_{(\cdot,\cdot,r_j) \in \calD} r_j $ is the Monte-Carlo estimate of $V(\pi)$ based on on-policy data $\calD$. 
    \item Repeat the above process $T$ times with different random seeds, and estimate the CDF of the relative SE as follows.
    \begin{align}
        \hat{F}_{Z}(z):= \frac{1}{T} \sum_{t=1}^T \mathbb{I} \big\{ \textit{rel-SE}_t(\hat{V}; \calD_{0,t}^*) \leq z \big\},
    \label{eq:empirical_cdf}
    \end{align}
    where $\textit{rel-SE} (\hat{V}; \calD_{0,t}^*)$ is the relative SE of $\hat{V}$ computed with the $t$-th bootstrapped samples $\calD_{0,t}^*$.
\end{enumerate}

Algorithm~\ref{algo:evaluation_of_ope} describes this experimental protocol for evaluating OPE estimators in detail.
Figure~\ref{fig:real_additional} reports the results with real bandit data for varying numbers of logged data ($n=10000,50000,500000$).
Note that we use the Random Forest implemented in \textit{scikit-learn} along with 2-fold cross-fitting~\citep{newey2018crossfitting} to obtain $\hat{q}(x,e)$ for the model-dependent estimators. We also use the Categorical Naive Bayes\footnote{\href{https://scikit-learn.org/stable/modules/generated/sklearn.naive_bayes.CategoricalNB.html}{https://scikit-learn.org/stable/modules/generated/sklearn.naive\_bayes.CategoricalNB.html}} to estimate $\hat{\pi}_{0}(a|x,e)$ for MIPS. 

Note that we use OBD's ``ALL" campaign, because it has the largest number of actions among the three available campaigns. We also regard the same action presented at a different position in a recommendation interface as different actions. As OBD has 80 unique actions and 3 different positions in its recommendation interface, the resulting action space has the cardinality of $80 \times 3 = 240$.

\end{document}